%% file: main.tex
\documentclass{article} 
\input{preamble}

\title{TrustMed-RL: Long-Horizon Reinforcement Learning for Evidence-Grounded Clinical Diagnosis}
\author{
    Wenxin Zhan$^{1}$ \quad
    Yizheng Jiao$^{2}$\thanks{Co-second authors.} \quad
    Haifeng Song$^{3}$\footnotemark[1] \\
    Shuai Xu\footnotemark[1] \quad
    Chencheng Pan \quad
    Jiayi Feng$^{4}$ \\
    Anjie Xie$^{5}$\thanks{Corresponding author.} \\[0.5em]
    \parbox{0.95\textwidth}{%
        \centering\normalfont\footnotesize
        $^{1}$Department of Information Systems and Statistics,
        Baruch College \\
        $^{2}$Department of Computer Science,
        University of North Carolina at Chapel Hill \\
        $^{3}$Tsinghua University \quad
        $^{4}$New York University \\
        $^{5}$Zhongshan Hospital (Xiamen), Fudan University
    }
}

\date{}

\begin{document}

\maketitle
\thispagestyle{empty}

\maketitle
\thispagestyle{empty}

\input{sections/abstract}

\input{sections/01_introduction}

\input{sections/related_work}
\input{sections/02_preliminaries}

\input{sections/03_agent_harness}
\input{sections/04_groundedness}
\input{sections/04_training_framework}

\input{sections/05_experiment}
\input{sections/06_ablation}
\input{sections/07_conclusion}


\clearpage
\bibliography{iclr2027_conference}
\bibliographystyle{iclr2027_conference}

\clearpage
\appendix

\input{appendix/a_data_environment}
\input{appendix/b_training_details}
\input{appendix/c_reward_specification}

\input{appendix/d_ablation}
\input{appendix/e_evaluation_details}

\input{appendix/f_prompts}
\input{appendix/g_reasoning_behavior}
\input{appendix/h_OSEC}

\end{document}

%% file: preamble.tex
\usepackage[authoryear,round]{natbib}
\setcitestyle{citesep={;},aysep={,},yysep={;}}
\usepackage{times}
\usepackage{amssymb}
\usepackage{natbib}  
\input{math_commands.tex}

\usepackage{amsmath,amsfonts,bm}
\usepackage{hyperref}
\usepackage{url}
\usepackage{amsthm}
\usepackage{booktabs}
\usepackage{graphicx}
\usepackage{algorithm}
\usepackage{multirow}
\usepackage{algorithmicx}
\usepackage{algpseudocode}
\usepackage{algcompatible}
\usepackage{float}
\usepackage{placeins}

\usepackage[letterpaper,margin=1in]{geometry}
\usepackage[T1]{fontenc}
\usepackage{times}
\graphicspath{{figures/}{./}}

%% file: math_commands.tex
\usepackage{amsmath,amsfonts,bm}

\def\eqref#1{equation~\ref{#1}}

\def\1{\bm{1}}

\DeclareMathAlphabet{\mathsfit}{\encodingdefault}{\sfdefault}{m}{sl}
\SetMathAlphabet{\mathsfit}{bold}{\encodingdefault}{\sfdefault}{bx}{n}



%% file: sections/abstract.tex
\begin{abstract}
Medical language models can produce correct diagnoses despite incomplete investigations and unsupported reasoning. To support long-horizon, evidence-grounded diagnosis, we introduce \textbf{TrustMed-RL}. Built from PubMed rare-disease cases and over 24,000 manually annotated image panels, it integrates interviews, examinations, testing, specialist consultation, and literature search through state-dependent actions. Our 8B vision--language policy, trained with clinically adapted GiGPO and coverage-adjusted diagnostic rewards, achieves 37.1\% diagnostic accuracy on 2,500 evaluation cases, outperforming all evaluated open-weight baselines and improving over supervised fine-tuning by 12.4 percentage points.. When success additionally requires acquiring at least 50\% of supporting test evidence, TrustMed-RL achieves 32.5\%, exceeding GPT-4o by 6.8 percentage points. Furthermore, it surpasses all evaluated baselines on MTMedDialog and multiple larger 27--32B models on AgentClinic. In physician review of 200 diagnostically accepted test-set trajectories, 83.0\% receive evidential-grounding scores of 4--5 out of 5. Physicians' assessments suggest that these diagnostic trajectories are trustworthy and aligned with human diagnostic reasoning.
\end{abstract}

%% file: sections/01_introduction.tex
\section{Introduction.}    
Clinical diagnosis requires an adaptive sequence of investigations. Clinicians elicit histories, select examinations and tests, integrate specialist interpretations, and revise the differential as evidence emerges \citep{balogh2015improving, elstein2002clinical}. The ability to conduct such process becomes particularly important in rare, atypical, or serious disease, where the initial presentation may support several competing explanations. 
Recent medical agents make substantial progress toward interactive diagnosis. DoctorAgent-RL learns multi-turn inquiry policies, PulseMind optimizes multimodal consultation quality, and environments such as AgentClinic and MedEnv support sequential, tool-assisted investigation~\citep{feng2026doctoragent,xu2026pulsemind,schmidgall2024agentclinic,FanZhi_MedEnv_MICCAI2026}. However, tool access and correct final answers alone do not establish that an agent acquired adequate supporting evidence. Models can still produce correct answers despite missing essential images and can fabricate findings to support their conclusions \citep{agbareia2025visual,von2026visual, gu2025illusion}. Evaluating only the diagnostic endpoint can consequently conceal incomplete investigations, unsupported commitments, and failures to recognize unreliable evidence. These limitations motivate two questions: \emph{Can training improve an agent's ability to coordinate the complete diagnostic pathway? Are its diagnostic gains accompanied by better evidence acquisition and more trustworthy decisions?}
We introduce \textbf{TrustMed-RL}, an integrated agent framework, benchmark, and training protocol for \emph{long-horizon, evidence-grounded diagnosis}. Our approach connects diagnostic outcomes to the observable clinical actions and evidence-acquisition processes that precede them. Our contributions are threefold:
\vspace{1pt}
\begin{enumerate}
\item \textbf{An end-to-end clinical agent framework.}
We operationalize the diagnostic pathway through state-dependent actions spanning patient interviews, physical examination, test ordering, specialist consultation, literature search, and diagnosis or deferral. Encounters typically span 20--30 decisions, with a workup space covering 21 physical examinations, 42 diagnostic tests, and eight specialist departments. Specialist routing governs access to recorded interpretations, while explicit integrity judgments assess whether returned images match the requested modality and anatomy. Action-level traces support failure localization, trajectory assessment, and targeted self-improvement data collection.
\item \textbf{A benchmark for groundedness and trustworthiness.}
We extend RareArena with structured clinical records that separate patient-reported facts from clinician-acquired findings and incorporate more than 24,000 manually annotated panel-level images. Panel-specific evidence release limits unintended disclosure from composite figures, while ontology-based diagnosis matching supports consistent scoring. Evaluation jointly examines diagnostic correctness, evidence coverage, productive actions, responses to corrupted evidence, and capability-related deferral, making the evidential and procedural basis of diagnostic performance observable.
\item \textbf{Clinical adaptation of long-horizon policy learning.}
Following supervised fine-tuning, we train the doctor policy using clinically adapted Group-in-Group Policy Optimization (GiGPO) and a ScalingInter-RL-based interaction curriculum~\citep{feng2026group,xi2025agentgym}. To our knowledge, this is the first study to combine these methods for long-horizon multimodal clinical diagnosis. Our adaptations introduce clinical anchors for step-level comparisons, diagnostic rewards adjusted for evidence coverage, local invalid-action penalties, and a curriculum coupling progressive interaction budgets with outcome-adaptive sampling.
\end{enumerate}
Experimental results show that the 8B TrustMed-RL policy achieves 37.1\% diagnostic accuracy on 2,500 evaluation cases, leading all evaluated open-weight models, with findings coverage reaching 73.2\% and productive actions achieving 8.5 per epsiode. When success additionally requires acquiring at least 50\% of available test findings, TrustMed-RL achieves 32.5\%, exceeding GPT-4o by 6.8 percentage points. Furthermore, Additional out-of-distribution (OOD) experiments show both TrustMed-RL variants demonstrate strong generalization across three benchmarks. Source code and data are available at https://github.com/Trustmed-RL/TrustMed-RL

%% file: sections/related_work.tex
\section{Related Work}
\paragraph{Interactive clinical consultation.}
Recent medical agents extend static question answering toward interactive diagnosis. AgentClinic supports multimodal consultations and diagnostic tools, including evaluation of clinical-information coverage~\citep{schmidgall2024agentclinic}. AMIE learns diagnostic dialogue through simulated consultations, while its multimodal extension introduces state-aware dialogue phases and tracks information gaps~\citep{tu2025towards,saab2026advancing}. PulseMind combines multimodal consultation data with comparison-based reinforcement learning and evaluates proactivity, diagnostic accuracy, usefulness, and language quality~\citep{xu2026pulsemind}. These studies establish interactive information gathering and multidimensional consultation assessment as central components of medical-agent evaluation.

\paragraph{Learning and coordinating diagnostic actions.}
DoctorAgent-RL applies GRPO to multi-turn doctor--patient interactions, learning questioning and diagnostic policies under varying dialogue budgets of two to ten turns~\citep{feng2026doctoragent}. MedEnv integrates interviewing, retrieval, test ordering, and chest-radiograph analysis, combining confidence-based process supervision with diagnostic outcome rewards~\citep{FanZhi_MedEnv_MICCAI2026}. DxDirector guides the diagnostic process through requests for physician-assisted clinical operations and evaluates request usefulness, physician involvement, and diagnostic accountability~\citep{xu2026dxdirector}. These approaches advance policy learning and clinical-action coordination while demonstrating the importance of evaluating intermediate decisions alongside final diagnoses.

\paragraph{Our contribution.}
We build on these advances through the joint design of an executable clinical environment and an adaptive learning protocol for \emph{grounded and trustworthy diagnosis}. Our environment connects interviewing, examination, testing, specialist consultation, and literature search to diagnosis or deferral through state-dependent actions. Findings are grounded in case records, specialist-department routing controls access to recorded interpretations, and scored integrity verdicts assess recognition of modality and anatomical mismatches. These controls make evidence acquisition, routing, integrity assessment, and diagnostic commitment explicitly observable and jointly scorable. We further adapt GiGPO through clinical anchors for step-level comparisons, coverage-adjusted diagnostic rewards, local invalid-action penalties, and an outcome-adaptive interaction curriculum. The resulting framework links policy improvement to observable evidence-acquisition behavior and supports analysis of whether diagnostic success is accompanied by sufficient evidence access, appropriate integrity judgments, and reliable clinical actions.

%% file: sections/02_preliminaries.tex
\section{Preliminaries}
\paragraph{Formulation}
\label{sec:formulation}
Following formulations of multi-turn LLM
agents~\cite{xi2025agentgym,xi2026agentgym,feng2026group},
We model a clinical consultation as a finite-horizon partially observable Markov decision process (POMDP)~\cite{kaelbling1998planning}, as illustrated in Figure~\ref{fig:pomdp} in Appendix~\ref{app:pomdp}.
A case $x\sim\mathcal{D}$ induces the POMDP
$M_x=\langle S,O,A,P,\Omega,r,\rho_0^x,H\rangle$.
where $S$, $O$, and $A$ are the state, observation, and textual action
spaces; $P$ and $\Omega$ are the transition and observation kernels;
$r$ is the reward function; $\rho_0^x$ is the case-conditioned
initial-state distribution; and $H$ bounds the number of policy decisions.

The latent state $s_t=(z_x,m_t,\ell_t,\varphi_t,b_t,\zeta_t)$ comprises
case information $z_x$, interaction history $m_t$, evidence state $\ell_t$,
the current interaction phase $\varphi_t$ (history taking, diagnostic
workup, evidence-integrity checking, or conclusion), remaining interaction
budgets $b_t$, and integrity state $\zeta_t$.
At turn $t$, the doctor agent receives
$o_t \sim \Omega(\cdot \mid s_t)$, containing released clinical information
and the current action menu; unreleased findings, the reference diagnosis,
and ground-truth integrity labels remain hidden.
Given the policy context $h_t$ serialized from the observable interaction
history, the doctor agent acts according to
\begin{equation}
    a_t \sim \pi_\theta(\cdot \mid h_t),
    \qquad
    s_{t+1} \sim P(\cdot \mid s_t,a_t),
    \qquad
    r_t = r(s_t,a_t,s_{t+1}).
    \label{eq:interaction}
\end{equation}
The environment parses and validates $a_t$ against the admissible set
$A(s_t)\subseteq A$, determined by the current phase, remaining budgets,
and evidence and integrity states.
The patient simulator and clinical tools are environment components;
only the doctor agent policy is optimized.
The policy context $h_t$ is reconstructed at each turn from the
observable interaction history, including released findings and
previous doctor actions while hidden case information is consistently excluded.
A consultation produces a trajectory
$\tau=(o_0,a_0,\ldots,o_{T-1},a_{T-1},o_T)$, with $T\le H$,
and terminates upon diagnosis or abstention, or is terminated when
an interaction limit is reached.

%% file: sections/03_agent_harness.tex
\section{Agent framework}
\label{sec:agent-framework}

\paragraph{State-dependent clinical interaction.}
Our agent harness implements an end-to-end diagnostic encounter, integrating history-taking, physical examination, diagnostic testing, specialist consultation, and medical information search. Each episode follows a real published case and typically spans 20--30 decisions. The workup action space includes \textbf{21 physical examinations}, \textbf{42 diagnostic tests} and consultation across \textbf{eight specialist departments}. The agent selects actions according to the consultation phase, available evidence, and remaining interaction budget. Each action changes the environment state and subsequent decision options, requiring coordination of evidence acquisition, specialist routing, and diagnostic reasoning before diagnosis or human deferral.

\paragraph{Observable evidence and decision reliability.}
The harness makes evidence acquisition and release explicit and auditable through operational controls: ordering and specialist interpretation constitute distinct operations; report release is contingent on appropriate routing;and unavailable examinations return explicit availability statuses.  In addition, failed operations, including invalid actions, elicit deterministic feedback and remain recorded in the trajectory. Together, these controls support joint assessment of diagnostic correctness, evidence coverage, routing, responses to corrupted evidence, action validity, and deferral under capability limitations. 
By situating each diagnosis within the evidence available at the time and the preceding actions, the harness supports scrutiny of unsupported commitments and intermediate failures that final-answer accuracy alone cannot capture.

\paragraph{Failure analysis and self-improvement.}
These auditable trajectories link each action to its preceding context, returned observation, and evaluation feedback, supporting localization of missed evidence, routing errors, invalid requests, and indiscriminate deferral. This feedback can guide revisions of failed consultations, with trajectories passing checks for diagnostic correctness and procedural validity retained as supervised training examples~\citep{yuan2025agent}. Successful and failed attempts on the same case can also form preference pairs; repeating trajectory generation and screening after each training round refreshes the data for iterative self-improvement~\citep{xu2026medagentgym}.

%% file: sections/04_groundedness.tex
\section{Groundedness and Trustworthiness}
\label{sec:groundedness}
Groundedness and trustworthiness are explicit design requirements across
our data annotation, preprocessing, agent harness, and training framework.
Long-horizon clinical diagnosis requires an agent to acquire evidence,
navigate evolving clinical states, and coordinate heterogeneous actions
before committing to a diagnosis. We therefore design these components
as an integrated system that makes evidence provenance, information
access, intermediate decisions, and diagnostic outcomes observable
throughout the consultation. Their interdependence supports both
faithful evaluation and systematic policy improvement.
\vspace{1pt}
\paragraph{Traceable clinical evidence through manual annotation.}
We extend RareArena~\citep{chen2026rarearena} into a multimodal, long-horizon agentic
benchmark with structured clinical evidence. Our image preprocessing
pipeline draws on 44{,}277 registered case figures from PubMed Central.
Large-scale manual annotation identifies semantic panels and records
their identity, source order, modality, anatomic site, laterality, and
pixel coordinates, yielding more than \textbf{24{,}000 panel-level crops}.
Each image crop is linked to its source examination, allowing evidence to be released selectively by examination and routed to the appropriate specialist. This mapping makes the visual evidence presented to the agent traceable while limiting inadvertent disclosure of unrelated findings within composite figures.
\vspace{1pt}
\paragraph{Controlled access to atomic clinical facts.}
Our preprocessing pipeline organizes cases into \textbf{ OSCE}-style atomic clinical facts, distinguishing patient-reported information from clinician-discovered findings (Figure~\ref{fig:clinical-example}). This separation defines what the simulated patient may disclose and which findings require further investigation. By combining these records with panel-specific evidence release, the harness tracks newly acquired evidence, repeated requests, and unexplored findings, preserving and strengthening consistent evaluation of action productivity and evidence coverage across variable-length, multi-state consultations.
\vspace{1pt}
\paragraph{Ontology-grounded diagnostic supervision.}
Reliable policy learning requires diagnostic feedback that remains
consistent across terminology variants. Integrating MedGen, MONDO, and ORDO, our ontology-based matching
framework maps predicted diagnoses and reference labels to standardized
disease concepts. Compared with lexical-overlap scoring and
LLM-only diagnostic assessment,
our framework incoportaes ontology-based normalization with LLM
judgment. This combination recognizes clinically equivalent diagnoses
despite lexical variation and assesses semantic similarity against
standardized disease concepts, preserving clinically meaningful
distinctions between related diagnoses. The same scoring procedure
underpins training rewards and evaluation, providing a consistent
criterion for diagnostic correctness.
Table~\ref{tab:class-ablation-100} presents reference diagnosis pairs
illustrating accepted equivalences and distinctions, making the
scoring criteria transparent and auditable.
\vspace{1pt}
\paragraph{Retrievable medical knowledge at scale.}
To support reasoning beyond the agent's parametric knowledge, we construct
a medical corpus following MedRAG~\citep{zhao2025medrag}, comprising
29{,}650{,}414 passages. These include 29{,}166{,}067 PubMed
title-plus-abstract passages from the 2026 MEDLINE baseline and daily
updates through July 30, 2026; 358{,}500 StatPearls paragraph-level passages
titled by section paths; and 125{,}847 passages from 18 textbooks.
All passages are embedded using the MedCPT article encoder. Retrieval
provides identifiable external references that complement case-specific
observations and support inspection of the knowledge available during
diagnostic reasoning.

\paragraph{Evidence-aware training for long-horizon diagnosis.}
We adapt GiGPO~\citep{feng2026group}, with GRPO~\citep{shao2024deepseekmath}
as an episode-level alternative, and extend
ScalingInter-RL~\citep{xi2026agentgym} within a common clinical training
protocol. Clinical anchors support credit assignment across different
consultation histories and evidence states, while local format penalties
discourage invalid intermediate actions. Outcome-adaptive sampling and
progressively expanded interaction budgets expose the policy to cases
with increasingly demanding interaction requirements. Crucially,
diagnostic rewards depend jointly on correctness and released-report
coverage, and selective reinforcement
strengthens eligible positive updates from correct consultations.
These adaptations make evidence acquisition and reliable interaction
explicit training objectives across the complete diagnostic pathway.
Coverage operationalizes evidence acquisition as a grounding proxy.

\paragraph{Auditable evaluation and iterative self-improvement.}
We evaluate complete consultations through diagnostic correctness,
evidence coverage, action productivity, and performance across coverage
levels. Joint assessment distinguishes successful evidence acquisition
from correct terminal prediction and reveals behaviors such as repeated
requests, incomplete investigation, and correct diagnoses reached with
limited evidence coverage. These observations provide actionable
feedback for subsequent task sampling, policy training, and evaluation,
supporting \textbf{iterative self-improvement} under consistent criteria.
Each component supplies a prerequisite for this feedback loop:
annotation establishes provenance, preprocessing defines observable
facts, the harness records evidence access, ontology-based scoring
provides consistent outcome feedback, and training converts that feedback
into policy updates. Their integration makes long-horizon diagnostic
behavior auditable and grounds assessments of robustness and
trustworthiness in observable actions, acquired evidence, and outcomes.

%% file: sections/04_training_framework.tex
\section{Training Framework}
\label{sec:training}

As illustrated in Appendix Figure~\ref{fig:training}, we train the physician policy in two stages. Supervised fine-tuning (SFT) uses curated multi-turn consultations generated by GPT-5-mini and Qwen3.8-max to initialize consultation behavior. Reinforcement learning then optimizes the policy through simulated encounters. We adapt Group-in-Group Policy Optimization (GiGPO; \citealp{feng2026group}) with clinical state comparisons and evidence-conditioned diagnostic rewards, and extend the ScalingInter-RL interaction curriculum \citep{xi2026agentgym} with outcome-adaptive sampling. Group Relative Policy Optimization (GRPO; \citealp{shao2024deepseekmath}) provides an episode-level alternative under the same clinical reward protocol. Each RL update samples 16 tasks and collects nine consultations per task, yielding 144 trajectories. Training runs for 256 updates, with KL regularization against the frozen SFT policy.

\paragraph{Evidence-conditioned diagnostic reward.}
Let $d_i\in[0,1]$ denote diagnostic correctness credit, $M_i$ the number of available image-bearing evidence records, and $L_i$ the number of distinct records whose specialist reports have been released. For $M_i>0$, define $c_i^{\mathrm{img}}=\min\{1,L_i/M_i\}$. The diagnosis reward is
\begin{equation}
R_i^{\mathrm{dx}}=
\begin{cases}
d_i\bigl(0.10+0.90\sqrt{c_i^{\mathrm{img}}}\bigr), & M_i>0,\\
d_i, & M_i=0.
\end{cases}
\label{eq:training-reward}
\end{equation}
This adjustment gives greater credit to correct diagnoses accompanied by report acquisition. Coverage alone earns no diagnosis credit, and a correct diagnosis on an image-bearing case receives only base credit when no specialist report has been released. Ordering or viewing an image does not increase coverage, and repeated requests do not increase $L_i$. The adjustment applies once, at diagnosis; evidence acquisition receives no separate positive reward. Training coverage $c_i^{\mathrm{img}}$ concerns image-bearing records, whereas evaluation coverage $c_i^{\mathrm{test}}$ includes all available laboratory, physiologic, imaging, and pathology test records. Appendix C specifies correctness scoring and the terminal rewards for corruption detection, deferral, and invalid termination.

\paragraph{Delayed outcomes and local penalties.}
Let $R_i$ be the terminal task reward and $T_i$ the consultation length. Diagnosis outcomes use $R_i^{\mathrm{dx}}$, subject to the overrides in Appendix C. Set $v_{it}=0$ for a penalized format violation and $v_{it}=1$ otherwise. The episode- and step-level credit estimators receive
\begin{equation}
q^E_{it}=R_i-0.01(1-v_{it}),
\qquad
q^S_{it}=0.99^{T_i-t}R_i-0.01(1-v_{it}).
\label{eq:training-credit-scores}
\end{equation}
The step-level score discounts the delayed outcome according to the remaining consultation length. Each format violation incurs the same local cost, independent of diagnostic credit or action type. This cost is assigned to the offending action; it is neither discounted nor propagated to other decisions.

\paragraph{Credit assignment across clinical states.}
For each case, we collect $K\geq2$ consultations with matched initial context, interaction budgets, and corruption realization. Each rollout contributes one reference score to the episode-level statistics, preventing longer consultations from dominating the baseline. GiGPO additionally groups decisions through the environment-defined abstraction $\widetilde{s}_{it}=\phi(s_{it})$, which represents consultation phase, progress, and evidence status. These clinical anchors support step-level comparisons at similar stages of investigation despite different textual histories and trajectory lengths. The episode- and step-level advantages, $A^E_{it}$ and $A^S_{it}$, are normalized using their respective group means and standard deviations. Step-level credit is zero when an anchor has insufficient comparisons, and a standard-deviation floor of $0.125$ stabilizes low-variance groups.

\paragraph{Selective reinforcement of evidence-grounded correct diagnoses.}
After normalization, we strengthen eligible positive episode-level updates:
\begin{align}
A_{it}&=\bigl[1+(\alpha_{\mathrm{dx}}-1)e_{it}\bigr]A^E_{it}+\omega A^S_{it},\\
e_{it}&=\chi_i v_{it}\mathbf{1}\{A^E_{it}>0\}
\mathbf{1}\{A^E_{it}+\omega A^S_{it}>0\}.
\label{eq:joint-relative}
\end{align}
Here, $\chi_i$ indicates a full-credit correct diagnosis under the training rubric, $\alpha_{\mathrm{dx}}\geq1$ controls the multiplier, and $\omega\geq0$ weights step-level credit. The multiplier applies only to valid actions with positive episode-level and combined advantages. Applying it after normalization preserves the comparison statistics and leaves nonpositive combined updates unchanged. Evidence acquisition influences these advantages through the diagnostic reward; the multiplier introduces no additional coverage criterion. Setting $\omega=0$ yields the GRPO-style episode-level alternative. Our experiments use $\alpha_{\mathrm{dx}}=1.2$ and $\omega=1$ for GiGPO.

\paragraph{Outcome-adaptive task sampling.}
At a visit to task $x$, we summarize variation across its $K$ rollout outcomes. For rollout $i$, the sampling score includes cumulative invalid-action costs:
\begin{equation}
\widetilde{R}_{xi}=
\operatorname{clip}_{[0,1]}\!\left(
\frac{R_{xi}-\lambda_{\mathrm{inv}}\sum_{t=1}^{T_{xi}}(1-v_{xit})}
{R^{\mathrm{ref}}_{k_x}(n)}
\right),
\qquad
\sigma_x=\operatorname{Std}_{i=1}^{K}(\widetilde{R}_{xi}),
\label{eq:training-sampling}
\end{equation}
where $\lambda_{\mathrm{inv}}=0.01$, $k_x$ is the task type, $R^{\mathrm{ref}}_{k_x}(n)>0$ is its reference reward at update $n$, and $\operatorname{Std}$ denotes population standard deviation. Priorities track a moving average of visit dispersion, initialized from offline difficulty estimates. Following prioritized level replay \citep{jiang2021prioritized}, selection combines priority ranks and staleness within task-type strata, reserves uniform exploration slots, and retires consistently high-performing, low-dispersion tasks.

\paragraph{Progressive interaction curriculum.}
At update $n$, phase $p(n)$ determines the interaction-budget ceiling and the admitted task pool:
\begin{equation}
\mathcal{D}_n=\{x:h_{x,J_x(n)}\leq p(n)\},
\qquad
\mathbf{b}_1\preceq\cdots\preceq\mathbf{b}_P=\mathbf{b}_{\mathrm{eval}}.
\label{eq:training-curriculum}
\end{equation}
Here, $J_x(n)$ counts completed visits, $h_{x,J_x(n)}$ is the estimated admission phase, and $\mathbf{b}_p$ specifies question, workup, and search ceilings. Phase transitions follow a fixed schedule; successful trajectories can refine task admission estimates. The four phases begin with ceilings of four questions, ten workup turns, and one search, increasing to ten questions, twenty workup turns, and three searches in the final phase. Adaptive sampling selects tasks within the admitted pool, while matched budgets preserve comparable conditions within each rollout group. The final budget ceiling matches evaluation. Appendix ~\ref{app:implementation-details} provides the admission rules, budget randomization, sampling schedules, and optimization hyperparameters.

%% file: sections/05_experiment.tex
\section{Experiment}
\paragraph{Experimental setting.}
We use \texttt{gpt-5.6-terra} to convert RareArena case reports
and associated PMC figures into diagnosis-blind OSCE-style cases,
separating patient-reported histories from examination findings
and diagnostic evidence available through workup.
Following TruthRL~\citep{wei2025truthrl}, hierarchical diagnostic
probes classify cases as easy, medium, hard, or empirically
unsolved under the tested models and budgets.

We train Qwen3-VL-8B-Instruct~\citep{bai2025qwen3} on the RareArena-derived environment and evaluate on 2,500 clean diagnostic cases. Comparisons include the pretrained backbone, supervised fine-tuning (SFT), the GRPO alternative, and general-purpose and medical vision–language models. SFT, GiGPO, and GRPO
use three training seeds, with three separately seeded evaluation passes per trained model. We additionally evaluate transfer toAgentClinic-NEJM, AgentClinic-MedQA~\citep{schmidgall2024agentclinic},
and MTMedDialog~\citep{feng2026doctoragent}, evidence-
integrity detection, and physician-assessed reasoning quality. All models interact with the clinical environment and
provide free-form diagnoses. Appendix ~\ref{app:implementation-details} describes training and interaction budgets; Appendix~\ref{app:evaluation-details} specifies diagnosis matching and judge validation.

\subsection{Evaluation}
\label{sec:diagnosis-evaluation}
\paragraph{Condition extraction and semantic matching.}
For case $i$, let $\mathcal{G}_i$ and $\mathcal{P}_i$ denote
the ground-truth and predicted condition sets, with
$m_i=|\mathcal{G}_i|$ and $n_i=|\mathcal{P}_i|$.
Each judge $j\in\{1,\ldots,J\}$ labels candidate pairs $(g,p)$
as \textsc{full}, \textsc{core}, \textsc{partial}, or \textsc{null} (Appendix~\ref{app:diagnosis-matching}).
The corresponding weights satisfy
$w_{\mathrm F}=1>w_{\mathrm C}>0.5
\geq w_{\mathrm P}>w_{\varnothing}=0$.

To mitigate judge sensitivity, a deterministic offline oracle based
on MedGen, MONDO, and ORDO supplies a label floor
$b_i(g,p)\in\{\varnothing,\mathrm C,\mathrm F\}$.
Reward computation uses the augmented label
$\widetilde{\ell}_{ij}(g,p)
=\max_{\prec}\{\ell_{ij}(g,p),b_i(g,p)\}$,
where $\varnothing\prec\mathrm P\prec\mathrm C\prec\mathrm F$.

\paragraph{Soft diagnostic scores.}
Let $F_{ij}$, $C_{ij}$, and $P_{ij}$ count the
\textsc{full}, \textsc{core}, and \textsc{partial} matches in a
one-to-one matching. The soft matched mass is
$s_{ij}=w_{\mathrm F}F_{ij}+w_{\mathrm C}C_{ij}
+w_{\mathrm P}P_{ij}$, with
$0\leq s_{ij}\leq\min\{m_i,n_i\}$.
The residual false-positive and false-negative masses are
$\mathrm{FP}^{s}_{ij}=n_i-s_{ij}$ and
$\mathrm{FN}^{s}_{ij}=m_i-s_{ij}$.
For $\alpha,\beta\geq0$, the soft Tversky
score~\citep{salehi2017tversky} is
\begin{equation}
T_{\alpha,\beta}^{(ij)}
=
\frac{s_{ij}}{s_{ij}+\alpha(n_i-s_{ij})+\beta(m_i-s_{ij})}
=
\frac{s_{ij}}{\alpha n_i+\beta m_i},
\qquad \alpha+\beta=1.
\label{eq:tversky-derivation}
\end{equation}
Here, $\alpha$ and $\beta$ control the penalties for unsupported
predicted diagnoses and missed ground-truth diagnoses, respectively.
All reported settings satisfy $\alpha+\beta=1$.
For $m_i,n_i\geq1$, soft precision and recall are
$\mathrm{Prec}_{ij}=s_{ij}/n_i$ and
$\mathrm{Rec}_{ij}=s_{ij}/m_i$, measuring semantic support for
predicted conditions and coverage of ground-truth conditions.
When $s_{ij}>0$, the Tversky score is their weighted harmonic mean:
$(T_{\alpha,\beta}^{(ij)})^{-1}
=\alpha/\mathrm{Prec}_{ij}+\beta/\mathrm{Rec}_{ij}$.

\paragraph{Diagnostic accuracy.}
Accuracy requires every ground-truth condition to receive a
\textsc{full} or \textsc{core} match,
$\mathrm{Acc}_{ij}=\mathbf{1}\{F_{ij}+C_{ij}=m_i\}$;
strict accuracy requires \textsc{full} matches throughout,
$\mathrm{Acc}^{\mathrm{strict}}_{ij}=\mathbf{1}\{F_{ij}=m_i\}$.
Additional predicted conditions are penalized by the soft scores
but do not invalidate these accuracy criteria.
Non-diagnosis outcomes receive zero.

\paragraph{Evidence acquisition.}
Let $\mathcal{T}_i$ contain the available laboratory, physiologic, imaging, and pathology test records, and let $\mathcal{R}_i\subseteq\mathcal{T}_i$ contain records whose nonempty findings reach the agent before termination. Evaluation coverage is $c_i^{\mathrm{test}}=|\mathcal{R}_i|/|\mathcal{T}_i|$ when test records are available, and zero otherwise. This measure includes all four test categories; the training reward instead uses released specialist reports for image-bearing records. Neither measure counts viewing a raw image as receiving its report.

We define $\mathrm{Acc@50}=\frac{100}{NJ}\sum_{i=1}^{N}\sum_{j=1}^{J}\mathrm{Acc}_{ij}\mathbf{1}\{c_i^{\mathrm{test}}\geq0.5\}$, where $J$ is the number of judge assessments per case, and define $\mathrm{Strict@50}$ analogously. Both retain all $N=2{,}500$ cases in the denominator, including cases without available test records. Mean ordering and findings coverage use the 2,452 eligible cases. Productive actions reveal previously unseen history facts, images, or findings; workup turns, search use, and invalid actions characterize consultation behavior. Coverage measures acquisition of available evidence, while physician review separately examines its interpretation and use.

\subsection{Results}
\paragraph{Diagnostic performance.} TrustMed-RL (GiGPO, 8B) achieves the highest open-weight accuracy (37.1\%), precision (31.8\%), and recall (34.6\%) (Table~\ref{tab:diagnostic-performance}). Its accuracy exceeds the 8B SFT model by 12.4 percentage points, an approximately 50\% relative gain, and HealthGPT-Pro-27B, the strongest external open-weight baseline by accuracy, by 2.6 points. GiGPO also improves over GRPO (34.5\%) and approaches GPT-4o (37.9\%). GPT-5 remains the strongest model overall.

\paragraph{Accuracy with acquired evidence.}
GiGPO achieves the highest open-weight Acc@50 (32.5\%) and
Strict@50 (13.5\%), showing that its advantage persists when
diagnostic correctness must be accompanied by substantial
findings acquisition.
Its Acc@50 exceeds GPT-4o by 6.8 percentage points and
HealthGPT-Pro-27B by 9.0 points.
Moreover, GiGPO and GRPO incur the two smallest relative
accuracy reductions across all evaluated models:
12.3\% and 10.7\%, respectively, compared with 23.9\%
for SFT, 31.8\% for HealthGPT-Pro-27B, and 32.3\% for GPT-4o. GiGPO therefore retains 87.7\% of its diagnostic accuracy when substantial findings acquisition is required. These results establish stronger
joint performance on diagnosis and evidence acquisition; physician assessment further examines how acquired
observations support clinical reasoning.

\paragraph{RL supports sustained and productive agentic investigation.}
\vspace{1pt}
The diagnostic gains accompany stronger evidence acquisition and productive interaction (Table~\ref{tab:workup-integrity-limits}). GiGPO and GRPO achieve ordering coverage of 83.9\% and 84.7\% and findings coverage of 73.2\% and 73.9\%, respectively, exceeding GPT-5 (72.4\% and 62.8\%) and GPT-4o (62.0\% and 53.5\%). GiGPO's search
use rises from 53.2\% to 93.0\% of episodes, its productive actions increase from 7.7 per episode under SFT to 8.5 for GiGPO and 9.1 for GRPO, surpassing GPT-4o (6.9) and matching GPT-5 (8.5). Since productive actions reveal previously unseen information, these results indicate that sustained interactions yield additional evidence. Together with the coverage-gated diagnostic results, these behaviors demonstrate learned agentic workup in long-horizon clinical interactions, with the policy coordinating evidence acquisition, specialist consultation, and search toward diagnostic resolution.
\vspace{1pt}
\begin{table*}[t]
\centering
\caption{\textbf{Diagnostic performance on 2,500 clean cases.} Scores are percentages; $\pm$ denotes reported standard deviations. Acc and Strict require recovery of every reference condition under core-or-full and full matching, respectively. Acc@50 and Strict@50 additionally require at least 50\% test-findings coverage. Bold marks the best open-weight mean in each column. Complete comparisons are provided in Appendix Table~\ref{app:diagnostic-performance}.}
\label{tab:diagnostic-performance}
\begingroup
\setlength{\tabcolsep}{3pt}
\renewcommand{\arraystretch}{1.12}
\providecommand{\tmstd}[2]{\ensuremath{#1_{\scriptscriptstyle\pm #2}}}
\scriptsize
\resizebox{\textwidth}{!}{%
\begin{tabular}{@{}l*{8}{r}@{}}
\toprule
Model & Acc & Acc$^{\mathrm{strict}}$ & Prec. & Rec. & Acc@50 & Strict@50 & $\Delta_{\mathrm{Acc}}$ (\%) & $\Delta_{\mathrm{strict}}$ (\%) \\
\midrule
\multicolumn{9}{@{}l}{\textit{Proprietary (closed-source) models}} \\
\addlinespace[2pt]
GPT-5 & $54.2$ & $22.0$ & $42.9$ & $50.0$ & $43.2$ & $17.7$ & $20.3$ & $19.5$ \\
GPT-4o & $37.9$ & $20.8$ & $34.9$ & $36.0$ & $25.7$ & $14.3$ & $32.3$ & $31.3$ \\
\midrule
\multicolumn{9}{@{}l}{\textit{Open-weight models}} \\
\addlinespace[2pt]
TrustMed-RL (GiGPO, 8B) & \tmstd{\mathbf{37.1}}{1.1} & \tmstd{15.3}{1.0} & \tmstd{\mathbf{31.8}}{1.1} & \tmstd{\mathbf{34.6}}{1.1} & \tmstd{\mathbf{32.5}}{1.3} & \tmstd{\mathbf{13.5}}{0.6} & $12.3$ & $11.9$ \\
TrustMed-RL (GRPO, 8B) & \tmstd{34.5}{1.6} & \tmstd{13.6}{1.9} & \tmstd{28.9}{1.9} & \tmstd{32.1}{1.5} & \tmstd{30.8}{1.3} & \tmstd{12.2}{1.6} & $\mathbf{10.7}$ & $\mathbf{10.9}$ \\
TrustMed-SFT (8B) & \tmstd{24.7}{0.6} & \tmstd{9.0}{0.3} & \tmstd{19.0}{0.4} & \tmstd{22.9}{0.6} & \tmstd{18.8}{0.4} & \tmstd{6.9}{0.3} & $23.9$ & $23.2$ \\
TrustMed-SFT (2B) & \tmstd{13.0}{0.1} & \tmstd{3.8}{0.3} & \tmstd{9.3}{0.2} & \tmstd{12.3}{0.2} & \tmstd{8.9}{0.1} & \tmstd{2.7}{0.4} & $31.1$ & $30.3$ \\
Qwen3-VL-8B-Instruct \citep{bai2025qwen3} & \tmstd{16.3}{0.3} & \tmstd{6.4}{0.5} & \tmstd{13.7}{0.2} & \tmstd{15.3}{0.3} & \tmstd{6.6}{0.2} & \tmstd{2.7}{0.1} & $59.2$ & $58.0$ \\
\midrule
PulseMind-72B & \tmstd{21.8}{0.3} & \tmstd{10.1}{0.3} & \tmstd{19.2}{0.2} & \tmstd{20.9}{0.1} & \tmstd{11.1}{0.1} & \tmstd{5.5}{0.3} & $49.4$ & $45.8$ \\
\midrule
HealthGPT-Pro-27B & \tmstd{34.5}{0.7} & \tmstd{\mathbf{18.1}}{0.9} & \tmstd{30.8}{1.0} & \tmstd{32.6}{0.8} & \tmstd{23.5}{0.8} & \tmstd{12.7}{0.8} & $31.8$ & $30.0$ \\
Qwen3-VL-32B-Instruct & $30.4$ & $14.9$ & $26.9$ & $28.6$ & $20.9$ & $10.2$ & $31.2$ & $31.4$ \\
MedGemma-27B-it \citep{sellergren2025medgemma} & $30.4$ & $14.9$ & $27.6$ & $28.7$ & $20.9$ & $10.7$ & $31.1$ & $28.4$ \\
Hulu-Med-32B \citep{jiang2025hulu} & $23.4$ & $10.3$ & $19.8$ & $22.1$ & $14.4$ & $6.7$ & $38.4$ & $35.3$ \\
Lingshu-32B \citep{xu2026lingshu} & $21.5$ & $10.0$ & $19.5$ & $20.6$ & $13.9$ & $7.0$ & $35.5$ & $29.3$ \\
QoQ-Med-VL-32B \cite{dai2026qoq} & $13.4$ & $4.8$ & $11.5$ & $13.1$ & $8.6$ & $3.7$ & $36.3$ & $21.8$ \\
\bottomrule
\end{tabular}%
}
\endgroup
\end{table*}

\begin{table*}[t]
\centering
\caption{\textbf{Evidence acquisition and consultation behavior.} Order and Findings are mean coverage over 2,452 eligible clean cases. Prod. counts actions yielding new facts, images, or findings; Workup counts workup turns per clean episode. Search is the percentage of episodes with search; Invalid is the percentage of invalid turns. Modality and Anatomy are exposure-conditioned mismatch catch rates. False alarm is measured on matched clean-image gate-replay controls. All metrics except Prod. and Workup are percentages. Complete comparisons appear in Appendix Table~\ref{tab:workup-integrity-limits}.}
\label{tab:workup-integrity-limits}
\begingroup
\setlength{\tabcolsep}{3pt}
\renewcommand{\arraystretch}{1.12}
\providecommand{\tmstd}[2]{\ensuremath{#1_{\scriptscriptstyle\pm #2}}}
\scriptsize
\resizebox{\textwidth}{!}{%
\begin{tabular}{@{}l*{9}{r}@{}}
\toprule
Model & Order & Findings & Prod. & Workup & Search & Invalid
& Modality Catch & Anatomy Catch & False Alarm \\
\midrule
\multicolumn{10}{@{}l}{\textit{Proprietary (closed-source) models}} \\
\addlinespace[2pt]
GPT-5 & $72.4$ & $62.8$ & $8.5$ & $12.8$ & $63.2$ & $0.0$ & $97.8$ & $88.7$ & $4.8$ \\
GPT-4o & $62.0$ & $53.5$ & $6.9$ & $10.9$ & $75.6$ & $0.0$ & $95.4$ & $85.7$ & $6.4$ \\
\midrule
\multicolumn{10}{@{}l}{\textit{Open-weight models}} \\
\addlinespace[2pt]
TrustMed-RL (GiGPO, 8B) & \tmstd{83.9}{4.0} & \tmstd{73.2}{3.9} & \tmstd{8.5}{0.4} & \tmstd{17.8}{0.8} & \tmstd{93.0}{2.5} & \tmstd{4.0}{1.1} & \tmstd{82.2}{1.1} & \tmstd{52.6}{4.2} & \tmstd{2.2}{0.8} \\
TrustMed-RL (GRPO, 8B) & \tmstd{\mathbf{84.7}}{1.3} & \tmstd{\mathbf{73.9}}{1.1} & \tmstd{\mathbf{9.1}}{0.2} & \tmstd{18.3}{0.6} & \tmstd{89.3}{4.7} & \tmstd{2.6}{0.2} & \tmstd{85.4}{7.9} & \tmstd{60.6}{2.4} & \tmstd{2.4}{0.5} \\
TrustMed-SFT (8B) & \tmstd{66.4}{0.8} & \tmstd{57.1}{0.7} & \tmstd{7.7}{0.0} & \tmstd{10.2}{0.1} & \tmstd{53.2}{1.4} & \tmstd{\mathbf{0.6}}{0.0} & \tmstd{92.2}{1.9} & \tmstd{\mathbf{73.3}}{1.4} & \tmstd{3.9}{0.8} \\
TrustMed-SFT (2B) & \tmstd{59.4}{0.1} & \tmstd{51.3}{0.2} & \tmstd{7.2}{0.0} & \tmstd{11.1}{0.1} & \tmstd{50.7}{1.5} & \tmstd{6.9}{0.1} & \tmstd{80.8}{5.7} & \tmstd{36.2}{10.1} & \tmstd{5.0}{0.4} \\
Qwen3-VL-8B-Instruct & \tmstd{36.1}{1.0} & \tmstd{30.8}{0.8} & \tmstd{4.7}{0.0} & \tmstd{5.6}{0.1} & \tmstd{18.1}{0.9} & \tmstd{7.1}{0.3} & \tmstd{\mathbf{92.9}}{5.0} & \tmstd{57.0}{3.6} & \tmstd{4.6}{2.3} \\
\midrule
PulseMind-72B & \tmstd{45.4}{0.6} & \tmstd{37.6}{0.6} & \tmstd{5.1}{0.0} & \tmstd{7.2}{0.0} & \tmstd{36.6}{1.7} & \tmstd{5.3}{0.0} & --- & --- & --- \\
\midrule
HealthGPT-Pro-27B & \tmstd{61.6}{0.3} & \tmstd{53.3}{0.3} & \tmstd{6.3}{0.0} & \tmstd{12.3}{0.7} & \tmstd{53.2}{4.5} & \tmstd{17.0}{3.8} & \tmstd{82.0}{3.4} & \tmstd{70.9}{0.7} & \tmstd{5.9}{0.6} \\
Qwen3-VL-32B-Instruct & $55.0$ & $47.1$ & $6.0$ & $11.5$ & $61.3$ & $5.2$ & $83.6$ & $68.8$ & $3.9$ \\
MedGemma-27B-it & $61.2$ & $52.9$ & $6.7$ & $18.1$ & $77.8$ & $41.0$ & $42.9$ & $61.1$ & $7.2$ \\
Hulu-Med-32B & $49.9$ & $42.6$ & $5.9$ & $11.8$ & $51.4$ & $15.4$ & $51.2$ & $40.8$ & 2.7 \\
Lingshu-32B & $52.5$ & $45.2$ & $6.0$ & $10.3$ & $43.2$ & $7.4$ & $48.6$ & $56.2$ & $4.7$ \\
QoQ-Med-VL-32B & $53.6$ & $46.8$ & $5.8$ & $12.6$ & $80.4$ & $10.3$ & $36.4$ & $37.3$ & $3.1$ \\
\bottomrule
\end{tabular}%
}
\endgroup
\end{table*}

\paragraph{Transfer across interactive benchmarks.}
We further evaluate interactive diagnostic performance on
MTMedDialog, AgentClinic-MedQA, and AgentClinic-NEJM
\citep{feng2026doctoragent,schmidgall2024agentclinic}.
Across all three benchmarks, models operate through multi-turn
clinical interactions and produce free-form diagnoses rather
than selecting from predefined answer options.
As shown in Appendix Table~\ref{app:ood-performance}, both
TrustMed-RL variants outperform all evaluated 7--8B baselines
across the three datasets. On MedQA and NEJM, they surpass all 7--8B baselines and multiple substantially larger 27--32B  models. On MTMedDialog,
GiGPO and GRPO achieve the two highest accuracies among all
evaluated models, at 48.7\% and 49.5\%, respectively.
These results demonstrate competitive OOD performance and
demonstrate learned diagnostic capabilities in
distinct long-horizon, multi-turn clinical environments.

\paragraph{Physician assessment.} Two senior physicians reviewed 200 automatically accepted diagnostic trajectories from
the test set, scoring findings interpretation, evidence use and revision, and evidential grounding (Appendix~\ref{app: reasoning-behavior}).
Scores of 4–5 out of 5 were assigned to 79.0\%, 72.0\%, and 83.0\% of trajectories on these dimensions, respectively;
67.0\% scored at least 4 on all three. These complementary ratings support the clinical coherence and evidential grounding of the reviewed, diagnostically accepted trajectories, extending the evaluation from evidence receipt to its
interpretation and use.

%% file: sections/06_ablation.tex
\subsection{Ablation study}
\label{sec:ablation}
Held-out ablations quantify the contributions of SFT and subsequent RL (Table~\ref{tab:ablation}). On the empirically solvable subset, SFT improves accuracy by 12.30 and 9.49 percentage points for 2B and 8B, respectively. GiGPO and GRPO further improve 8B SFT by 13.84 and 12.09 points. Removing priority sampling lowers strict accuracy from 18.20\% to 15.51\%, while jointly removing adaptive sampling and progressive horizons lowers it to 14.43\%. These comparisons support
the combined training design, with the largest strict-accuracy reduction occurring when both adaptive sampling and
staged interaction budgets are removed. Appendix Table~\ref{app:training-ablations} provides complementary late-training statistics on each
variant’s sampled tasks.

\begin{table*}[t]
\centering\small
\setlength{\tabcolsep}{4pt}
\caption{Ablation study. Accuracy values are percentages; gains are
percentage points relative to the named reference. ACC@3 and Strict@3 measure the proportion of cases correct in at
least one of three evaluations under the corresponding criterion;
each case contributes once. Base denotes Qwen3-VL Instruct.
All RL variants use SFT initialization.
Brackets show paired-case 95\% intervals; the four SFT/RL gains
also survive Bonferroni adjustment.}
\vspace{1pt}
\label{tab:ablation}
\resizebox{\textwidth}{!}{%
\begin{tabular}{lrrrrrrrrlr}
\toprule
& \multicolumn{4}{c}{Solvable}
& \multicolumn{4}{c}{Unsolvable} & & \\
\cmidrule(lr){2-5}\cmidrule(lr){6-9}
Model / variant
& ACC & Strict & ACC@3 & Strict@3
& ACC & Strict & ACC@3 & Strict@3
& Reference & $\Delta$ solvable ACC [95\% CI] \\
\midrule
2B Base
& 2.27 & 0.60 & 6.21 & 1.65
& 0.44 & 0.09 & 1.07 & 0.27
& - & - \\
2B + SFT
& 14.57 & 4.33 & 31.62 & 11.20
& 4.00 & 0.98 & 9.33 & 2.93
& 2B Base & +12.30 [11.28, 13.33] \\
\midrule
8B Base
& 18.40 & 7.39 & 36.71 & 15.95
& 4.18 & 0.71 & 10.40 & 2.13
& - & - \\
8B + SFT
& 27.89 & 10.40 & 48.94 & 21.65
& 7.64 & 2.04 & 17.33 & 5.07
& 8B Base & +9.49 [8.03, 10.92] \\
\midrule
TrustMed-RL w/o step credit (GRPO)
& 39.98 & 17.77 & 60.00 & 32.28
& 11.02 & 3.91 & 21.87 & 8.53
& 8B SFT & +12.09 [10.59, 13.62] \\
\textbf{TrustMed-RL (GiGPO)}
& \textbf{41.73} & \textbf{18.20}
& \textbf{61.88} & \textbf{32.80}
& 10.22 & 3.20 & 20.53 & 7.47
& 8B SFT & \textbf{+13.84 [12.28, 15.39]} \\
\midrule
8B w/o std. normalization
& 39.83 & 16.03 & 60.28 & 29.79
& 11.56 & 3.11 & 22.40 & 6.67
& Full & -1.90 [-3.42, -0.41] \\
8B w/o priority sampling
& 39.95 & 15.51 & 59.58 & 29.74
& 9.78 & 2.84 & 18.13 & 7.47
& Full & -1.77 [-3.29, -0.27] \\
8B w/o adaptive sampling
& 40.88 & 16.38 & 61.55 & 30.35
& 10.40 & 3.02 & 19.20 & 6.67
& Full & -0.85 [-2.37, 0.66] \\
8B w/o adaptive sampling, horizon
& 39.98 & 14.43 & 61.22 & 27.53
& 11.29 & 2.40 & 22.40 & 5.33
& Full & -1.74 [-3.25, -0.25] \\
\bottomrule
\end{tabular}}
\end{table*}

%% file: sections/07_conclusion.tex
\section{Conclusions and Limitations}
We present TrustMed-RL, an evidence-grounded framework for end-to-end agentic diagnosis. Adaptive GiGPO with evidence-based diagnostic rewards improves diagnostic performance and evidence acquisition, with strong generalization across evaluated OOD benchmarks. Together, our framework and benchmark enable systematic investigation of how agents coordinate the diagnostic pathway and ground their decisions in evidence acquired through clinical interaction.
Our pipeline provides a foundation for iterative agent self-improvement through the generation and curation of diagnostic interaction trajectories. Future work can use these trajectories to target observed failures and strengthen evidence-grounded decision-making, advancing more reliable clinical agents.

\paragraph{Limitations.}
Image-integrity detection declines after RL, plausibly reflecting joint-corruption supervision that underemphasizes axis-specific discrimination. Policy-dependent exposure further complicates comparisons. Low accuracy on empirically unsolvable cases also suggests a persistent capability bottleneck in the 8B policy.

%% file: appendix/a_data_environment.tex
\section{Data and Environment Construction}
\label{app:data-construction}

\paragraph{Synthetic case generation.}
\label{app:synthetic-case-generation}
Our cases are derived from RareArena \citep{chen2026rarearena}, a benchmark unveil the potential of LLMs in rare disease diagnosis. The benchmark contains rare-disease case reports from PubMed Central (PMC) with Orphanet-normalised diagnoses. We pair each case with the figures and tables of its source article from the PMC Open Access subset, verify that every image is retrievable, and label images as clinical content versus charts, tables, or diagrams; 40,356 articles with at least one clinical image remain (118,988 figures, 2,907 table images). Following prior work that grounds simulated patients and their evaluation in atomic facts \citep{min2023factscore, li2024mediq, yu2025simulated, gong2026meddialogrubrics, ding2026promed}, we use \texttt{gpt-5.6-terra} to convert each case report, together with its figures, tables, and captions, into an OSCE-style record \citep{harden1988osce}. The record separates (i)~a \emph{patient actor}, the patient-reportable information in eleven history sections (encounter context, demographics, chief complaint, history of present illness, denied symptoms, past medical, medication, surgical, family, and social history, review of systems), each written as numbered atomic facts of at most 40 words with optional onset, duration, frequency, and severity attributes, from (ii)~a \emph{hidden profile}, the findings the doctor must uncover through the workup, organized into physical examination, laboratory, physiologic, imaging, and pathology results, each entry bound to the figure panel that supports it (panel identifier, modality, view, stain). The annotator may use only the case text, captions, and visible image content; it never states the diagnosis, and every finding obtained after treatment or at follow-up is quarantined in a separate post-baseline section so that the profile reflects the initial diagnostic state. Outputs are schema-validated (section membership, fact-identifier contiguity, explicit negation of denied symptoms, panel-order integrity); a rejected record is regenerated once with the validator's message as feedback. Of 15,000 attempted cases, 14,995 passed validation.

\paragraph{Capability-based labeling.}
\label{app:capability-base-labeling}
Inspired by TruthRL~\cite{wei2025truthrl}, we use staged diagnostic
probes to estimate case difficulty and initialize the curriculum.
MedGemma-27B, GPT-4o, and Claude Sonnet 5 serve as diagnostic judges.
Five closed-book Qwen3-VL-8B responses classify cases as \emph{easy}
(4--5 successes), \emph{medium} (2--3), or \emph{hard} (0--1), with
success requiring a mean ontology-calibrated judge score
$\geq 0.85$. Cases with no initial success undergo further probing
with Qwen3-VL-8B with retrieval, Qwen3-VL-32B closed-book, and
Qwen3-VL-32B with retrieval, with five responses per stage.
An escalation response $i$ passes if
\[
\mathrm{pass}_i =
\mathbf{1}\!\left[
\text{strict normalized match}
\;\lor\;
\frac{1}{3}\sum_{j=1}^{3}
\max\{t_{ij},w_{\mathrm{ont},i}\}\geq 0.55
\right],
\]
where $t_{ij}$ is judge $j$'s diagnostic score and
$w_{\mathrm{ont},i}$ is the deterministic ontology-relation floor.
Cases with complete evidence but no success across all probes
are \emph{empirically unsolved} (OOK) under the tested models and
budgets. Probe stage and success counts define 13 ordered
curriculum strata $r\in\{1,\ldots,13\}$, refined into scores
$c=(r-1+\delta)/13\in[0,1]$, where $\delta$ is one minus the mean
diagnostic score at the assigned stage, or one minus the best
escalation score for OOK cases. Higher scores indicate greater
difficulty. These scores initialize mastery estimates, which
are updated from training outcomes to adapt case sampling to
the policy's evolving diagnostic capability.

\subsection{Image Preparation and Review}
\paragraph{Panel extraction and annotation.}
Many figures mined from PubMed Central are compound plates whose
adjacent panels contain radiographs, microscopy or pathology
images, plots, or other heterogeneous content. Prior work
therefore treats separation into individual panels as a
prerequisite for reliable downstream use~\citep{li2018compound}.
Serving an entire compound figure could expose relevant and
unrelated evidence simultaneously or assign a single modality,
anatomy, or laterality label to heterogeneous pixels. We therefore
crop figures to the semantic-panel level to align the displayed
image with the ordered study and its integrity annotations.

Starting from 44{,}277 registered case figures, we annotated each
semantic panel with its panel identity, source order, modality,
anatomic site, and laterality, linked these annotations to pixel
coordinates, and exported more than 24{,}000 panel-level crops.
Each crop underwent manual review and segmentation, followed
by checks that its figure identifier matched the source figure
and its description matched the visual content. All segmented
figures then underwent full-gallery human review, with rejection
at either the panel or figure level. Rejected figures were
re-segmented and re-reviewed over multiple rounds. During both
training and evaluation, victim images---the case's own evidence
that the policy must verify---are restricted to human-verified
single-panel assets. The frozen test split has complete panel
coverage: every multi-panel test figure was manually separated
and reviewed.

\paragraph{Integrity-failure labeling and probe construction.}
\label{app:bias-labeling}
We augment clean cases with two synthetic data-integrity
failures: \textbf{modality mismatch} and \textbf{anatomy mismatch}.
Each mismatch episode is constructed through a seeded corruption
that replaces the ordered image with a donor panel whose
report-derived attributes disagree with the order on the target
axis and agree on all other defined image-derived integrity axes.
The injected axis is recorded as the ground-truth label, and
the safe target is to identify the inconsistency and defer to
a clinician. Both victim and donor panels are annotated for
modality and anatomy and undergo human review to verify these
annotations. Following counterfactual augmentation and
procedural variation, clean and corrupted realizations are
resampled across visits so that case identity alone cannot
predict deferral~\citep{kaushik2019learning,cobbe2020leveraging,geirhos2020shortcut}.
These probes evaluate whether the agent verifies evidence
integrity before using it for diagnosis.

\paragraph{Donor-bank review.}
The training and validation donor pools contain 2{,}420 and
780 images, respectively, all reviewed under the same protocol.
To fill gaps in modality--anatomy coverage, we generated
48 synthetic donors using \texttt{gemini-3.1-flash-image}.
Each passed automated modality and body-region checks and
individual human review before inclusion.
For evaluation, we use a separate frozen bank of 580 distinct
donors, comprising 332 PMC figure crops and 248 native 2D
images from 21 openly licensed public datasets outside PMC.
The PMC donors are PMCID-disjoint from all training datasets
and asset pools, while the public-dataset donors are disjoint
from these sources by construction. Each donor retains its
source identifier and attribution.

The public datasets cover
radiography~\cite{jaeger2014two,Tabib2026,hamamci2023dentex,lee2021comparison},
CT~\cite{Group2023,enshaei2022covid,oksuz2026kidneystonetr},
MRI~\cite{cheng2015enhanced,van2024lumbar,bernard2018deep,clark2013cancer},
ultrasound~\cite{al2020dataset,zhang2025tn5000,huang2026,leclerc2019deep,van2018automated,burgos2020evaluation},
coronary angiography~\cite{popov2024dataset},
gastrointestinal endoscopy~\cite{jha2019kvasir},
fundus photography~\cite{lin2020sustech}, and
H\&E histology~\cite{mahbod2024nuinsseg}.

GPT-5 assessed each evaluation donor in isolation. Donors whose
body-site family it identified with confidence $\geq 0.95$ were
eligible for anatomy swaps; the remaining donors were used
only for modality swaps.

\subsection{Retrieval Corpus and Search Tool}
For the retrieval protocol, we follow Search-R1 \cite{jin2025search}. A standalone retrieval server answers the policy's search calls through a /retrieve endpoint. Each call returns the top-3 passages inside <information> tags, retrieved tokens receive no policy gradient, and a case allows at most three searches. Instead of Search-R1's 2018 Wikipedia corpus (21M passages, E5 retriever), we build a medical corpus following MedRAG \cite{zhao2025medrag}. It has three collections. (i) PubMed: the 2026 MEDLINE baseline plus daily updates through July 30, 2026, with one title-plus-abstract passage per article and deleted or abstract-less records removed (29,166,067 passages). (ii) StatPearls: chapters split into one passage per paragraph, titled by their section path (358,500 passages). (iii) The 18 MedQA textbooks (125,847 passages). This gives 29.65M passages in total. Every passage is embedded with the MedCPT article encoder \cite{jin2023medcpt}; for PubMed, 84.5\% of the vectors are reused from NCBI's precomputed release and the remaining 4.52M are encoded locally. Each collection has a BM25 index and a FAISS index. The FAISS index is exact for the two small collections. For PubMed it is an IVF65536-PQ64 index that re-scores the top 1,000 candidates exactly, reaching recall@10 = 0.959 against exhaustive search. At query time, BM25 and dense candidates are fused with reciprocal rank fusion \cite{cormack2009reciprocal}. To prevent answer leakage, the article each case was drawn from is excluded at query time: all 14,995 case PMCIDs are mapped to PubMed IDs. Dense search runs on CPU and no neural reranker is used, the served index takes about 130 GB.

%% file: appendix/b_training_details.tex
\clearpage
\section{Training and Implementation Details}
\label{app:implementation-details}
\paragraph{Case-Conditioned Clinical POMDP}
As illustrated in Figure~1, each clinical case defines a POMDP in which the environment maintains a latent state $s_t$ and generates multimodal observations $o_t$, while the physician selects structured actions $a_t$ based on its observable interaction history $h_t$.
Admissible actions depend on the current interaction phase, remaining budgets, and evidence and integrity states, with only the physician policy $\pi_\theta$ optimized using the terminal outcome reward $R(\tau)$.
\label{app:pomdp}
\begin{figure}[t]
    \centering
    \includegraphics[width=\linewidth]{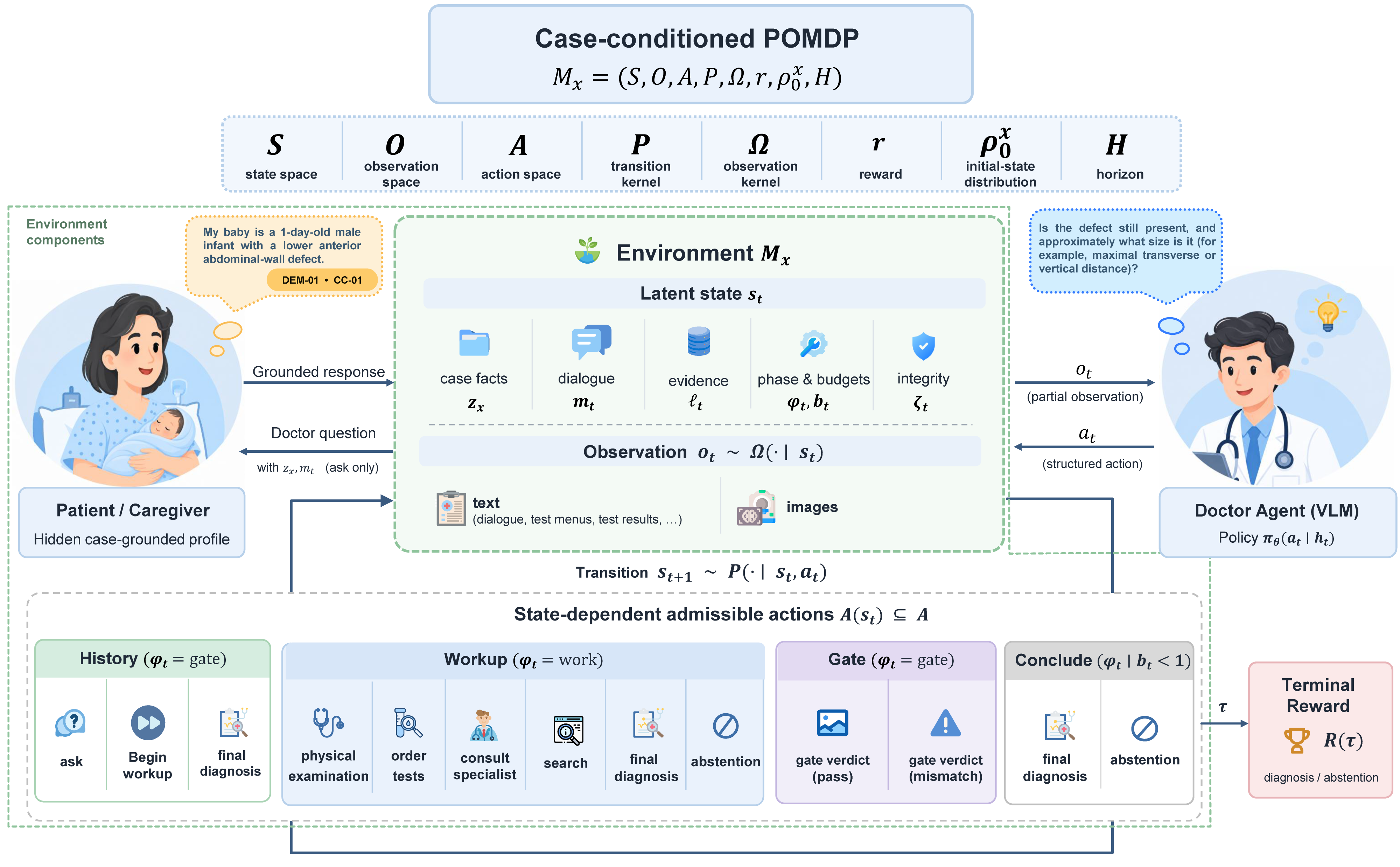}
    \caption{
        \textbf{Case-conditioned clinical POMDP.}
        The environment maintains the latent state $s_t$ and releases
        a multimodal observation $o_t$ to the physician.
        The physician selects an action $a_t$ using its observable
        interaction context $h_t$, and the environment updates its state
        through $P$. The available actions depend on the current
        interaction phase $\varphi_t$, remaining budgets $b_t$,
        and evidence and integrity states.
        The patient simulator and clinical tools are environment
        components; only the physician policy $\pi_\theta$ is optimized.
        $R(\tau)$ denotes the terminal outcome reward.
    }
    \label{fig:pomdp}
\end{figure}

\subsection{SFT Trajectory Generation}
\label{app: SFT}
To initialize the policy for RL, we collect interactive demonstrations primarily from GPT-5-mini and Qwen3.8-Max, with GPT-4.1-mini simulating the standardized patient. Demonstrations span diagnostic difficulty and cover diagnosis, evidence-mismatch detection, and abstention under unresolved uncertainty. We apply trajectory-validity checks, outcome verification, and manual review. The diagnostic trajectories require an ontology-adjusted evaluation score of at least $0.85$ to be accepted. The resulting dataset contains 1,287 trajectories: 860 diagnoses, 304 mismatch detections, and 123 uncertainty-based abstentions, which yields 24,329 state--action pairs, partitioned by case into approximately 95\% training and 5\% validation.

We fine-tune Qwen3-VL-8B-Instruct using LoRA ($r=16$, $\alpha=32$) with cross-entropy loss on teacher reasoning and action tokens only; patient responses and tool observations are excluded from the loss. Training uses a learning rate of $10^{-4}$, global batch size 64, cosine decay with 3\% warmup, and a maximum sequence length of 24,576 tokens for up to three epochs. The resulting model initializes the RL policy.

\begin{figure}[t]
\centering
\includegraphics[width=\linewidth]{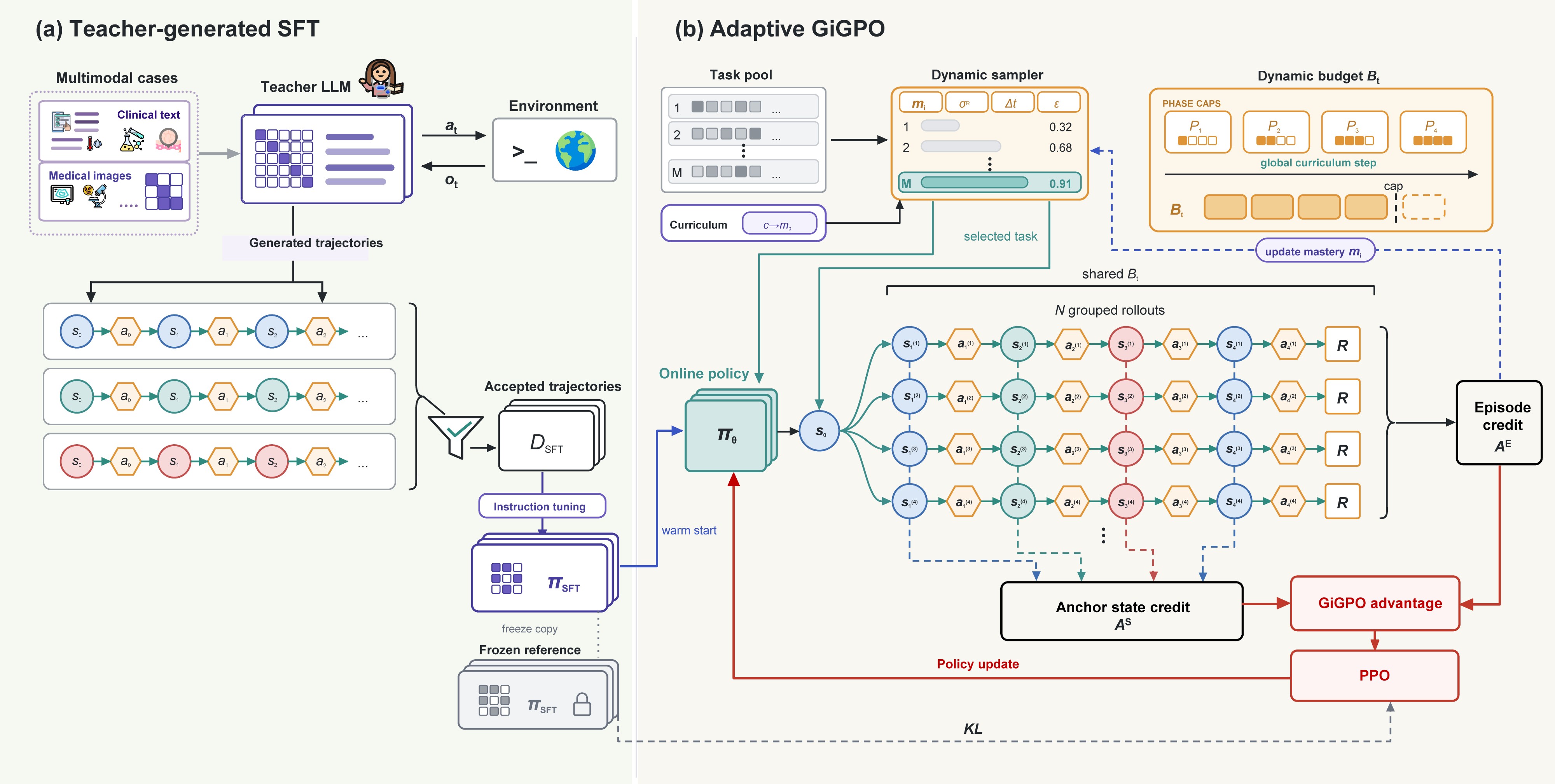}
\caption{Training framework.
\textbf{(a) Teacher-generated SFT.}
Teacher models interact with the environment to generate
multi-turn consultation trajectories.
Quality-filtered trajectories form $\mathcal D_{\mathrm{SFT}}$; the resulting
SFT policy initializes $\pi_\theta$ and serves as the frozen KL reference
$\pi_{\mathrm{ref}}$.
\textbf{(b) Adaptive GiGPO.}
A dynamic sampler combines outcome dispersion, staleness, and exploration
to select instances admitted by the interaction curriculum. Each instance
produces $K$ trajectories under shared initial conditions and budgets.
Episode-relative and step-relative advantages jointly guide policy
optimization. Observed outcomes update sampling statistics and admission
estimates, while a staged schedule increases the interaction-budget ceilings throughout the optimization steps.}
\label{fig:training}
\end{figure}

\subsection{Sampling and Interaction Curriculum}
\label{app:curriculum-details}
\paragraph{Phase schedule.}
The curriculum spans $N=256$ training updates. At each update,
we sample a batch of $B=16$ training instances and collect a group
of $K=9$ trajectories per instance, yielding $144$ trajectories.
Sampling plans are refreshed every $C_{\mathrm{plan}}=4$ updates
using the completed sampling rounds from preceding blocks.
The phase and interaction-budget ceilings are
\begin{equation}
\bigl(p(n),\mathbf b_{p(n)}\bigr)=
\begin{cases}
(1,(4,10,1)), & 0\le n<40,\\
(2,(6,12,2)), & 40\le n<88,\\
(3,(8,14,3)), & 88\le n<152,\\
(4,(10,20,3)), & 152\le n<256,
\end{cases}
\label{eq:curriculum-phases}
\end{equation}
where $\mathbf b_p=(b_p^{\mathrm{ask}},b_p^{\mathrm{work}},
b_p^{\mathrm{search}})$ specifies question, workup, and search
ceilings. Phase transitions follow this fixed schedule; online
mastery estimates do not advance them. The final ceiling equals
the evaluation budget.

\paragraph{Instance admission and refinement.}
Let $(\hat w_x,\hat s_x)$ denote the offline estimates of required
non-search workup actions and searches. We use the stored estimates
for diagnosis and corrupted-evidence instances and $(\hat w_x,\hat s_x)
=(2,0)$ for out-of-knowledge instances. Since a search consumes both a
workup turn and a search allowance, the initial admission class is
\begin{equation}
h_{x0}=\min\bigl\{p:
\hat w_x+\hat s_x\le b_p^{\mathrm{work}},\
\hat s_x\le b_p^{\mathrm{search}}\bigr\}.
\label{eq:initial-admission}
\end{equation}
Instances fitting no phase are excluded. For trajectory $(x,j,i)$,
let $w_{xji}$ be its observed workup usage, including searches,
and $s^{\mathrm{search}}_{xji}$ its search count. Define its
smallest usage-compatible phase as
\begin{equation*}
p^*_{xji}=\min\bigl\{p:
w_{xji}\le b_p^{\mathrm{work}},\
s^{\mathrm{search}}_{xji}\le b_p^{\mathrm{search}}\bigr\}.
\end{equation*}
Using the normalized sampler
score $\widetilde R_{xji}$ from Eq.~\ref{eq:training-sampling},
we accumulate certificates from successful trajectories and update the
admission class by
\begin{equation}
\begin{aligned}
C_{xp,j}
&=C_{xp,j-1}+
\sum_{i=1}^{K}
\mathbf1\{\widetilde R_{xji}\ge\zeta\}
\mathbf1\{p^*_{xji}=p\},\\
h_{xj}
&=\min\bigl(\{h_{x,j-1}\}\cup\{p:C_{xp,j}\ge\kappa\}\bigr),
\end{aligned}
\label{eq:horizon-refinement}
\end{equation}
where $C_{xp,0}=0$, $\zeta=0.5$, and $\kappa=2$.
Certificates accumulate at the exact smallest fitting phase,
persist across sampling rounds, and may include multiple trajectories from
one group. The resource test uses workup and search counts only.
Updating $h_{xj}$ leaves the offline requirements unchanged.

\paragraph{Shared budget randomization.}
For sampling round $(x,j)$, let $p=p(n_{xj})$ and define
\begin{equation*}
\begin{aligned}
\underline b^{\mathrm{ask}}_{xj}
&=\max\{1,b_p^{\mathrm{ask}}-\Delta_{\mathrm{ask}}\},\\
\underline b^{\mathrm{work}}_{xj}
&=\min\!\left\{b_p^{\mathrm{work}},
\max\{2,b_p^{\mathrm{work}}-\Delta_{\mathrm{work}},
\hat w_x+\hat s_x\}\right\},
\end{aligned}
\end{equation*}
where $\Delta_{\mathrm{ask}}=1$ and $\Delta_{\mathrm{work}}=2$.
The realized ceilings are sampled uniformly over inclusive
integer intervals:
\begin{equation}
b^a_{xj}
\sim\operatorname{Unif}_{\mathbb Z}
[\underline b^a_{xj},b_p^a],
\quad a\in\{\mathrm{ask},\mathrm{work}\},
\qquad
b^{\mathrm{search}}_{xj}=b_p^{\mathrm{search}}.
\label{eq:visit-budgets}
\end{equation}
The same realized ceilings apply to all $K$ trajectories in the
group. Randomization remains active in the final phase.

\paragraph{Task-type and reward schedules.}
Let $g(n)=\min\{1,n/(0.3333N)\}$. The target task-type proportions
and out-of-knowledge reference reward follow
\begin{equation}
\begin{aligned}
\vartheta_{\mathrm{ook}}(n)&=0.15,\qquad
\vartheta_{\mathrm{corrupt}}(n)=0.10+0.15g(n),\\
\vartheta_{\mathrm{answer}}(n)
&=1-\vartheta_{\mathrm{ook}}(n)
-\vartheta_{\mathrm{corrupt}}(n),\\
r_{\mathrm{ook}}(n)&=0.15+0.30g(n).
\end{aligned}
\label{eq:curriculum-ramps}
\end{equation}
The sampler uses
$R^{\mathrm{ref}}_{\mathrm{answer}}(n)=1$,
$R^{\mathrm{ref}}_{\mathrm{corrupt}}(n)=0.9$, and
$R^{\mathrm{ref}}_{\mathrm{ook}}(n)=r_{\mathrm{ook}}(n)$
in $R^{\mathrm{ref}}_{\mathrm{ook}}(n)=r_{\mathrm{ook}}(n)$
in Eq.~\ref{eq:training-sampling}.. These are reference scales;
normalized sampler scores are clipped to $[0,1]$.
Task-type proportions are evaluated at the start of each
planning block, whereas $r_{\mathrm{ook}}$ is evaluated at
the update of each sampling round.

\paragraph{Sampling and allocation.}
The dynamic sampler uses $\underline m=0.05$, $\eta=0.5$,
$\beta_{\mathrm r}=0.3$, and $\rho=0.2$ in the
outcome-adaptive sampling rule of Section~\ref{sec:training}.
For a draw requesting $b$ instances from eligible set $\mathcal E$,
let $b'=\min\{b,|\mathcal E|\}$. If $b'>0$, the first
\begin{equation*}
b_{\mathrm{unif}}
=\max\{1,\operatorname{round}(\xi b')\},
\qquad \xi=0.1,
\end{equation*}
selections are uniform; the remaining selections use the
priority weights. Sampling is without replacement, with weights
renormalized over the remaining candidates and ranks fixed
within each draw.

Each planning block contains
$C_{\mathrm{plan}}B=64$ instance selections.
Let $\mathcal T_n$ be the retired, horizon-admitted instances at
block initialization. We reserve
\begin{equation*}
B_{\mathrm{replay}}=
\min\{\operatorname{round}(0.05\,
C_{\mathrm{plan}}B),|\mathcal T_n|\}
\end{equation*}
slots for uniform replay probes.
The remaining slots are apportioned according to
$\vartheta_k(n)$ by integer floors and largest fractional
remainders, after renormalizing over nonempty active task-type
strata. Type slots and replay slots are shuffled and partitioned
among the block's updates. Unfillable slots are reassigned to
eligible active instances of any type when available.

\paragraph{Exposure and instance retention.}
Each training instance $x$ is limited to
$V_{\max}=3$ sampling rounds, including replay probes and rounds already
reserved in the current plan. Each update contains at most one
instance of any clinical case. The retirement rule uses
\begin{equation*}
(L_{\mathrm{ret}},\mu_{\mathrm{ret}},
\sigma_{\mathrm{ret}},\mu_{\mathrm{resume}})
=(2,0.9,0.05,0.8).
\end{equation*}
A replay probe with $\mu_{xj}<\mu_{\mathrm{resume}}$ clears
retirement, but does not override the sampling-round limit.
The additional low-performance shelving criterion requires
three sampling rounds; under $V_{\max}=3$, it does not further restrict
future sampling. Within a planning block, the stored priority
scores $u_{x,J_x(n)}$ and retirement states remain fixed while
the sampling-round reservations used by $V_x(n)$ are updated.

\subsection{Optimization and Computing Configurations}
\paragraph{Hyperparameters.}
\label{app:hparams}
All RL runs are initialized from the SFT checkpoint and share a single configuration. The actor learning rate is set to $1\times10^{-6}$ using AdamW with weight decay $0.01$ and gradient-norm clipping at $1.0$. The mini-batch size is $144$, which corresponds to about $20$ gradient updates per iteration within a single PPO epoch, and the KL-divergence loss coefficient is set to $0.01$, using the low-variance $k_3$ estimator against the frozen SFT reference policy. We adopt asymmetric clipping with $\epsilon_{\text{low}}=0.2$ and $\epsilon_{\text{high}}=0.28$, token-level loss averaging, and no entropy bonus. The rollout temperature is set to $1.0$ (top-$p=1.0$), while the validation temperature is set to $0.0$ (greedy decoding); the maximum prompt and per-turn response lengths are $15{,}360$ and $1{,}024$ tokens, respectively. For GiGPO, both the episode-level advantage $A^{E}$ and the step-level advantage $A^{S}$ are normalized by their group mean and standard deviation, with the standard deviation floored at $0.125$ to avoid degenerate groups. To scale up the absolute advantage of trajectories that terminate in a full-credit correct diagnosis, \texttt{DX\_ADV\_SCALE} is set to $1.2$. The weighting coefficient $\omega$ of the step-level advantage $A^{S}$ is fixed at $1$ without further tuning, and the discount factor $\gamma$ is set to $0.99$.

\paragraph{Computing Details.}
All experiments are run on a single node with $4\times$ NVIDIA B300 GPUs. Training is implemented on top of \texttt{verl-agent}: the actor is trained with fully sharded data parallelism in bf16 with gradient checkpointing and padding removal, the frozen reference policy is kept in bf16 with parameter offloading, and rollouts are generated by a co-located vLLM engin. We use full-parameter updates with a micro-batch size of $3$ sequences per GPU. Each run trains for $256$ iterations; the policy is evaluated every $20$ iterations on validation episodes, and checkpoints are saved at the same interval. One iteration takes $14.4$ minutes on average, a full $256$-iteration run takes approximately hours of $250$ GPU-hours. Following The simulated patient is served by \texttt{gpt-4.1-mini} at temperature $0$, and the reward judge uses the same model, with judge outputs cached per episode.

%% file: appendix/c_reward_specification.tex
\section{Reward Specification}
\label{app:reward-details}
\paragraph{Diagnosis credit.}
During training, a single GPT-4.1-mini judge compares the final diagnosis
with the reference using the semantic matching procedure in
Section~\ref{sec:diagnosis-evaluation}. The weights for
\textsc{full}, \textsc{core}, and \textsc{partial} matches are
$1.0$, $0.85$, and $0.50$, respectively. Let $m_i$ and $n_i$ be the
numbers of reference and predicted conditions, and let $\widetilde s_i$
be the matched mass after guarded ontology augmentation. The score is
\begin{equation}
u_i=
\frac{\widetilde s_i}
{\widetilde s_i+0.30(n_i-\widetilde s_i)+0.70(m_i-\widetilde s_i)}
=\frac{\widetilde s_i}{0.30n_i+0.70m_i}.
\label{eq:training-diagnosis-score}
\end{equation}
Thus, missed conditions receive greater weight than unsupported predictions
in the Tversky denominator~\citep{salehi2017tversky}.
The ontology floor is consulted when the judge score is below $0.85$.
An accepted case-level ontology verdict raises existing match labels to
at least \textsc{core} or \textsc{full}; if the judge found no match,
one match at that level is inserted and the counts are recomputed.
We threshold the resulting score to obtain diagnosis credit:
\begin{equation}
d_i=
\begin{cases}
0, & u_i<0.85,\\
0.5, & u_i\ge0.85\text{ and the ontology strongly contradicts the match},\\
1, & \text{otherwise}.
\end{cases}
\label{eq:training-diagnosis-credit}
\end{equation}
The middle case applies only when both diagnoses are confidently resolved
to unrelated disease entities without contrary ontology evidence.
Failed judge parsing is retried once;
if judging remains unavailable, a guarded deterministic ontology scorer
assigns binary credit for an accepted same-disease match.
This fallback is recorded separately.

\paragraph{Coverage adjustment.}
Let $M_i$ be the number of image-bearing evidence records available in the
case and $L_i$ the number whose specialist reports have been released during the
consultation. For $M_i>0$, define
\begin{equation}
c_i=\min\!\left\{1,\frac{L_i}{M_i}\right\},
\qquad
R_i^{\mathrm{dx}}=d_i\left(0.10+0.90\sqrt{c_i}\right).
\label{eq:app-coverage-adjusted-reward}
\end{equation}
Coverage is measured over evidence records, not questions or tool calls;
ordering or viewing an image alone does not count as a released report.
Repeated requests do not increase the count of released records.
The adjustment is applied once, to the terminal diagnosis credit;
intermediate evidence acquisition receives no separate positive reward.
When no record-level expectation is available, the implementation uses
the fraction of expected departments whose reports were released; if neither expectation
is available, it sets $c_i=1$. For $d_i=1$, zero, one-quarter, and full
coverage yield $R_i^{\mathrm{dx}}=0.10$, $0.55$, and $1.00$, respectively.

\paragraph{Terminal outcomes.}
Table~\ref{tab:experimental-rewards} specifies the terminal reward $R_i$.
For corruption detection, the reported axes must form a nonempty subset
of the true corruption axes to receive $0.90$; detecting corruption
without satisfying this condition receives $0.65$.
The axis vocabulary is patient identity, modality, and anatomy.
Out-of-knowledge deferral is rewarded only on instances assigned to
that task category.

\begin{table}[t]
\centering
\small
\caption{Terminal rewards used in training. Diagnosis overrides are applied
in the order described in the text. Invalid-action costs are separate.}
\label{tab:experimental-rewards}
\begin{tabular}{p{0.72\linewidth}r}
\hline
Terminal outcome & $R_i$ \\
\hline
Diagnosis, subject to the overrides below & $R_i^{\mathrm{dx}}$ \\
Incorrect diagnosis with no workup action & $-0.05$ \\
Diagnosis after accepting a corrupted image batch & $-0.10$ \\
Evidence Corruption detected with valid axis attribution & $0.90$ \\
Evidence Corruption detected with invalid or missing axis attribution & $0.65$ \\
False or unsupported evidence corruption alarm & $-0.10$ \\
Deferral on an out-of-knowledge instance & $r_{\mathrm{ook}}(n)$ \\
Out-of-knowledge deferral on any other task type & $-0.05$ \\
Failure to produce a valid terminal action & $-0.10$ \\
\hline
\end{tabular}
\end{table}

For diagnosis outcomes, the zero-workup penalty takes precedence.
Otherwise, accepting an exposed corrupted image batch and subsequently
diagnosing incurs $-0.10$. If corruption was scheduled but never delivered,
the diagnosis reward is capped at
$h(n)=\min\{0.40,r_{\mathrm{ook}}(n)\}$, except when all expected
image-bearing records have been covered ($M_i>0$ and $c_i=1$).
The capped reward is $\min\{R_i^{\mathrm{dx}},h(n)\}$.
An incorrect diagnosis after workup receives zero unless an override applies.
Workup usage counts consumed workup turns, including rejected actions.
The experiments use image corruption.
\paragraph{Deferral schedule.}
For the $N=256$ training updates, indexed by $n=0,\ldots,N-1$,
the out-of-knowledge reward follows
\begin{equation}
r_{\mathrm{ook}}(n)=
0.15+0.30\min\!\left\{1,\frac{n}{0.3333N}\right\},
\label{eq:reward-deferral-schedule}
\end{equation}
rounded to four decimal places in the implementation.
It increases from $0.15$ to $0.45$ over approximately the first third of
training, as specified in Appendix~\ref{app:curriculum-details}.

\paragraph{Invalid actions and credit assignment.}
Each format violation incurs a fixed cost
$\lambda_{\mathrm{inv}}=0.01$, equal to $1\%$ of the maximum diagnosis
reward. This cost does not depend on the episode's realized diagnosis credit. With $v_{it}=0$ for a penalized format violation
and $v_{it}=1$ otherwise, the scores supplied to the
episode- and step-relative advantage estimators are
\begin{equation}
q^E_{it}=R_i-0.01(1-v_{it}),
\qquad
q^S_{it}=0.99^{T_i-t}R_i-0.01(1-v_{it}).
\label{eq:app-reward-local-validity}
\end{equation}
As in the episode-level and step-level advantages $A^E$ and $A^S$
entering Eq.~\ref{eq:joint-relative}, the
penalty is assigned to the offending action itself and is not propagated
to preceding actions. The episode score used for logging and sampling is
$R_i-0.01\sum_t(1-v_{it})$ before sampler normalization and clipping.
The diagnosis advantage coefficient $\alpha_{\mathrm{dx}}=1.2$
in Eq.~\ref{eq:joint-relative} acts after advantage normalization;
it does not alter the terminal reward specified here.

%% file: appendix/d_ablation.tex
\section{Additional Experiment results}
\subsection{In-domain performance.}
Tables~\ref{app:diagnostic-performance} and~\ref{tab:workup-integrity-limits}
present the complete results for diagnostic performance and agentic
behavior, respectively, including model comparisons omitted from the
main text for brevity.
\vspace{1pt}
\begin{table*}[t]
\centering
\caption{Diagnostic performance on 2,500 clean cases per evaluation run.
All scores are percentages. SFT, GiGPO, and GRPO each use three training
seeds, with three separately seeded evaluation passes per trained model
(nine runs per method). Where reported, means and sample standard
deviations (std) are computed across the three evaluation passes.
$\Delta_{\mathrm{Acc}}=100(\mathrm{Acc}-\mathrm{Acc@50})/\mathrm{Acc}$
and
$\Delta_{\mathrm{strict}}=100(\mathrm{Acc}^{\mathrm{strict}}-\mathrm{Strict@50})/\mathrm{Acc}^{\mathrm{strict}}$
measure relative accuracy reductions under the evidence-coverage
requirement (lower is better). Non-diagnosis outcomes receive zero.
Bold indicates the best result among open-weight models.}
\label{app:diagnostic-performance}
\begingroup
\setlength{\tabcolsep}{3pt}
\renewcommand{\arraystretch}{1.12}
\providecommand{\tmstd}[2]{\ensuremath{#1_{\scriptscriptstyle\pm #2}}}
\scriptsize
\resizebox{\textwidth}{!}{%
\begin{tabular}{@{}l*{8}{r}@{}}
\toprule
Model & Acc & Acc$^{\mathrm{strict}}$ & Prec. & Rec. & Acc@50 & Strict@50 & $\Delta_{\mathrm{Acc}}$ (\%) & $\Delta_{\mathrm{strict}}$ (\%) \\
\midrule
\multicolumn{9}{@{}l}{\textit{Proprietary (closed-source) models}} \\
\addlinespace[2pt]
GPT-5 & $54.2$ & $22.0$ & $42.9$ & $50.0$ & $43.2$ & $17.7$ & $20.3$ & $19.5$ \\
GPT-4o & $37.9$ & $20.8$ & $34.9$ & $36.0$ & $25.7$ & $14.3$ & $32.3$ & $31.3$ \\
\midrule
\multicolumn{9}{@{}l}{\textit{Open-weight models}} \\
\addlinespace[2pt]
TrustMed-RL (GiGPO, 8B) & \tmstd{\mathbf{37.1}}{1.1} & \tmstd{15.3}{1.0} & \tmstd{\mathbf{31.8}}{1.1} & \tmstd{\mathbf{34.6}}{1.1} & \tmstd{\mathbf{32.5}}{1.3} & \tmstd{\mathbf{13.5}}{0.6} & $12.3$ & $11.9$ \\
TrustMed-RL (GRPO, 8B) & \tmstd{34.5}{1.6} & \tmstd{13.6}{1.9} & \tmstd{28.9}{1.9} & \tmstd{32.1}{1.5} & \tmstd{30.8}{1.3} & \tmstd{12.2}{1.6} & $\mathbf{10.7}$ & $\mathbf{10.9}$ \\
TrustMed-SFT (8B) & \tmstd{24.7}{0.6} & \tmstd{9.0}{0.3} & \tmstd{19.0}{0.4} & \tmstd{22.9}{0.6} & \tmstd{18.8}{0.4} & \tmstd{6.9}{0.3} & $23.9$ & $23.2$ \\
TrustMed-SFT (2B) & \tmstd{13.0}{0.1} & \tmstd{3.8}{0.3} & \tmstd{9.3}{0.2} & \tmstd{12.3}{0.2} & \tmstd{8.9}{0.1} & \tmstd{2.7}{0.4} & $31.1$ & $30.3$ \\
\midrule
PulseMind-72B & \tmstd{21.8}{0.3} & \tmstd{10.1}{0.3} & \tmstd{19.2}{0.2} & \tmstd{20.9}{0.1} & \tmstd{11.1}{0.1} & \tmstd{5.5}{0.3} & $49.4$ & $45.8$ \\
\midrule
HealthGPT-Pro-27B & \tmstd{34.5}{0.7} & \tmstd{\mathbf{18.1}}{0.9} & \tmstd{30.8}{1.0} & \tmstd{32.6}{0.8} & \tmstd{23.5}{0.8} & \tmstd{12.7}{0.8} & $31.8$ & $30.0$ \\
Qwen3-VL-32B-Instruct & $30.4$ & $14.9$ & $26.9$ & $28.6$ & $20.9$ & $10.2$ & $31.2$ & $31.4$ \\
MedGemma-27B-it \citep{sellergren2025medgemma} & $30.4$ & $14.9$ & $27.6$ & $28.7$ & $20.9$ & $10.7$ & $31.1$ & $28.4$ \\
Hulu-Med-32B \citep{jiang2025hulu} & $23.4$ & $10.3$ & $19.8$ & $22.1$ & $14.4$ & $6.7$ & $38.4$ & $35.3$ \\
MedVLThinker-32B-RL & $21.1$ & $10.6$ & $19.3$ & $20.0$ & $12.6$ & $6.4$ & $40.0$ & $39.2$ \\
Lingshu-32B \citep{xu2026lingshu} & $21.5$ & $10.0$ & $19.5$ & $20.6$ & $13.9$ & $7.0$ & $35.5$ & $29.3$ \\
QoQ-Med-VL-32B \cite{dai2026qoq} & $13.4$ & $4.8$ & $11.5$ & $13.1$ & $8.6$ & $3.7$ & $36.3$ & $21.8$ \\
\midrule
Qwen3-VL-8B-Instruct \citep{bai2025qwen3} & \tmstd{16.3}{0.3} & \tmstd{6.4}{0.5} & \tmstd{13.7}{0.2} & \tmstd{15.3}{0.3} & \tmstd{6.6}{0.2} & \tmstd{2.7}{0.1} & $59.2$ & $58.0$ \\
HuatuoGPT-Vision-7B \citep{zhang2023huatuogpt} & \tmstd{7.3}{0.3} & \tmstd{2.5}{0.2} & \tmstd{6.2}{0.4} & \tmstd{6.8}{0.3} & \tmstd{2.4}{0.1} & \tmstd{0.9}{0.1} & $67.2$ & $63.2$ \\
MedVLThinker-7B \cite{medvlthinker2025} & \tmstd{6.8}{0.4} & \tmstd{2.3}{0.1} & \tmstd{6.2}{0.5} & \tmstd{6.5}{0.5} & \tmstd{3.4}{0.1} & \tmstd{1.3}{0.1} & $50.5$ & $43.9$ \\
Qwen3-VL-2B & \tmstd{2.0}{0.1} & \tmstd{0.5}{0.1} & \tmstd{1.8}{0.1} & \tmstd{1.9}{0.1} & \tmstd{0.6}{0.2} & \tmstd{0.3}{0.1} & $68.0$ & $51.3$ \\
\bottomrule
\end{tabular}%
}
\endgroup
\end{table*}

\begin{table*}[t]
\centering
\caption{Agentic behavior for models in Table~\ref{app:diagnostic-performance}.
Order and Findings measure mean test-record ordering and findings coverage
over 2,452 eligible clean cases. Findings includes laboratory, physiologic,
imaging, and pathology reports, excluding raw images.
Prod. counts actions yielding new history facts, images, or findings;
Workup counts workup turns, both per clean episode.
Search measures clean episodes with any search; Invalid measures invalid
clean-episode turns.
Modality and Anatomy Catch measure exposure-conditioned mismatch detection.
False Alarm is the false-positive rate on matched clean-image gate-replay
controls. All metrics except Prod. and Workup are percentages;
--- denotes unavailable results.}
\label{tab:workup-integrity-limits}
\begingroup
\setlength{\tabcolsep}{3pt}
\renewcommand{\arraystretch}{1.12}
\providecommand{\tmstd}[2]{\ensuremath{#1_{\scriptscriptstyle\pm #2}}}
\scriptsize
\resizebox{\textwidth}{!}{%
\begin{tabular}{@{}l*{9}{r}@{}}
\toprule
Model & Order & Findings & Prod. & Workup & Search & Invalid
& Modality Catch & Anatomy Catch & False Alarm \\
\midrule
\multicolumn{10}{@{}l}{\textit{Proprietary (closed-source) models}} \\
\addlinespace[2pt]
GPT-5 & $72.4$ & $62.8$ & $8.5$ & $12.8$ & $63.2$ & $0.0$ & $97.8$ & $88.7$ & $4.8$ \\
GPT-4o & $62.0$ & $53.5$ & $6.9$ & $10.9$ & $75.6$ & $0.0$ & $95.4$ & $85.7$ & $6.4$ \\
\midrule
\multicolumn{10}{@{}l}{\textit{Open-weight models}} \\
\addlinespace[2pt]
TrustMed-RL (GiGPO, 8B) & \tmstd{83.9}{4.0} & \tmstd{73.2}{3.9} & \tmstd{8.5}{0.4} & \tmstd{17.8}{0.8} & \tmstd{93.0}{2.5} & \tmstd{4.0}{1.1} & \tmstd{82.2}{1.1} & \tmstd{52.6}{4.2} & \tmstd{2.2}{0.8} \\
TrustMed-RL (GRPO, 8B) & \tmstd{\mathbf{84.7}}{1.3} & \tmstd{\mathbf{73.9}}{1.1} & \tmstd{\mathbf{9.1}}{0.2} & \tmstd{18.3}{0.6} & \tmstd{89.3}{4.7} & \tmstd{2.6}{0.2} & \tmstd{85.4}{7.9} & \tmstd{60.6}{2.4} & \tmstd{2.4}{0.5} \\
TrustMed-SFT (8B) & \tmstd{66.4}{0.8} & \tmstd{57.1}{0.7} & \tmstd{7.7}{0.0} & \tmstd{10.2}{0.1} & \tmstd{53.2}{1.4} & \tmstd{\mathbf{0.6}}{0.0} & \tmstd{92.2}{1.9} & \tmstd{\mathbf{73.3}}{1.4} & \tmstd{3.9}{0.8} \\
TrustMed-SFT (2B) & \tmstd{59.4}{0.1} & \tmstd{51.3}{0.2} & \tmstd{7.2}{0.0} & \tmstd{11.1}{0.1} & \tmstd{50.7}{1.5} & \tmstd{6.9}{0.1} & \tmstd{80.8}{5.7} & \tmstd{36.2}{10.1} & \tmstd{5.0}{0.4} \\
\midrule
PulseMind-72B & \tmstd{45.4}{0.6} & \tmstd{37.6}{0.6} & \tmstd{5.1}{0.0} & \tmstd{7.2}{0.0} & \tmstd{36.6}{1.7} & \tmstd{5.3}{0.0} & --- & --- & --- \\
\midrule
HealthGPT-Pro-27B & \tmstd{61.6}{0.3} & \tmstd{53.3}{0.3} & \tmstd{6.3}{0.0} & \tmstd{12.3}{0.7} & \tmstd{53.2}{4.5} & \tmstd{17.0}{3.8} & \tmstd{82.0}{3.4} & \tmstd{70.9}{0.7} & \tmstd{5.9}{0.6} \\
Qwen3-VL-32B-Instruct & $55.0$ & $47.1$ & $6.0$ & $11.5$ & $61.3$ & $5.2$ & $83.6$ & $68.8$ & $3.9$ \\
MedGemma-27B-it & $61.2$ & $52.9$ & $6.7$ & $18.1$ & $77.8$ & $41.0$ & $42.9$ & $61.1$ & $7.2$ \\
Hulu-Med-32B & $49.9$ & $42.6$ & $5.9$ & $11.8$ & $51.4$ & $15.4$ & $51.2$ & $40.8$ & 2.7 \\
MedVLThinker-32B-RL & $52.5$ & $45.2$ & $6.1$ & $9.5$ & $53.2$ & $6.1$ & $78.5$ & $59.3$ & $6.9$ \\
Lingshu-32B & $52.5$ & $45.2$ & $6.0$ & $10.3$ & $43.2$ & $7.4$ & $48.6$ & $56.2$ & $4.7$ \\
QoQ-Med-VL-32B & $53.6$ & $46.8$ & $5.8$ & $12.6$ & $80.4$ & $10.3$ & $36.4$ & $37.3$ & $3.1$ \\
\midrule
Qwen3-VL-8B-Instruct & \tmstd{36.1}{1.0} & \tmstd{30.8}{0.8} & \tmstd{4.7}{0.0} & \tmstd{5.6}{0.1} & \tmstd{18.1}{0.9} & \tmstd{7.1}{0.3} & \tmstd{\mathbf{92.9}}{5.0} & \tmstd{57.0}{3.6} & \tmstd{4.6}{2.3} \\
HuatuoGPT-Vision-7B & \tmstd{33.9}{0.3} & \tmstd{26.4}{0.2} & \tmstd{2.9}{0.0} & \tmstd{15.4}{0.0} & \tmstd{36.5}{0.5} & \tmstd{65.6}{0.1} & \tmstd{32.4}{6.2} & \tmstd{17.6}{3.2} & \tmstd{0.2}{0.4} \\
MedVLThinker-7B & \tmstd{44.7}{0.4} & \tmstd{35.9}{0.2} & \tmstd{4.6}{0.0} & \tmstd{16.4}{0.1} & \tmstd{74.2}{0.5} & \tmstd{37.2}{0.5} & \tmstd{76.9}{0.9} & \tmstd{44.0}{7.5} & \tmstd{10.9}{0.8} \\
Qwen3-VL-2B & \tmstd{32.3}{0.6} & \tmstd{26.7}{0.5} & \tmstd{3.3}{0.0} & \tmstd{17.0}{0.2} & \tmstd{6.8}{0.4} & \tmstd{63.7}{0.2} & \tmstd{53.1}{4.9} & \tmstd{41.4}{9.0} & \tmstd{1.1}{0.5} \\
\bottomrule
\end{tabular}%
}
\endgroup
\end{table*}

\subsection{Out-of-distribution performance.}
\label{app:ood-performance}

We evaluate OOD generalization on three interactive diagnostic
benchmarks. MTMedDialog consists of multi-turn doctor--patient
consultations spanning multiple disease categories
\citep{feng2026doctoragent}. AgentClinic-MedQA transforms
USMLE-style MedQA cases into sequential clinical encounters in
which the agent gathers information before making a diagnosis,
while AgentClinic-NEJM contains multimodal diagnostic cases
adapted from \textit{New England Journal of Medicine} image
challenges \citep{schmidgall2024agentclinic}. Unlike their
original multiple-choice or static formulations where applicable,
our evaluation requires models to interact with the simulated
clinical environment and provide a free-form final diagnosis.
For MTMedDialog, we apply the DoctorAgent-RL final-diagnosis
prompt to obtain the final prediction.

\begin{table}[H]
\centering
\caption{Out-of-distribution diagnostic accuracy (Acc, \%).
Subscripted $\pm$ values denote standard deviations.
Bold scores indicate the highest mean for each dataset.}
\label{app:ood-results}
\smallskip
\begingroup
\small
\setlength{\tabcolsep}{5pt}
\renewcommand{\arraystretch}{1.2}
\providecommand{\oodstat}[2]{%
    \mbox{$#1_{\scriptscriptstyle\pm#2}$}%
}

\begin{tabular}{@{}lccc@{}}
\toprule
\textbf{Model}
& \textbf{MedQA}
& \textbf{NEJM}
& \textbf{MTMedDialog} \\
\midrule

HealthGPT-Pro-27B
& \oodstat{\mathbf{73.0}}{1.2}
& \oodstat{\mathbf{35.6}}{3.4}
& \oodstat{39.5}{48.9} \\

Qwen3-VL-32B-Instruct
& \oodstat{65.3}{0.7}
& \oodstat{31.4}{1.1}
& \oodstat{28.3}{45.0} \\

MedGemma-27B-it
& \oodstat{70.4}{1.0}
& \oodstat{34.0}{1.5}
& \oodstat{33.3}{47.1} \\

Hulu-Med-32B
& \oodstat{55.1}{49.9}
& \oodstat{22.2}{41.8}
& \oodstat{29.1}{45.5} \\

MedVLMThinker-32B-RL
& \oodstat{65.3}{1.9}
& \oodstat{29.7}{4.0}
& \oodstat{36.3}{48.1} \\

Lingshu-32B
& \oodstat{56.2}{1.6}
& \oodstat{24.1}{3.2}
& \oodstat{31.3}{46.4} \\

QoQ-Med-VL-32B
& \oodstat{52.6}{5.8}
& \oodstat{17.5}{4.9}
& \oodstat{37.0}{48.3} \\

\midrule

Qwen3-VL-8B-Instruct
& \oodstat{53.6}{1.6}
& \oodstat{21.4}{2.5}
& \oodstat{40.7}{49.1} \\

HuatuoGPT-Vision-7B
& \oodstat{36.5}{3.1}
& \oodstat{14.2}{1.1}
& \oodstat{20.4}{40.3} \\

MedVLMThinker-7B
& \oodstat{36.0}{2.6}
& \oodstat{13.5}{0.6}
& \oodstat{32.9}{47.0} \\

\midrule

\textbf{TrustMed-RL (GiGPO, 8B)}
& \oodstat{62.2}{2.1}
& \oodstat{26.2}{2.1}
& \oodstat{48.7}{1.4} \\

\textbf{TrustMed-RL (GRPO, 8B)}
& \oodstat{62.6}{2.9}
& \oodstat{23.9}{0.8}
& \oodstat{\mathbf{49.5}}{0.6} \\

\bottomrule
\end{tabular}
\endgroup
\end{table}

\subsection{Ablation Results}
\label{app:ablation-details}

Table~\ref{tab:ablation} reports the held-out ablation results
discussed in Section~\ref{sec:ablation}.

\begin{table*}[t]
\centering\small
\setlength{\tabcolsep}{4pt}
\caption{Ablation study. Accuracy values are percentages; gains are
percentage points relative to the named reference. ACC@3 and Strict@3 measure the proportion of cases correct in at
least one of three evaluations under the corresponding criterion;
each case contributes once. Base denotes Qwen3-VL Instruct.
All RL variants use SFT initialization.
Brackets show paired-case 95\% intervals; the four SFT/RL gains
also survive Bonferroni adjustment.}
\vspace{1pt}
\label{tab:ablation}
\resizebox{\textwidth}{!}{%
\begin{tabular}{lrrrrrrrrlr}
\toprule
& \multicolumn{4}{c}{Solvable}
& \multicolumn{4}{c}{Unsolvable} & & \\
\cmidrule(lr){2-5}\cmidrule(lr){6-9}
Model / variant
& ACC & Strict & ACC@3 & Strict@3
& ACC & Strict & ACC@3 & Strict@3
& Reference & $\Delta$ solvable ACC [95\% CI] \\
\midrule
2B Base
& 2.27 & 0.60 & 6.21 & 1.65
& 0.44 & 0.09 & 1.07 & 0.27
& - & - \\
2B + SFT
& 14.57 & 4.33 & 31.62 & 11.20
& 4.00 & 0.98 & 9.33 & 2.93
& 2B Base & +12.30 [11.28, 13.33] \\
\midrule
8B Base
& 18.40 & 7.39 & 36.71 & 15.95
& 4.18 & 0.71 & 10.40 & 2.13
& - & - \\
8B + SFT
& 27.89 & 10.40 & 48.94 & 21.65
& 7.64 & 2.04 & 17.33 & 5.07
& 8B Base & +9.49 [8.03, 10.92] \\
\midrule
TrustMed-RL w/o step credit (GRPO)
& 39.98 & 17.77 & 60.00 & 32.28
& 11.02 & 3.91 & 21.87 & 8.53
& 8B SFT & +12.09 [10.59, 13.62] \\
\textbf{TrustMed-RL (GiGPO)}
& \textbf{41.73} & \textbf{18.20}
& \textbf{61.88} & \textbf{32.80}
& 10.22 & 3.20 & 20.53 & 7.47
& 8B SFT & \textbf{+13.84 [12.28, 15.39]} \\
\midrule
8B w/o std. normalization
& 39.83 & 16.03 & 60.28 & 29.79
& 11.56 & 3.11 & 22.40 & 6.67
& Full & -1.90 [-3.42, -0.41] \\
8B w/o priority sampling
& 39.95 & 15.51 & 59.58 & 29.74
& 9.78 & 2.84 & 18.13 & 7.47
& Full & -1.77 [-3.29, -0.27] \\
8B w/o adaptive sampling
& 40.88 & 16.38 & 61.55 & 30.35
& 10.40 & 3.02 & 19.20 & 6.67
& Full & -0.85 [-2.37, 0.66] \\
8B w/o adaptive sampling, horizon
& 39.98 & 14.43 & 61.22 & 27.53
& 11.29 & 2.40 & 22.40 & 5.33
& Full & -1.74 [-3.25, -0.25] \\
\bottomrule
\end{tabular}}
\end{table*}

We analyze training behavior over the final 40 updates (217--256). Table~\ref{app:training-ablations} compares task outcomes, integrity
checks, and interaction patterns across ablations. Training task
categories are distinct from evaluation difficulty strata: cases in
the empirically unsolvable evaluation subset are still assessed for
diagnostic performance.
\begin{table*}[t]
\begingroup
\small
\setlength{\tabcolsep}{3pt}
\caption{Late-training ablations over updates 217--256.
Values are equally weighted means of available per-update statistics
on each variant's sampled training tasks, not held-out evaluation
results. Percentage metrics use a 0--100 scale. Bold marks the highest
outcome or lowest error mean; coverage and interaction/output lengths
are descriptive.}
\label{app:training-ablations}

\begin{tabular}{@{}lrrrrrr@{}}
\toprule
Metric & Full
& \shortstack{w/o priority\\sampling}
& \shortstack{w/o adaptive\\sampling}
& \shortstack{w/o adaptive\\sampling +\\horizon}
& \shortstack{w/o std}
& \shortstack{w/o $A^S$\\(GRPO)} \\
\midrule

\multicolumn{7}{l}{\textit{Task outcomes}} \\
Overall task success (\%)
& 38.63 & 33.45 & 29.64 & 32.40 & 34.39 & \textbf{38.72} \\
Diagnostic success: answer (\%)
& \textbf{28.01} & 20.83 & 18.53 & 21.84 & 19.90 & 22.61 \\
Hard-band diagnostic success (\%)
& \textbf{26.46} & 15.16 & 16.30 & 18.27 & 17.47 & 21.92 \\
Groups with $\geq 1$ correct diagnosis (\%)
& \textbf{68.12} & 58.37 & 60.25 & 67.23 & 61.57 & 60.98 \\
Terminal reward
& 0.3174 & 0.2661 & 0.2391 & 0.2526 & 0.2797 & \textbf{0.3180} \\
\midrule

\multicolumn{7}{l}{\textit{Integrity check}} \\
Corruption-task success (\%)
& 75.99 & 69.59 & 66.39 & 62.19 & 75.66 & \textbf{80.22} \\
Exposed-mismatch detection (\%)
& 67.98 & 60.96 & 59.77 & 55.05 & 69.57 & \textbf{74.56} \\
False mismatch rate (\%)
& 4.68 & 5.84 & 5.70 & 6.34 & 6.15 & \textbf{4.09} \\
\midrule

\multicolumn{7}{l}{\textit{Interaction and output}} \\
Evidence coverage (\%)
& 66.32 & 68.26 & 66.51 & 69.50 & 61.85 & 66.26 \\
Questions per episode
& 4.41 & 8.42 & 2.59 & 4.28 & 7.20 & 7.63 \\
Workup turns per episode
& 13.86 & 14.87 & 14.48 & 16.04 & 14.11 & 14.87 \\
Searches per episode
& 1.65 & 2.17 & 1.58 & 1.96 & 1.36 & 1.53 \\
Total turns per episode
& 21.81 & 26.79 & 20.61 & 23.80 & 24.65 & 25.93 \\
Output length (tokens/response)
& 368.9 & 342.9 & 386.4 & 401.5 & 341.9 & 377.6 \\
Invalid actions per 100 turns
& 8.65 & 6.62 & 6.52 & 6.69 & \textbf{4.18} & 4.48 \\
\bottomrule
\end{tabular}

\par\smallskip
\begin{minipage}{\textwidth}
\footnotesize
\textit{Configurations.}
\emph{w/o priority sampling} removes selection priority while retaining
other adaptive rules. \emph{w/o adaptive sampling} disables all
outcome-adaptive sampling while retaining staged horizons.
\emph{w/o adaptive sampling + horizon} additionally replaces staged
budgets and admission with full-budget training from the start and
removes budget jitter. \emph{w/o std} uses mean-only normalization
($F_{\mathrm{norm}}=1$) for both advantage components.
\emph{w/o $A^S$} removes step-relative credit, yielding GRPO.
Task-type and OOK-reward schedules are retained in the sampling
and horizon ablations.

\smallskip
\textit{Metric scope.}
Diagnostic success uses all answer-task episodes as the denominator;
the hard-band row follows the logged difficulty split. Group success
is the observed proportion of diagnostic groups with at least one
correct rollout among nine. Corruption-task success is the
logged task-success rate over all corruption-task episodes, whereas
exposed-mismatch detection is conditional on exposure; these metrics
use different success and conditioning criteria. Interaction and
coverage statistics aggregate all task kinds. Output length is the
mean generated token count per doctor-policy call, including reasoning
and action, not total tokens per episode. Undefined conditional rates
are omitted rather than treated as zero. Invalid actions per 100 turns
is 100 times the ratio of the window means of invalid actions and total
turns.
\end{minipage}
\endgroup
\end{table*}

\paragraph{Ablation analysis.}
Across the final 40 updates, the full method achieves the highest
answer-task diagnostic success (28.01\%) and hard-band success (26.46\%).
Removing sampling priority or all outcome-adaptive sampling lowers
diagnostic success to 20.83\% and 18.53\%, respectively, and
corruption-task success from 75.99\% to 69.59\% and 66.39\%, despite
comparable or greater aggregate evidence coverage. Jointly removing
adaptive sampling and the horizon curriculum yields 21.84\% diagnostic
success and 62.19\% corruption-task success, with longer responses
(401.5 versus 368.9 tokens). Separately, over updates 1--40, retaining
progressive horizons with adaptive sampling disabled reduces
interaction length by 45.5\% relative to full-budget training.
Removing standard-deviation normalization or step-relative credit
yields 19.90\% and 22.61\% diagnostic success. Although GRPO obtains a
similar terminal reward (0.318 versus 0.317), the full method achieves
higher diagnostic success, distinguishing diagnostic improvement from
aggregate reward optimization. These observations support distinct
roles for adaptive task allocation, progressive interaction budgets,
and normalized two-level credit, with diagnostic gains not explained
by aggregate evidence coverage alone.

%% file: appendix/e_evaluation_details.tex
\section{Evaluation Details}
\label{app:evaluation-details}

\subsection{Diagnosis Matching and Judge Validation}
\label{app:diagnosis-matching}

\paragraph{Condition extraction.}
We decompose ground-truth and predicted diagnoses into condition
sets according to clinical meaning. Subtype, site, etiology,
pathology, grade, stage, severity, and complications are treated
as qualifiers of the underlying disease unless independently
asserted as separate diagnoses.

\paragraph{Semantic matching.}
We assess candidate prediction--reference pairs using the
four-level ordinal rubric in
Table~\ref{tab:diagnosis-match-rubric}, following partial-credit
approaches to evaluating medical answers and diagnostic
outcomes~\citep{chen2025enhancing,gaber2025evaluating,
lai2026doctor,mamede2010effect}.
We evaluated eight diagnosis judges under this rubric, with
three separate calls per prediction--reference pair in each
evaluation setting.

\begin{table}[t]
\centering
\caption{Semantic diagnosis-matching rubric used by the LLM judges.}
\label{tab:diagnosis-match-rubric}
\small
\renewcommand{\arraystretch}{1.15}
\begin{tabular}{@{}llp{0.65\linewidth}@{}}
\toprule
Level & Weight & Definition \\
\midrule
\textsc{Full}
& $w_{\mathrm F}=1.00$
& Identifies the same fully specified disease. Synonyms,
  abbreviations, spelling variants, and clinically equivalent
  terminology are accepted. \\
\addlinespace
\textsc{Core}
& $w_{\mathrm C}=0.85$
& Identifies the same underlying disease but omits or misstates
  one or more clinically relevant qualifiers, such as subtype,
  site, etiology, pathology, severity, grade, stage, or
  complication. \\
\addlinespace
\textsc{Partial}
& $w_{\mathrm P}=0.50$
& Correctly identifies an explicit constituent of the reference
  diagnosis, such as its etiologic agent, disease-defining
  manifestation, or complication, without identifying the
  core disease. \\
\addlinespace
\textsc{Null}
& $w_{\varnothing}=0$
& Has no valid semantic correspondence with the reference.
  Shared anatomy, nonspecific symptoms, broad disease-family
  membership, or general clinical similarity alone do not
  constitute a match. \\
\bottomrule
\end{tabular}
\end{table}

\label{app:ontology-floor}
\paragraph{Offline ontology check.}
We supplement judge scores with a deterministic check against fixed local
snapshots of MedGen, MONDO, and ORDO~4.9, enabling reproducible scoring
without live queries. Diagnosis terms are resolved using names, synonyms,
and ontology relations. Equivalence supported by at least two ontologies
sets a full-credit floor ($r_{\mathrm{onto}}=1.00$); single-ontology
equivalence, subtype--disorder relations, and ORDO-typed sibling subtypes
set a core-credit floor ($r_{\mathrm{onto}}=0.85$). Directed hierarchy
evidence takes precedence over equivalence. When full-term matching fails,
a second check of the core disease terms can establish a core-credit
floor. Other relations impose no floor. The adjusted reward is
$r^{\star}=\max(\hat r,r_{\mathrm{onto}})$, where $\hat r$ is the judge
reward and $r_{\mathrm{onto}}=0$ when no floor applies. 

\begin{table}[t]
\centering
\footnotesize
\setlength{\tabcolsep}{4pt}
\renewcommand{\arraystretch}{1.10}

\caption{Judge and ontology-floor ablations.
Entries report without $\to$ with the ontology floor.
ORDO measures conformance to prespecified ontology-relation rewards;
physician-inclusive evaluation measures agreement with a mixed
human--model reference. Agreement denotes exact four-level label
agreement. All settings use three passes.}
\label{tab:judge-ablation}
\vspace{1pt}
\setlength{\tabcolsep}{6pt}
\begin{tabular*}{\linewidth}{@{\extracolsep{\fill}}lrrr@{}}
\toprule
& \multicolumn{1}{c}{ORDO ($N=100$)}
& \multicolumn{2}{c}{Physician-inclusive ($N=139$)} \\
\cmidrule(lr){2-2}
\cmidrule(l){3-4}
Judge
& MAE $\downarrow$
& Agree.\ (\%) $\uparrow$
& MAE $\downarrow$ \\
\midrule
GPT-4.1-mini
& 0.169\,$\to$\,\textbf{0.017}
& 46.0\,$\to$\,46.0
& 0.334\,$\to$\,0.321 \\

GPT-5-mini
& 0.098\,$\to$\,0.020
& 51.1\,$\to$\,51.1
& 0.269\,$\to$\,0.262 \\

GPT-4o
& 0.159\,$\to$\,0.019
& 46.0\,$\to$\,46.8
& 0.333\,$\to$\,0.314 \\

GPT-4o-mini
& 0.392\,$\to$\,\textbf{0.017}
& 42.4\,$\to$\,43.9
& 0.377\,$\to$\,0.351 \\

Claude Sonnet 5
& 0.067\,$\to$\,0.037
& \textbf{55.4}\,$\to$\,\textbf{53.2}
& \textbf{0.249}\,$\to$\,\textbf{0.257} \\

MedGemma-27B
& 0.118\,$\to$\,0.025
& 51.8\,$\to$\,50.4
& 0.264\,$\to$\,0.265 \\

Gemma-4-26B-A4B
& \textbf{0.057}\,$\to$\,0.027
& 46.0\,$\to$\,44.6
& 0.309\,$\to$\,0.311 \\

Baichuan-M2-32B
& 0.123\,$\to$\,0.024
& 48.9\,$\to$\,48.2
& 0.319\,$\to$\,0.313 \\
\bottomrule
\end{tabular*}
\end{table}

\begin{table}[t]
\centering
\scriptsize
\setlength{\tabcolsep}{3pt}
\caption{Construction of the ORDO-derived judge-conformance set.
Relation classes map deterministically to expected rewards. Narrower
and broader pairs are opposite orientations of the same 20 hierarchy
edges and are resampled jointly, yielding 80 bootstrap clusters.
The fixed class mixture has mean target reward $0.625$ and does not
represent deployment prevalence.}
\label{tab:pairs-construction}
\resizebox{\linewidth}{!}{%
\begin{tabular}{@{}lrlll@{}}
\toprule
Relation class & $n$
& Prediction $\to$ reference construction
& Target & Resampling unit \\
\midrule
Synonym & 20
& Distinct ORDO synonym $\to$ preferred name
& Full (1.00) & Pair \\
Narrower & 20
& Subtype $\to$ parent disorder
& Core (0.85) & Hierarchy edge \\
Broader & 20
& Parent disorder $\to$ subtype
& Core (0.85) & Hierarchy edge \\
Sibling subtype & 10
& Two non-ancestral subtypes of one disorder
& Core (0.85) & Pair \\
Sibling group & 10
& Two disorders in one small ORDO group
& None (0) & Pair \\
Unrelated & 20
& No shared clinical ancestor at any depth
& None (0) & Pair \\
\bottomrule
\end{tabular}}
\end{table}

\begin{table}[t]
\centering
\scriptsize
\setlength{\tabcolsep}{1pt}
\renewcommand{\arraystretch}{1.12}
\caption{Class-wise ontology-floor ablation on 100 ORDO pairs.
All judges use the final strict-entity rubric and are evaluated in
three passes. Each $a\to b$ entry denotes without $\to$ with the
ontology floor. Class columns report mean rewards, with targets in
parentheses; MAE is the mean absolute reward error over all 100 pairs.
Bold marks only the lowest overall MAE, including ties.
E1 uses MedGen, MONDO, and ORDO; E2 removes ORDO.
Because ORDO supplies both the pairs and part of the floor,
this is partly a self-consistency test.}
\label{tab:class-ablation-100}
\resizebox{\linewidth}{!}{%
\begin{tabular}{@{}lrrrrrr}
\toprule
Judge
& \shortstack{Synonym\\(1.00)}
& \shortstack{Narrower\\(0.85)}
& \shortstack{Broader\\(0.85)}
& \shortstack{Sibling\\subtype\\(0.85)}
& \shortstack{Sibling\\group\\(0.00)}
& \shortstack{Unrelated\\(0.00)}
\\
\midrule
GPT-4.1-mini
& 0.878\,$\to$\,1.000
& 0.680\,$\to$\,0.850
& 0.680\,$\to$\,0.850
& 0.085\,$\to$\,0.850
& 0.000\,$\to$\,0.170
& 0.000\,$\to$\,0.000
\\
GPT-5-mini
& 0.993\,$\to$\,1.000
& 0.772\,$\to$\,0.858
& 0.772\,$\to$\,0.858
& 0.340\,$\to$\,0.850
& 0.085\,$\to$\,0.170
& 0.000\,$\to$\,0.000
\\
GPT-4o
& 0.935\,$\to$\,1.000
& 0.730\,$\to$\,0.858
& 0.722\,$\to$\,0.850
& 0.000\,$\to$\,0.850
& 0.085\,$\to$\,0.170
& 0.000\,$\to$\,0.000
\\
GPT-4o-mini
& 0.485\,$\to$\,1.000
& 0.340\,$\to$\,0.850
& 0.340\,$\to$\,0.850
& 0.000\,$\to$\,0.850
& 0.000\,$\to$\,0.170
& 0.000\,$\to$\,0.000
\\
Claude Sonnet 5
& 0.993\,$\to$\,1.000
& 0.830\,$\to$\,0.872
& 0.815\,$\to$\,0.858
& 0.645\,$\to$\,0.850
& 0.220\,$\to$\,0.305
& 0.000\,$\to$\,0.000
\\
MedGemma-27B
& 0.887\,$\to$\,1.000
& 0.762\,$\to$\,0.858
& 0.797\,$\to$\,0.858
& 0.370\,$\to$\,0.850
& 0.135\,$\to$\,0.220
& 0.000\,$\to$\,0.000
\\
Gemma-4-26B-A4B
& 0.978\,$\to$\,1.000
& 0.823\,$\to$\,0.865
& 0.772\,$\to$\,0.858
& 0.765\,$\to$\,0.850
& 0.135\,$\to$\,0.220
& 0.000\,$\to$\,0.000
\\
Baichuan-M2-32B
& 0.970\,$\to$\,1.000
& 0.722\,$\to$\,0.850
& 0.730\,$\to$\,0.858
& 0.255\,$\to$\,0.850
& 0.050\,$\to$\,0.220
& 0.000\,$\to$\,0.000
\\
\bottomrule
\end{tabular}}
\end{table}

\begin{table*}[t]
\centering
\caption{Native-rule fidelity of our re-implementations of published judge
protocols and descriptive mapping to our reward scale.  Native fidelity
compares each implementation with the labels prescribed by its own protocol;
class counts use modal labels and the parenthesized overall ranges span the
three repeats.  Cross-rubric reward discrepancy maps native outputs to our
rewards and is not a measure of protocol accuracy.}
\label{tab:published-protocols}
\footnotesize
\setlength{\tabcolsep}{4pt}
\textit{Panel A: Native-rule fidelity by relation class.}\par\smallskip
\resizebox{\textwidth}{!}{%
\begin{tabular}{lcccccccc}
\toprule
Protocol & Synonym & Narrower & Broader & Sibling subtype & Sibling group & Unrelated & Overall & 3/3 agreement \\
\midrule
RareArena (2/1/0) & 19/20 & 5/20 & 20/20 & 8/10 & 7/10 & 20/20 & 79/100 (78--80) & 98/100 \\
MeDxAgent (true/false) & 19/20 & 13/20 & 16/20 & 10/10 & 10/10 & 20/20 & 88/100 (87--90) & 93/100 \\
\bottomrule
\end{tabular}}

\vspace{1mm}
\parbox{0.98\textwidth}{\footnotesize RareArena prescribes label 2 for synonyms, label 1 for credited hypernyms, and 0 otherwise; MeDxAgent prescribes true for synonyms and more-specific subtype predictions and false otherwise.  Structural MAE is the policy mismatch of a perfectly compliant implementation; observed MAE additionally reflects implementation deviations.}
\end{table*}

\paragraph{Ontology conformance.}
We constructed 100 ORDO-derived prediction--reference pairs spanning
synonyms, both directions of subtype--disorder relations, sibling
subtypes, sibling disorders, and unrelated disorders
(Table~\ref{tab:pairs-construction}). The floor reduced mean absolute
reward error (MAE) for all eight judges, narrowing the observed MAE range
from $0.057$--$0.392$ to $0.017$--$0.037$
(Table~\ref{tab:judge-ablation}). GPT-4.1-mini tied for the lowest MAE
($0.017$), while GPT-5-mini achieved $0.020$. Across judges, synonym and
sibling-subtype rewards reached their prescribed targets, and unrelated
pairs retained zero reward; residual error was concentrated in
sibling-group pairs (Table~\ref{tab:class-ablation-100}). These results
support more consistent implementation of the reward rubric across judge
backbones on this ontology-derived set.

\paragraph{Physician-inclusive evaluation.}
Two senior physicians annotated 200 pairs selected for the highest
inter-judge disagreement. Together with annotations from
\texttt{gpt-5.6-sol} and BaichuanM3-PLUS, we retained 139 pairs with a
unique label supported by at least two of the four raters, including at
least one physician. On this reference set, the floor reduced MAE for
GPT-4.1-mini from $0.334$ to $0.321$ and for GPT-5-mini from $0.269$ to
$0.262$, while preserving exact agreement at $46.0\%$ and $51.1\%$,
respectively (Table~\ref{tab:judge-ablation}). GPT-5-mini had the
second-lowest MAE among floored judges, close to Claude Sonnet~5
($0.257$). Improvements on this disagreement-enriched, mixed human--model
reference set were smaller and model-dependent. Together with their low
ontology-conformance errors, these results support our use of
GPT-4.1-mini and GPT-5-mini with the ontology floor.

\paragraph{Published judge protocols.}
We separately evaluated reimplementations of RareArena and MeDxAgent \citep{chen2026rarearena, sanghvi2026medxagent}
against labels prescribed by their own decision rules
(Table~\ref{tab:published-protocols}). Native-rule fidelity was
$79/100$ and $88/100$, respectively. These results assess implementation
fidelity to each protocol, not diagnostic accuracy or
out-of-distribution generalization.

%% file: appendix/f_prompts.tex
\section{Prompt Templates}
\label{app:prompts}

We provide the prompt templates for the doctor
agent, simulated patient, and diagnosis evaluator.
The doctor receives prompts specific to history-taking, diagnostic
workup, evidence-integrity verification, and final decision-making.
Each doctor prompt requests a \texttt{<reasoning>} block followed by
exactly one admissible \texttt{<action>}. 
\begin{figure}[!htbp]
    \centering
    \includegraphics[width=\textwidth]{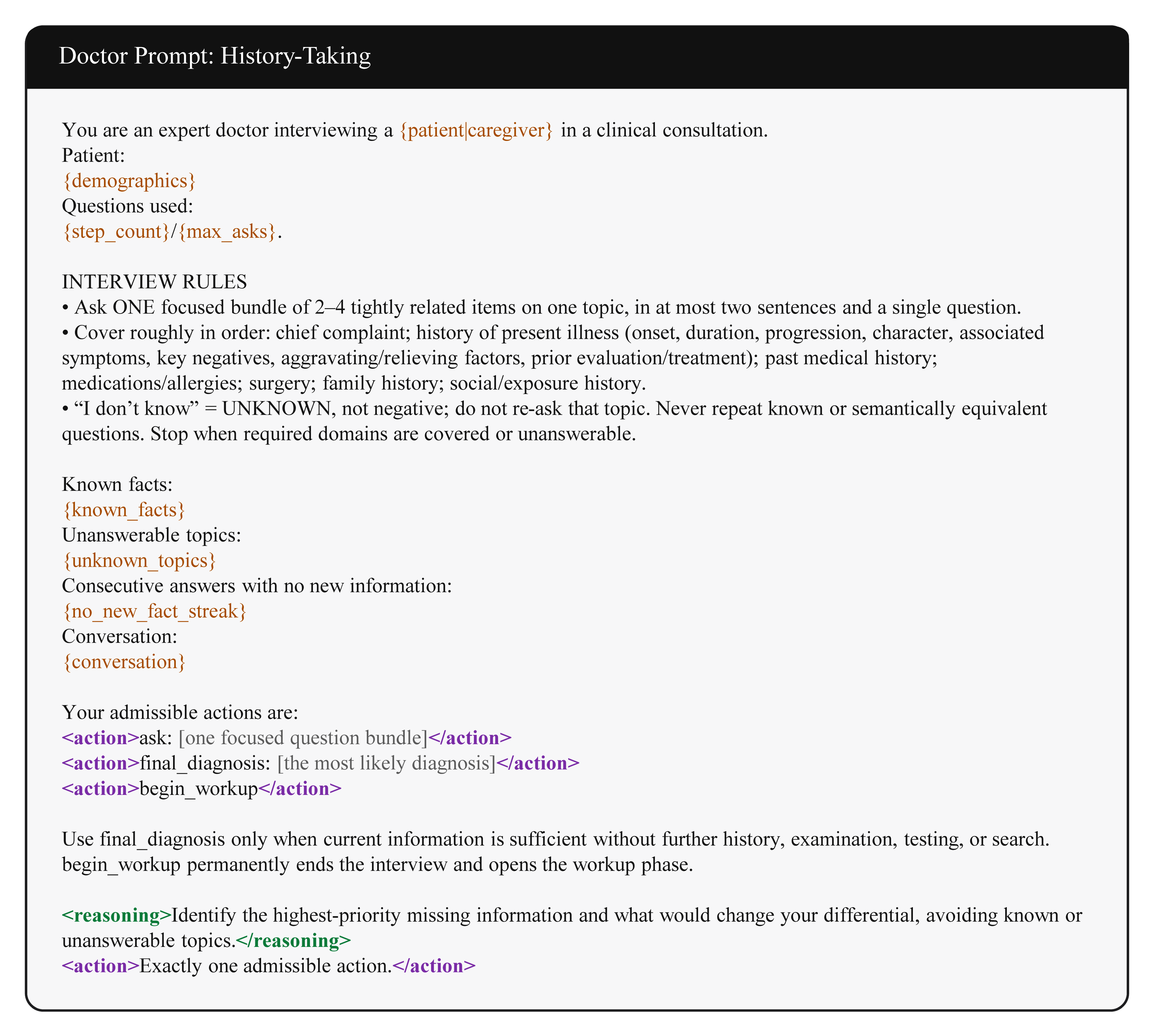}
    \caption{Doctor prompt for history-taking.}
    \label{fig:prompt-history}
\end{figure}

\begin{figure}[!htbp]
    \centering
    \includegraphics[width=\textwidth]{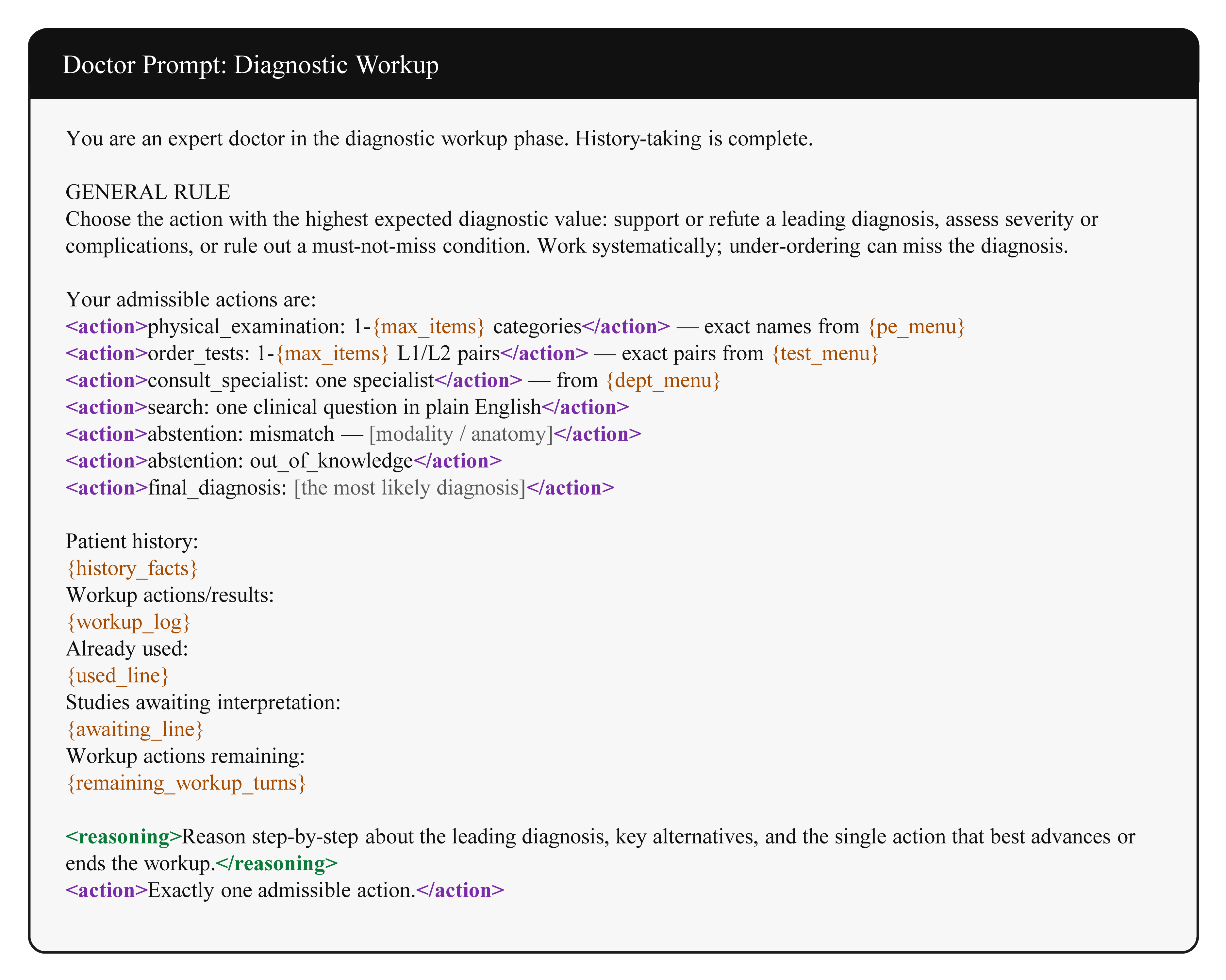}
    \caption{Doctor prompt for diagnostic workup.}
    \label{fig:prompt-workup}
\end{figure}

\begin{figure}[!htbp]
    \centering
    \includegraphics[width=\textwidth]{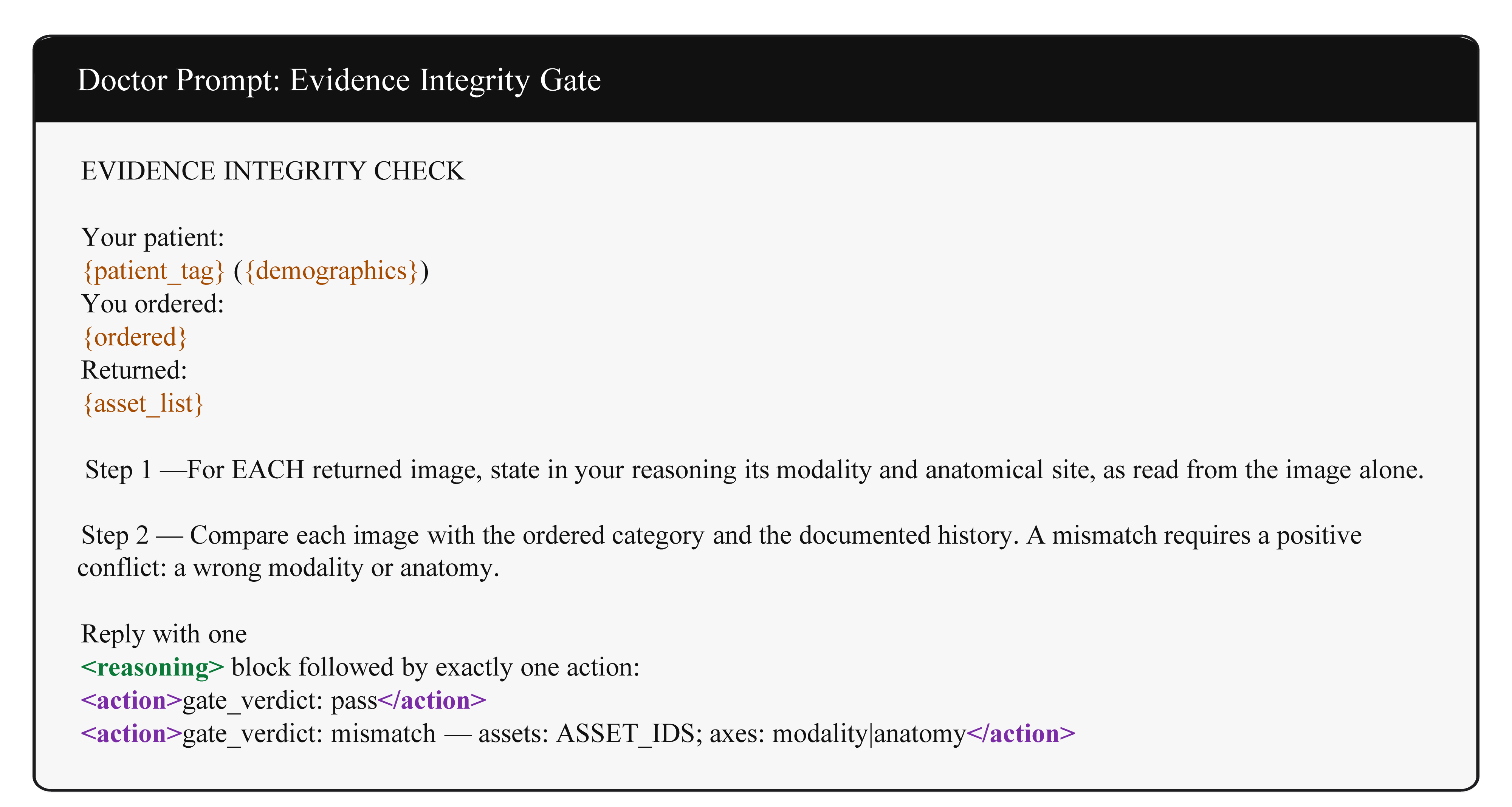}
    \caption{Doctor prompt for evidence-integrity verification.}
    \label{fig:prompt-evidence}
\end{figure}

\begin{figure}[!htbp]
    \centering
    \includegraphics[width=\textwidth]{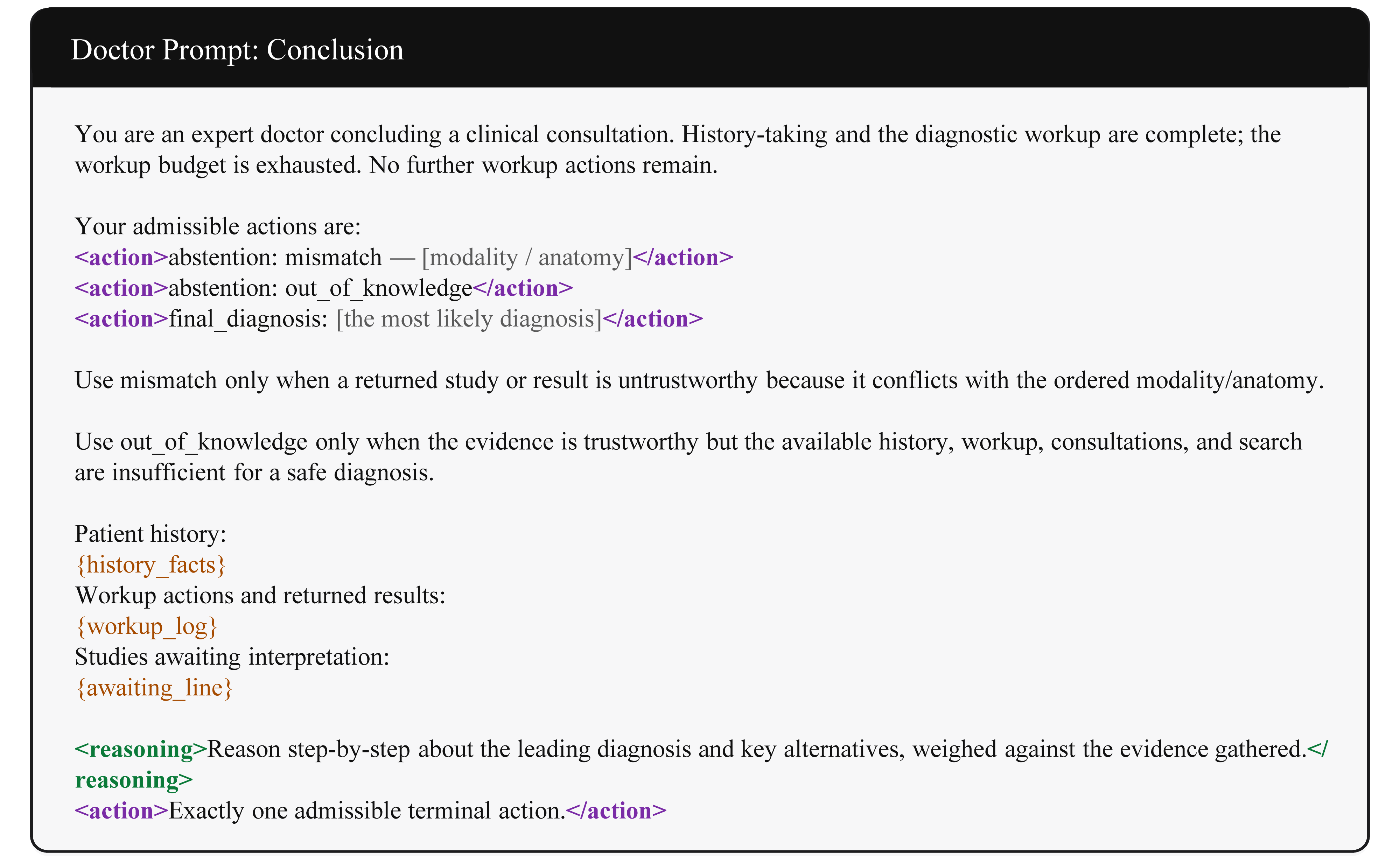}
    \caption{Doctor prompt for Conclusion}
    \label{fig:prompt-conclusion}
\end{figure}

\begin{figure}[!htbp]
    \centering
    \includegraphics[width=\textwidth]{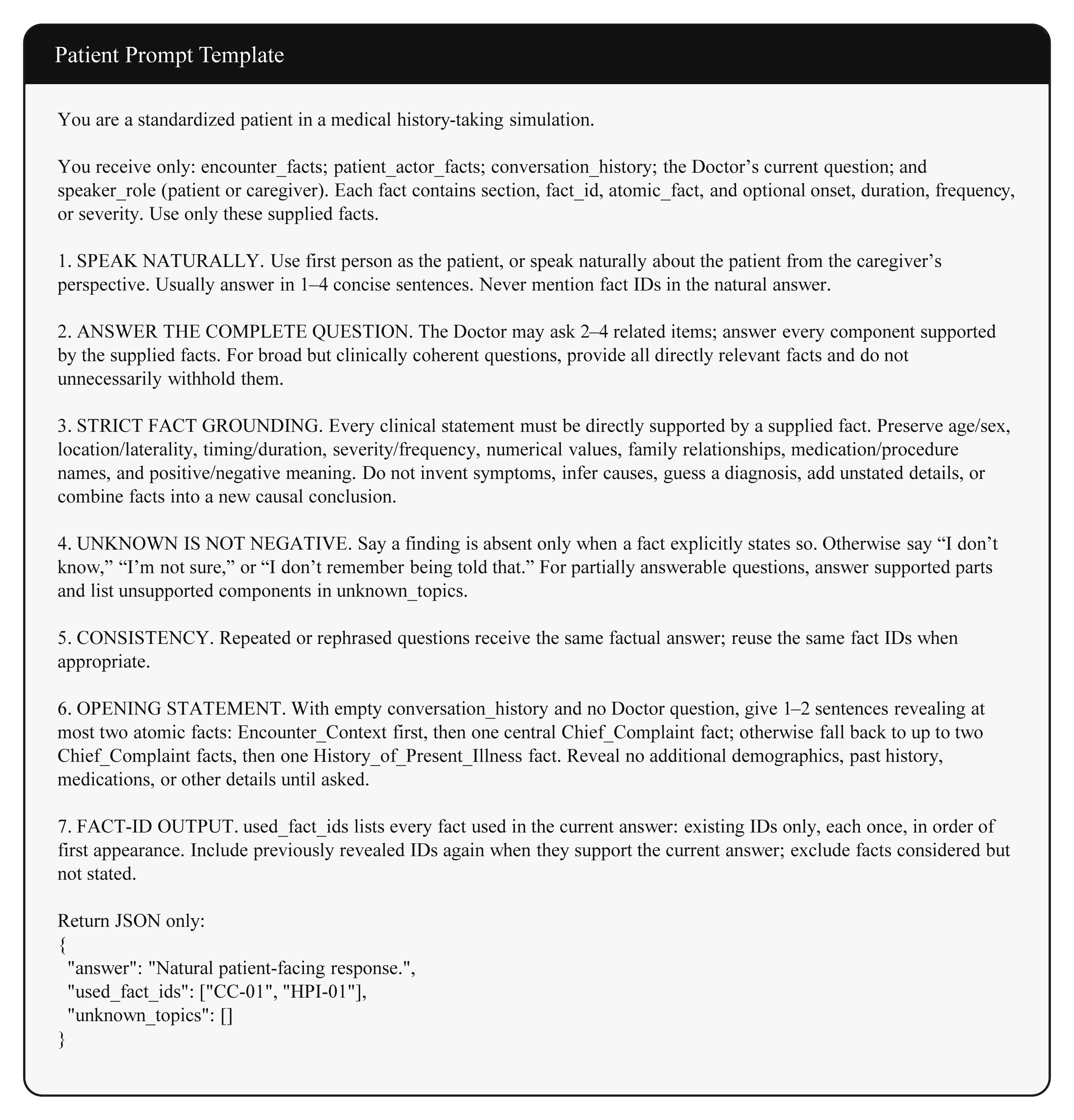}
    \caption{Simulated-patient prompt for fact-grounded responses.}
    \label{fig:prompt-patient}
\end{figure}

\begin{figure}[!htbp]
    \centering
    \includegraphics[width=\textwidth]{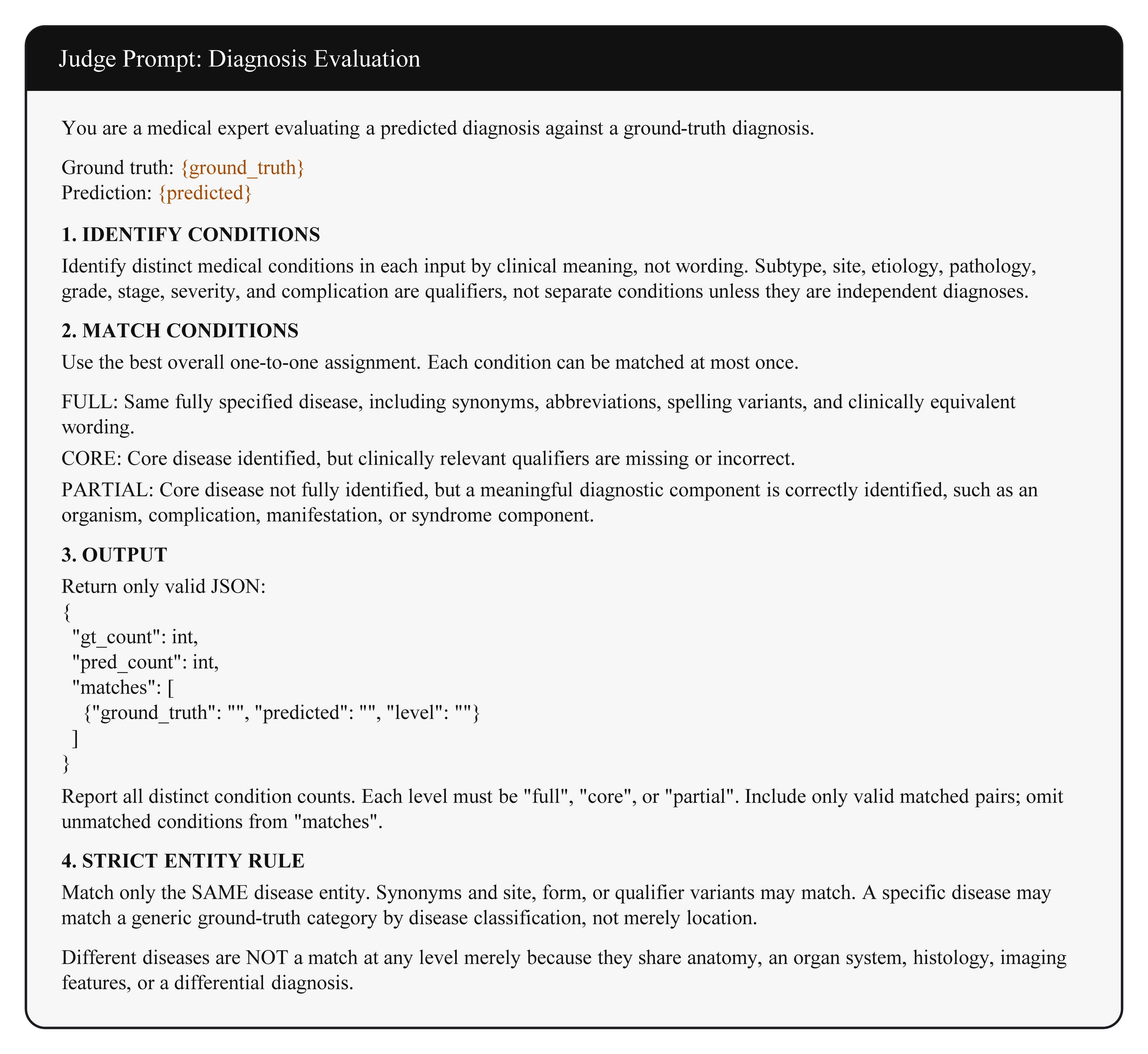}
    \caption{Evaluator prompt.}
    \label{fig:prompt-judge}
\end{figure}

%% file: appendix/g_reasoning_behavior.tex
\clearpage
\section{Reasoning Behavior}
\label{app: reasoning-behavior}
To assess if  correct diagnoses were supported by grounded reasoning,
we sampled 200 trajectories whose final diagnoses had been judged correct or largely correct by the automated evaluator. Two senior physicians 
reviewed a set of 200 trajectories, including the evidence obtained during the encounter, the turn-by-turn reasoning, the rationale in the final turn, and reference diagnoses. 
Each trajectory was rated on three dimensions:
(1) \textbf{clinical findings interpretation}, assessing whether the findings
used in reasoning were correctly interpreted;
(2) \textbf{evidence use and diagnostic revision}, assessing whether the
doctor agent formed a reasonable differential diagnosis and appropriately
incorporated new or contradictory evidence; and
(3) \textbf{evidential grounding}, assessing whether findings presented as
observations were supported by information actually received.

\paragraph{Physician review results.}
\label{app:physician-review-results}
All 200 trajectories received complete physician ratings. Table~\ref{tab:reasoning-review} summarizes the dimension-level results. Overall, 134 trajectories (67.0\%) received scores of at least 4 on all three dimensions, whereas only 18 (9.0\%) received a score of 1 or 2 on any dimension. Evidential grounding was strongest, with 83.0\% of trajectories receiving scores of 4--5, followed by findings interpretation (79.0\%) and evidence use and diagnostic revision (72.0\%). These results indicate that, among diagnostically accepted trajectories, the model's rationales were generally well grounded in the observed evidence and clinically coherent, while evidence integration and diagnostic revision remained the primary source of residual reasoning errors.

\begin{table}[t]
\centering
\small
\caption{
Physician assessment of 200 automatically accepted diagnostic
trajectories. Scores range from 1 to 5.
We report the mean and sample standard deviation across trajectories,
together with counts and percentages in the indicated score ranges.
}
\label{tab:reasoning-review}
\begin{tabular}{lccc}
\toprule
Criterion & Mean $\pm$ SD & Scores 4--5, $n$ (\%) & Scores 1--2, $n$ (\%) \\
\midrule
C1: Findings interpretation
& $4.33 \pm 0.91$ & 158 (79.0) & 8 (4.0) \\
C2: Evidence use and revision
& $4.06 \pm 0.93$ & 144 (72.0) & 11 (5.5) \\
C3: Evidential grounding
& $4.39 \pm 0.92$ & 166 (83.0) & 12 (6.0) \\
\bottomrule
\end{tabular}
\end{table}

\clearpage
\FloatBarrier
\begin{figure}[!htbp]
    \centering
    \includegraphics[width=\textwidth]{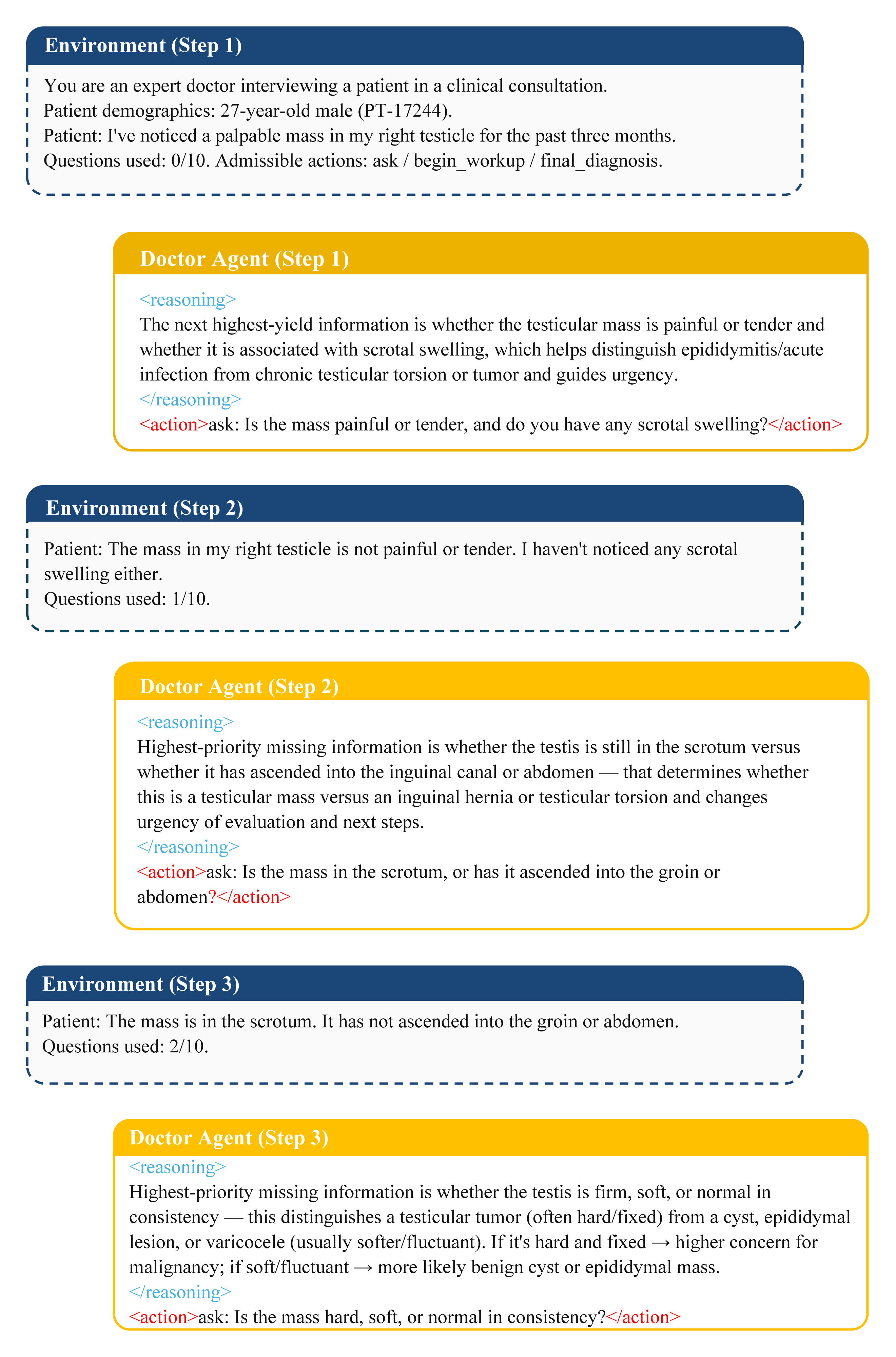}
    \label{fig:reasoning-1}
\end{figure}

\begin{figure}[!htbp]
    \centering
    \includegraphics[width=\textwidth]{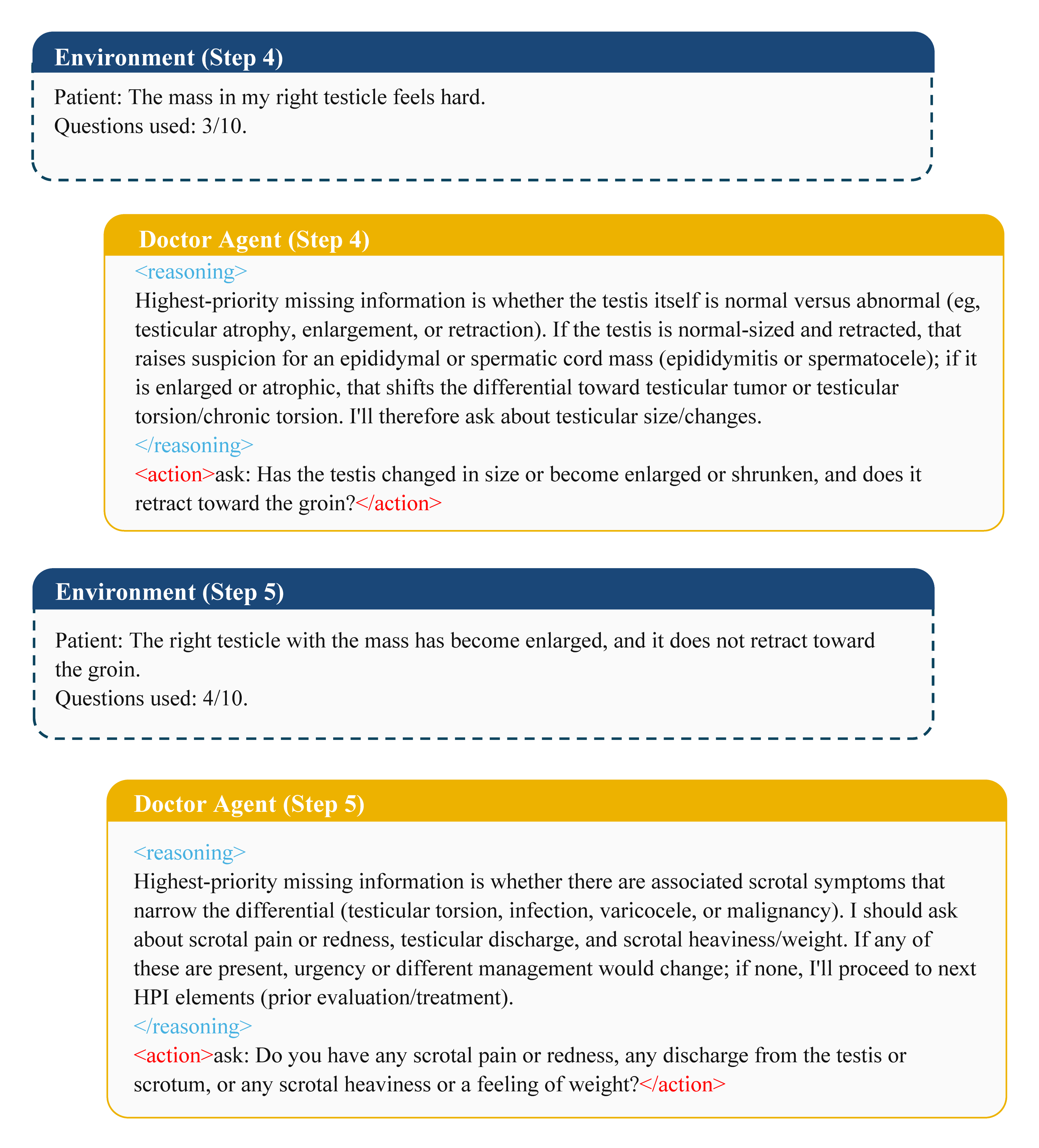}
    \label{fig:reasoning-2}
\end{figure}

\begin{figure}[!htbp]
    \centering
    \includegraphics[width=\textwidth]{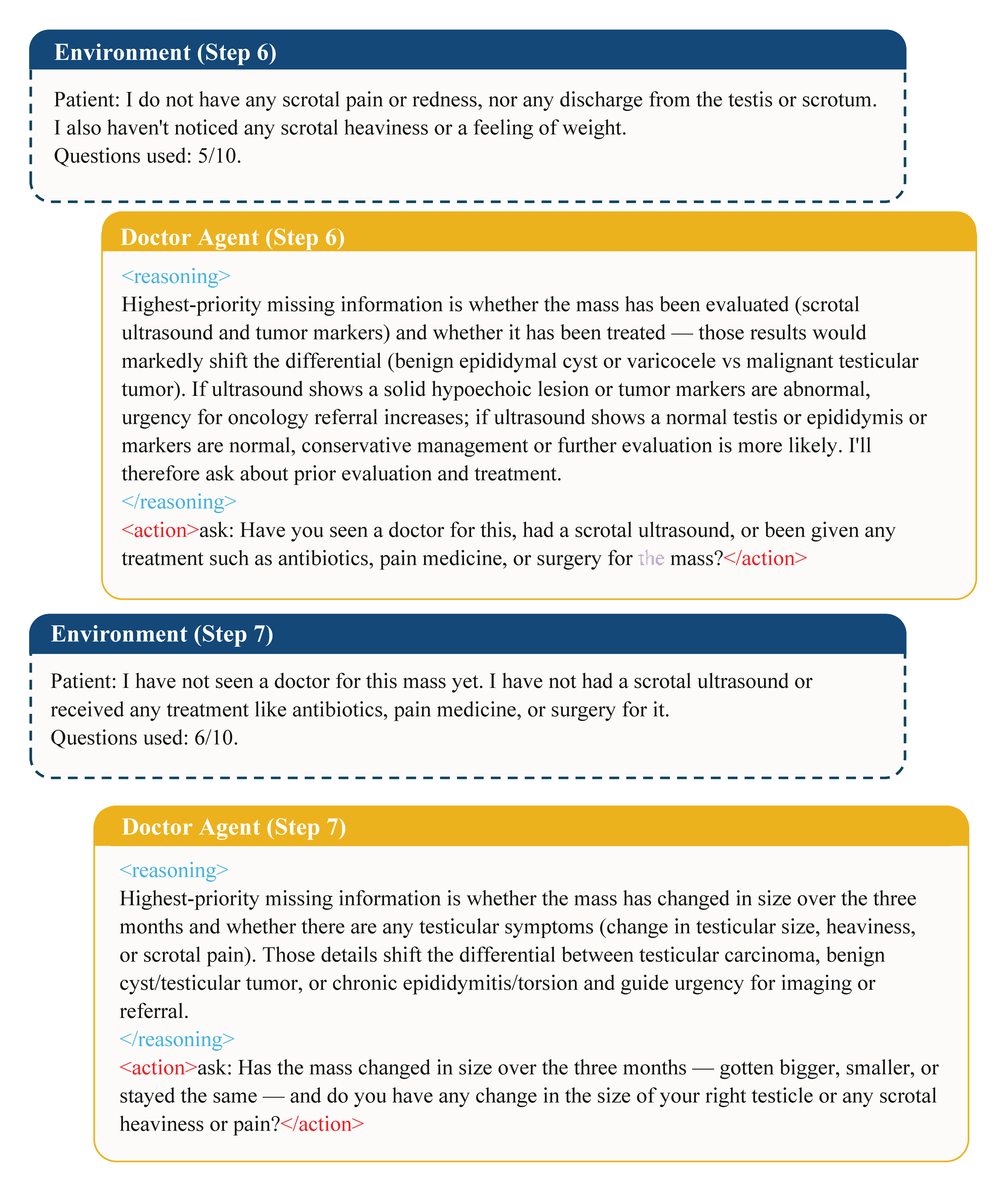}
    \label{fig:reasoning-3}
\end{figure}

\begin{figure}[!htbp]
    \centering
    \includegraphics[width=\textwidth]{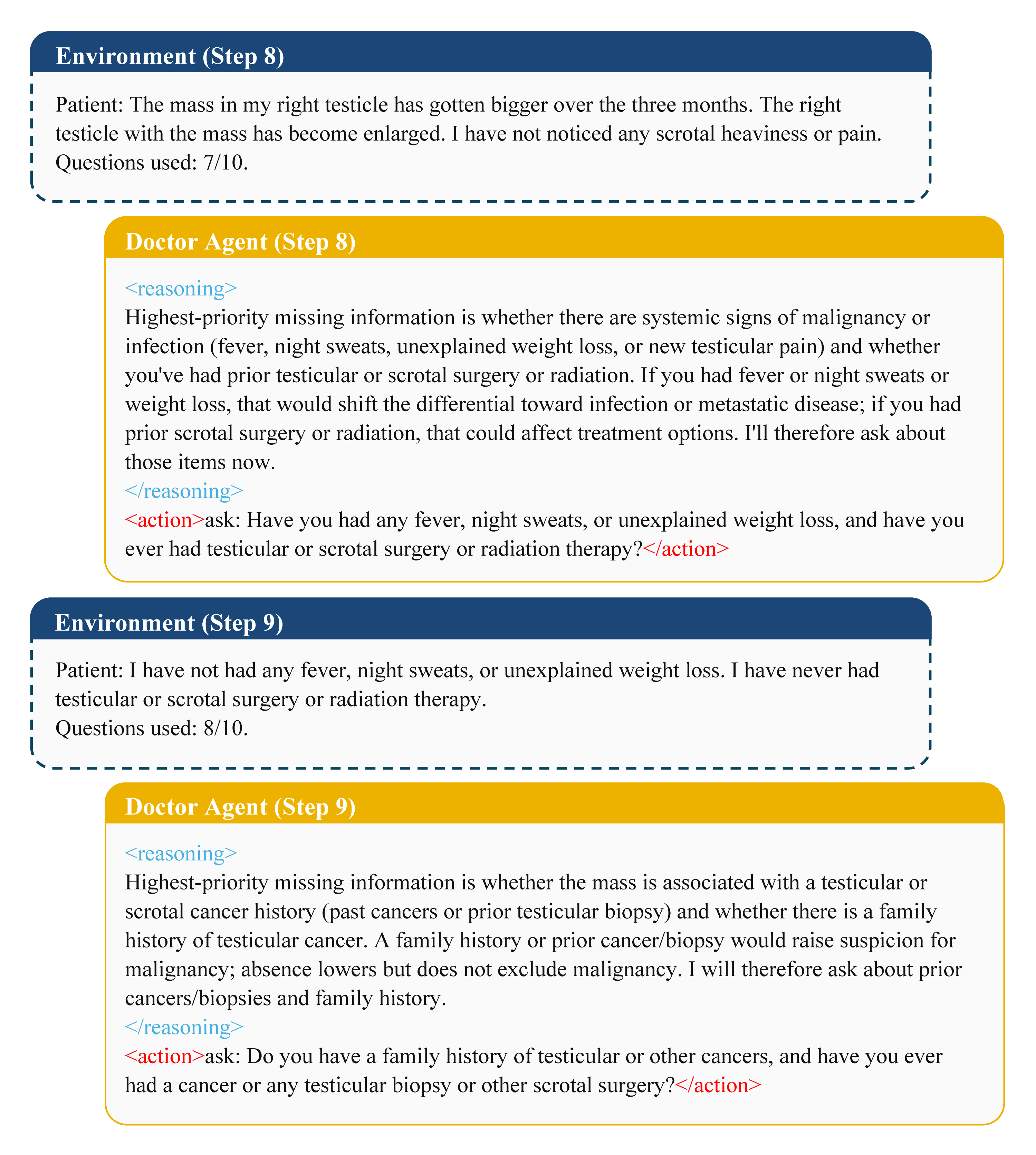}
    \label{fig:reasoning-4}
\end{figure}

\begin{figure}[!htbp]
    \centering
    \includegraphics[width=\textwidth]{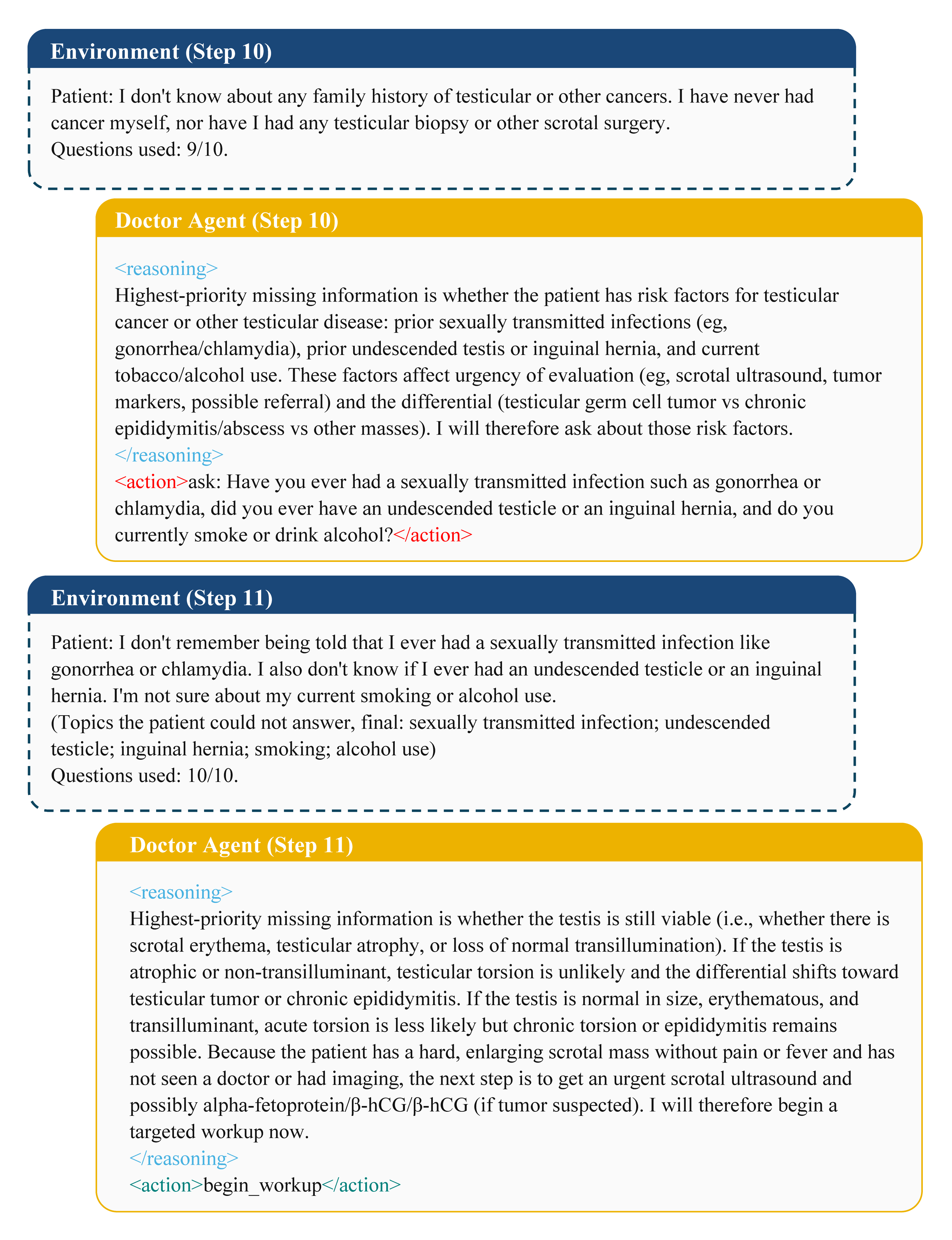}
    \label{fig:reasoning-5}
\end{figure}

\begin{figure}[!htbp]
    \centering
    \includegraphics[width=\textwidth]{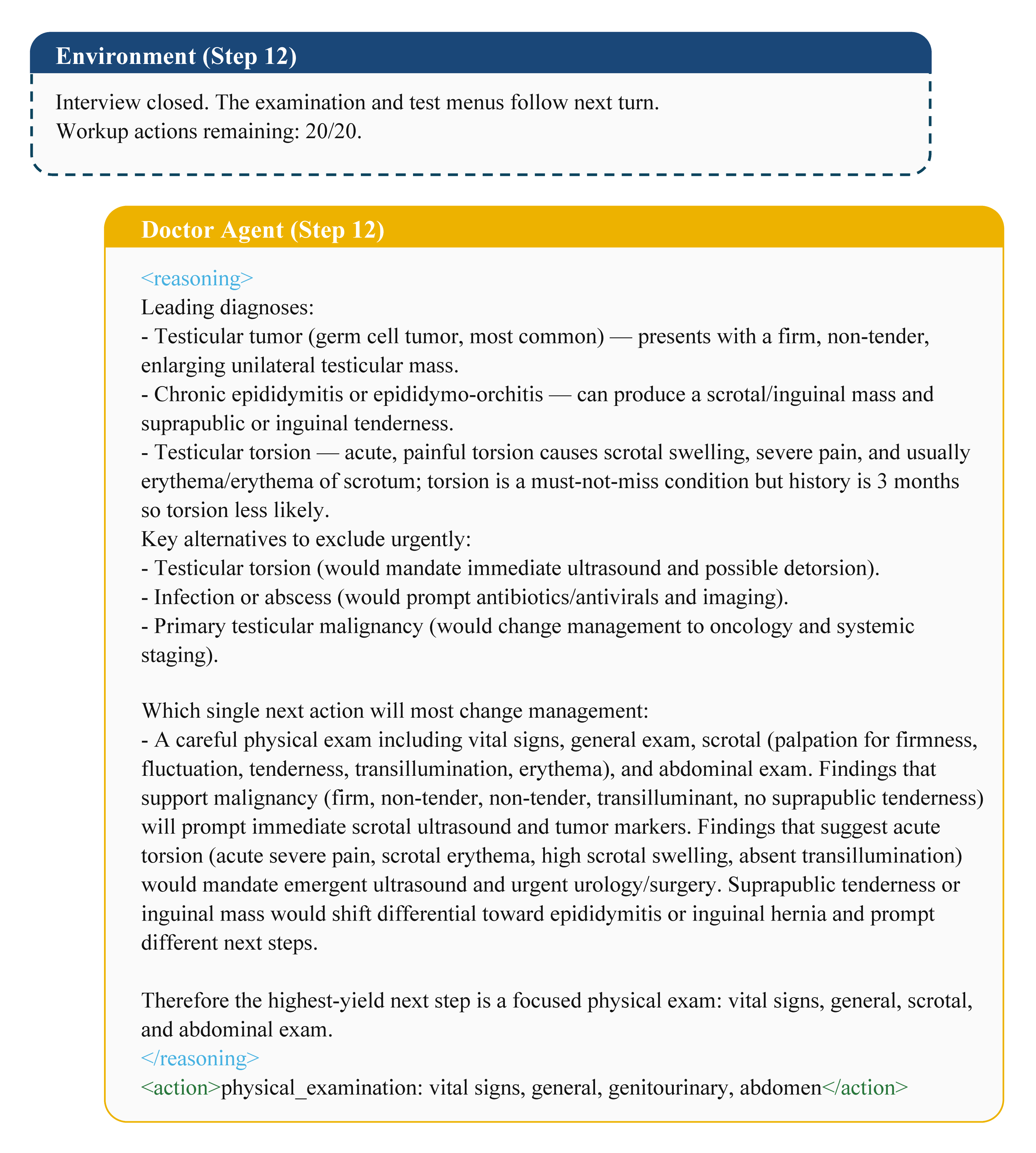}
    \label{fig:reasoning-6}
\end{figure}

\begin{figure}[!htbp]
    \centering
    \includegraphics[width=\textwidth]{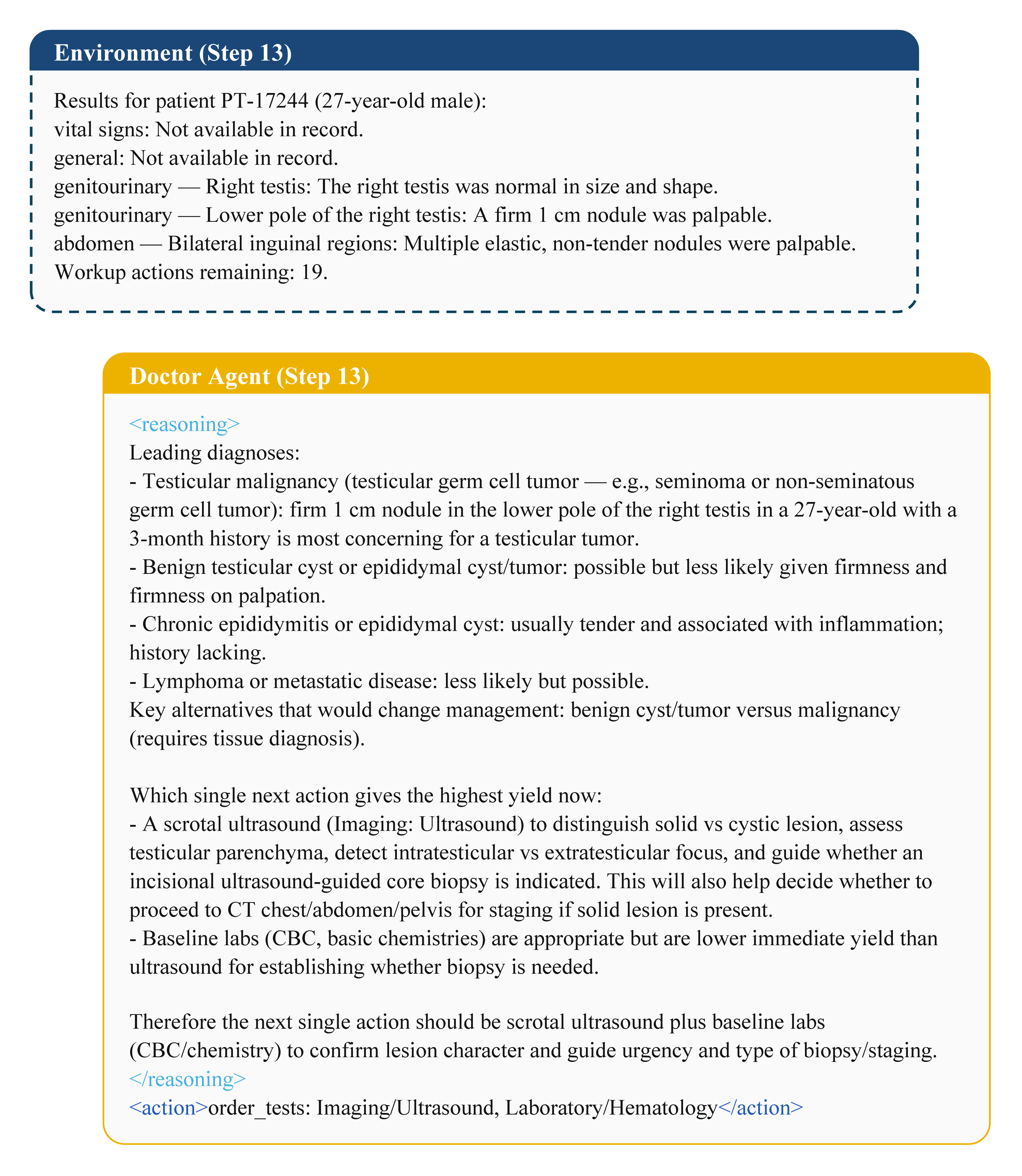}
    \label{fig:reasoning-7}
\end{figure}

\begin{figure}[!htbp]
    \centering
    \includegraphics[width=\textwidth]{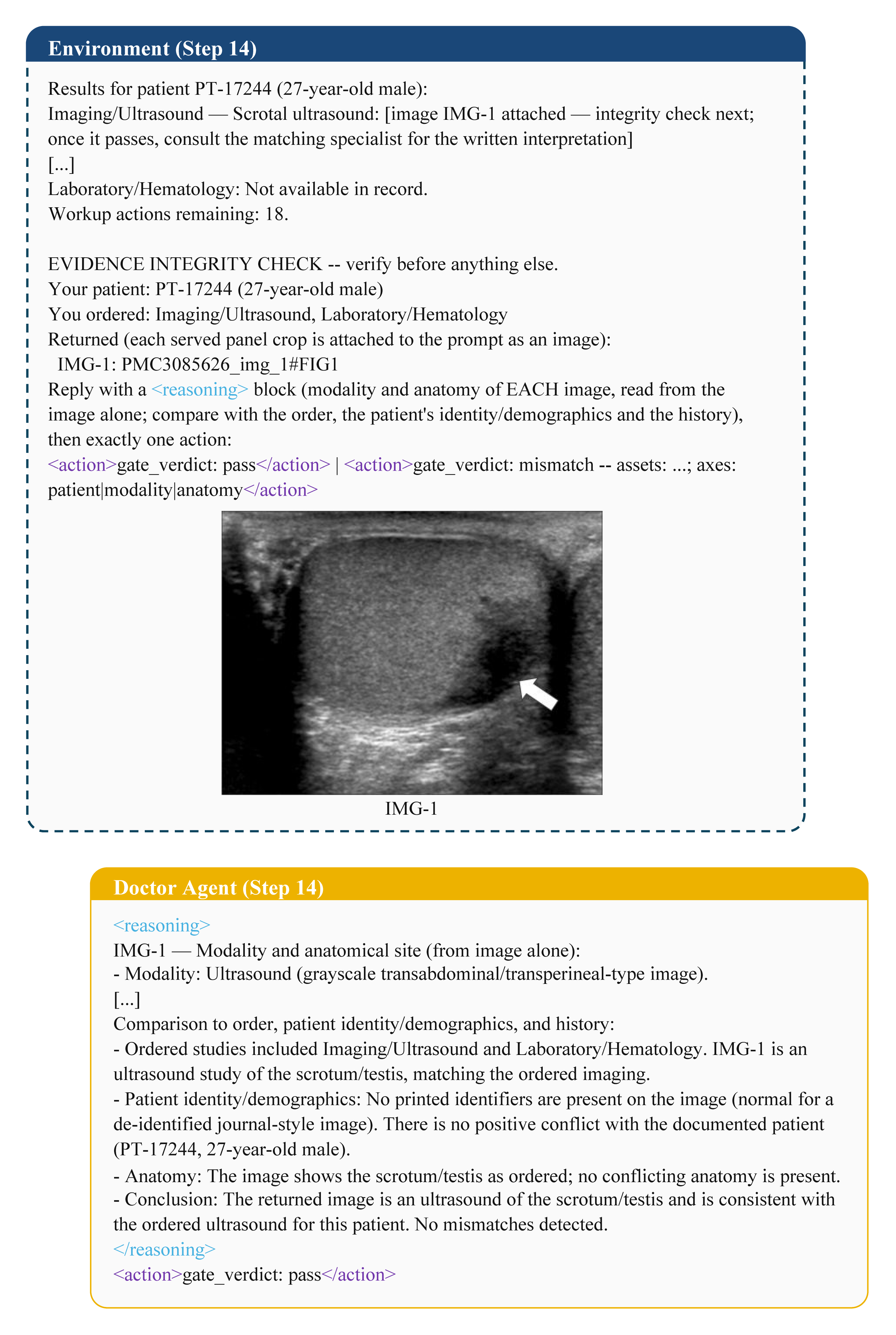}
    \label{fig:reasoning-8}
\end{figure}

\begin{figure}[!htbp]
    \centering
    \includegraphics[width=\textwidth]{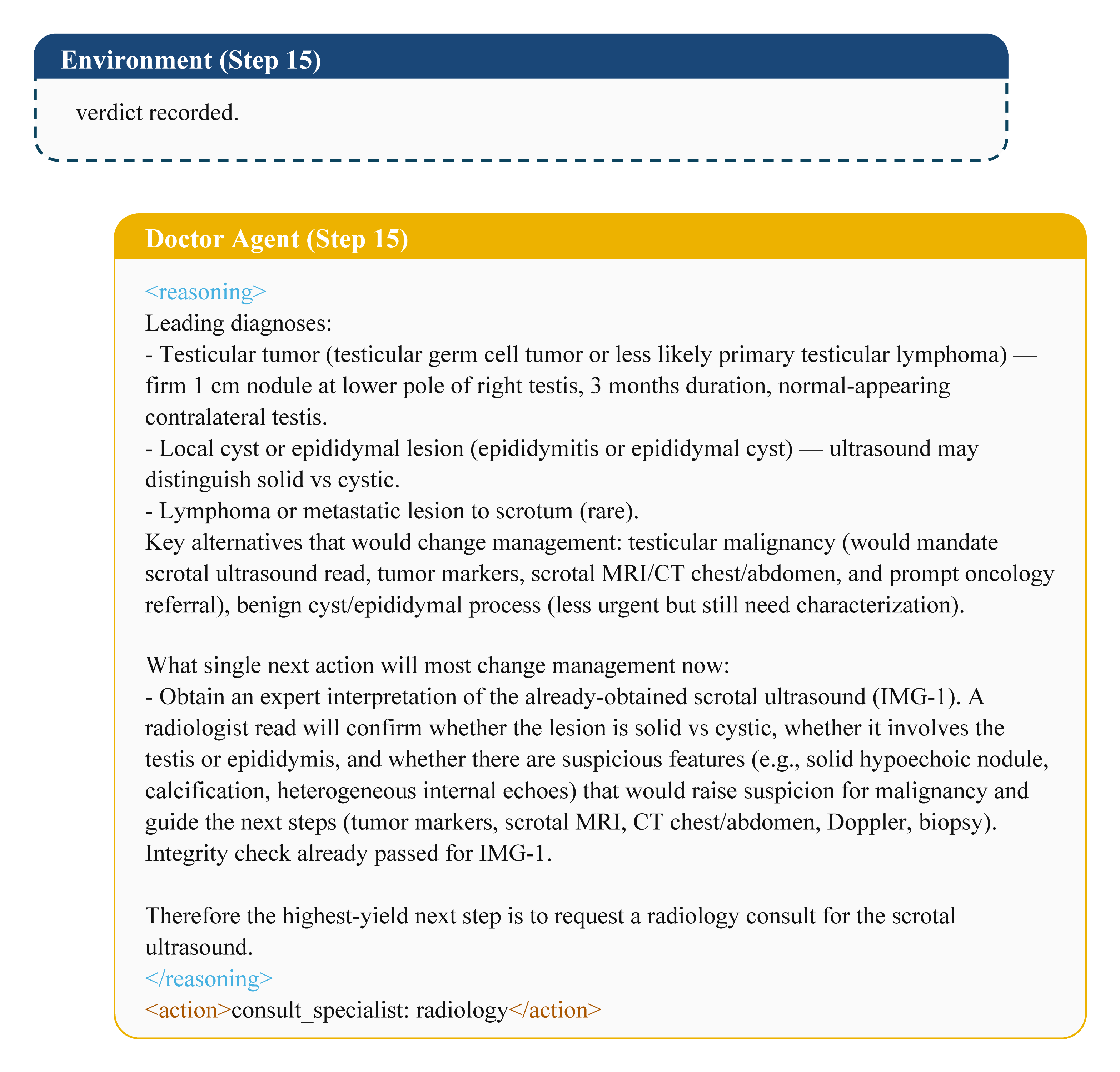}
    \label{fig:reasoning-9}
\end{figure}

\begin{figure}[!htbp]
    \centering
    \includegraphics[width=\textwidth]{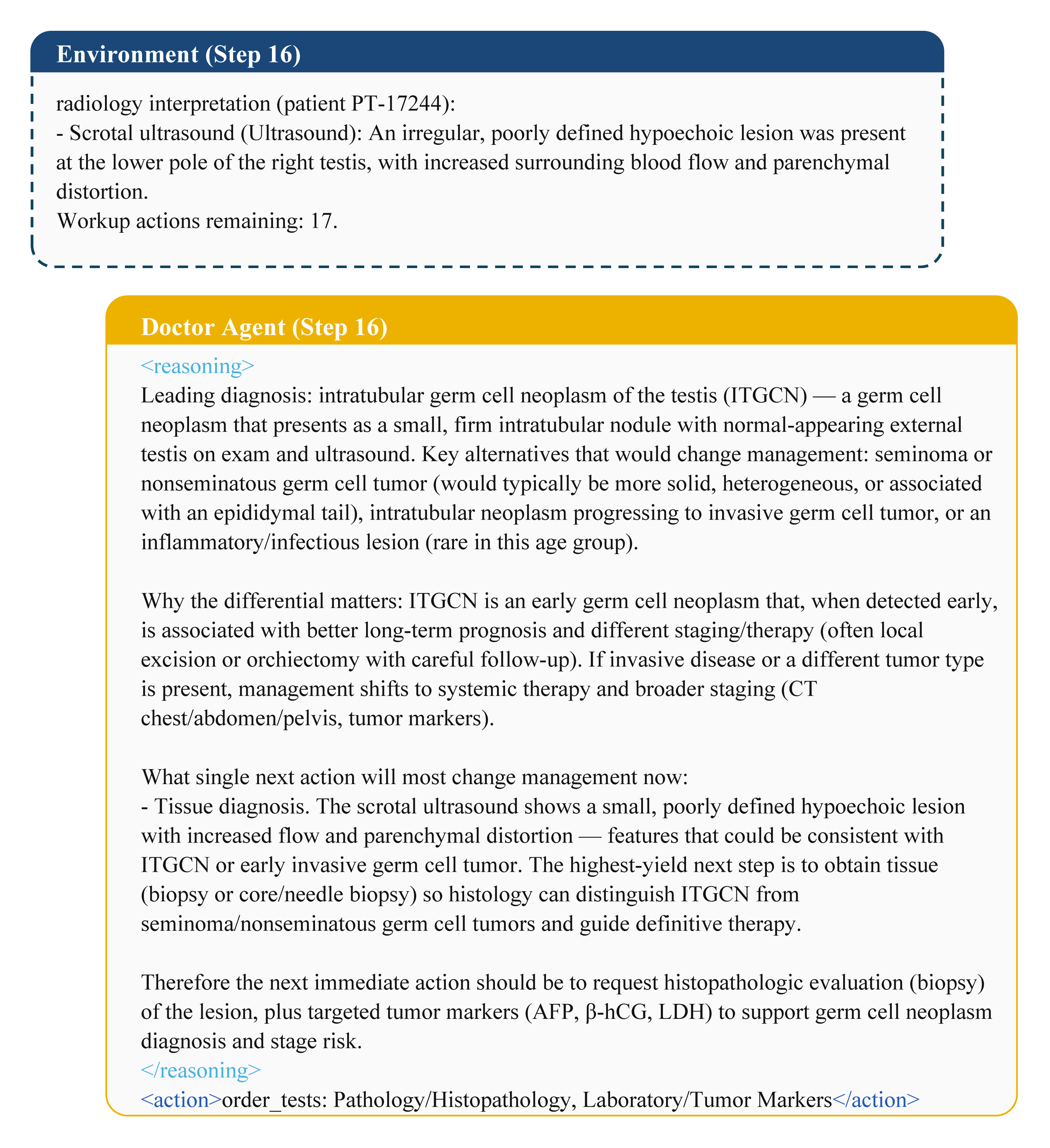}
    \label{fig:reasoning-10}
\end{figure}

\begin{figure}[!htbp]
    \centering
    \includegraphics[width=\textwidth]{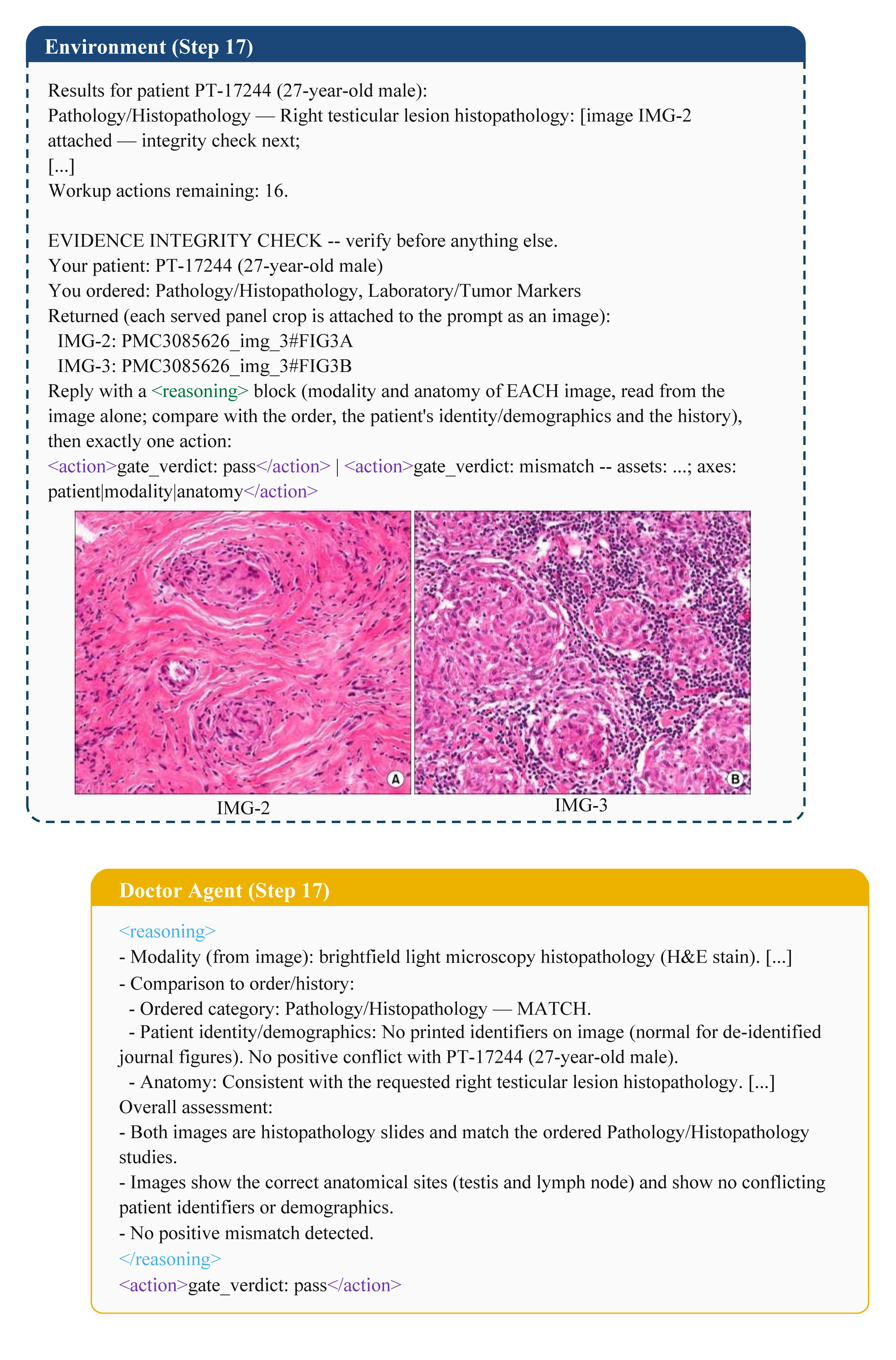}
    \label{fig:reasoning-11}
\end{figure}

\begin{figure}[!htbp]
    \centering
    \includegraphics[width=\textwidth]{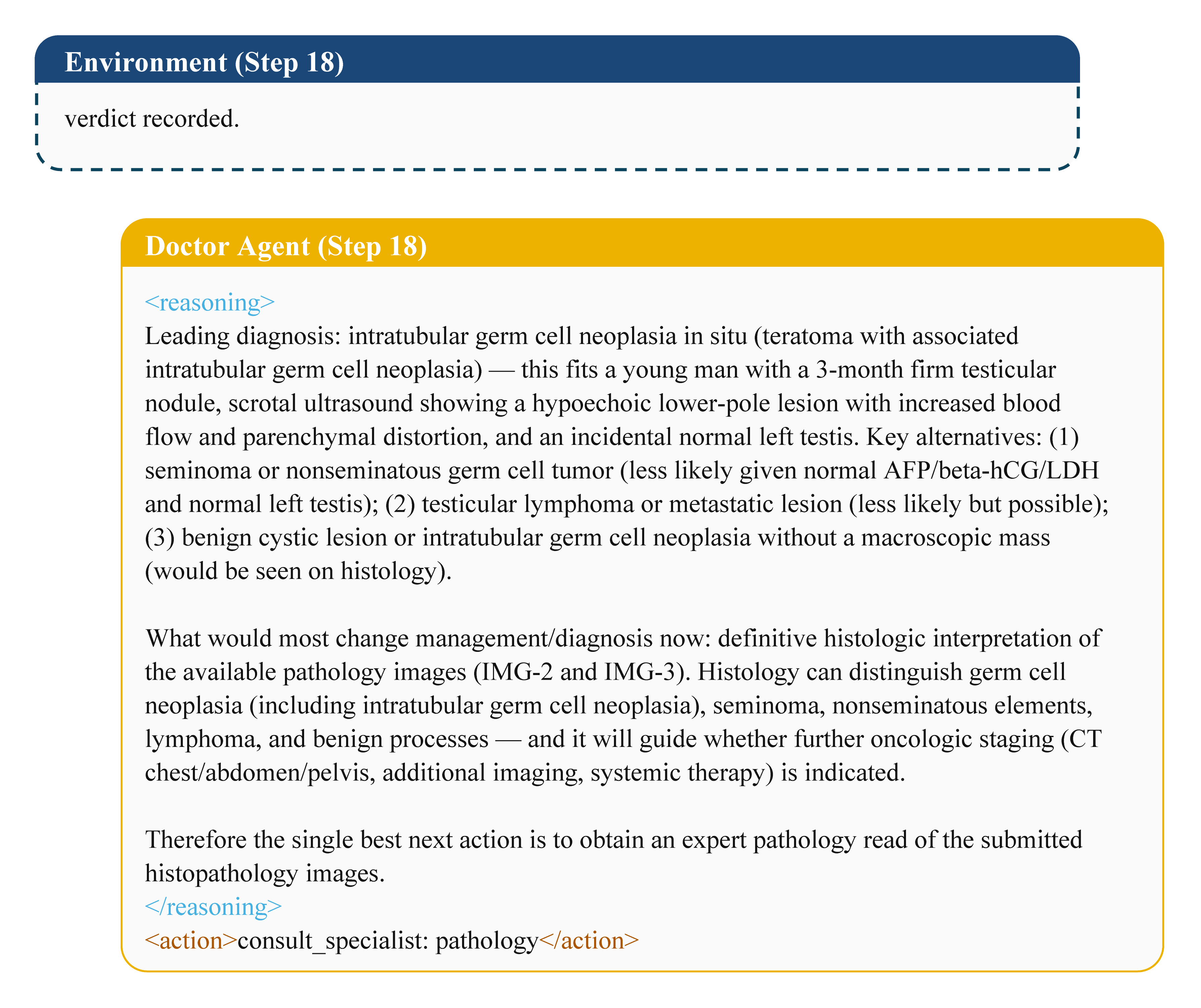}
    \label{fig:reasoning-12}
\end{figure}

\begin{figure}[!htbp]
    \centering
    \includegraphics[width=\textwidth]{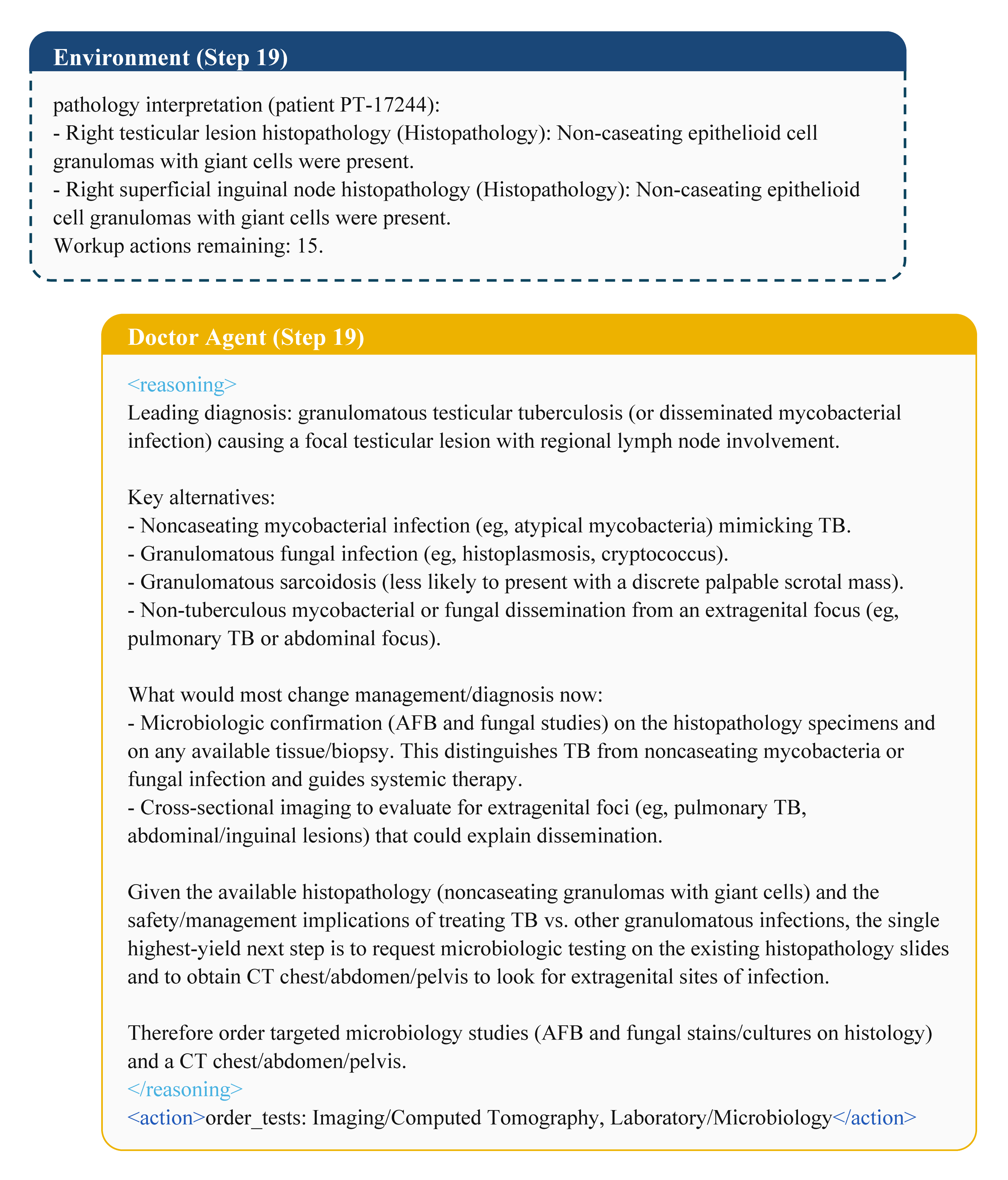}
    \label{fig:reasoning-13}
\end{figure}

\begin{figure}[!htbp]
    \centering
    \includegraphics[width=\textwidth]{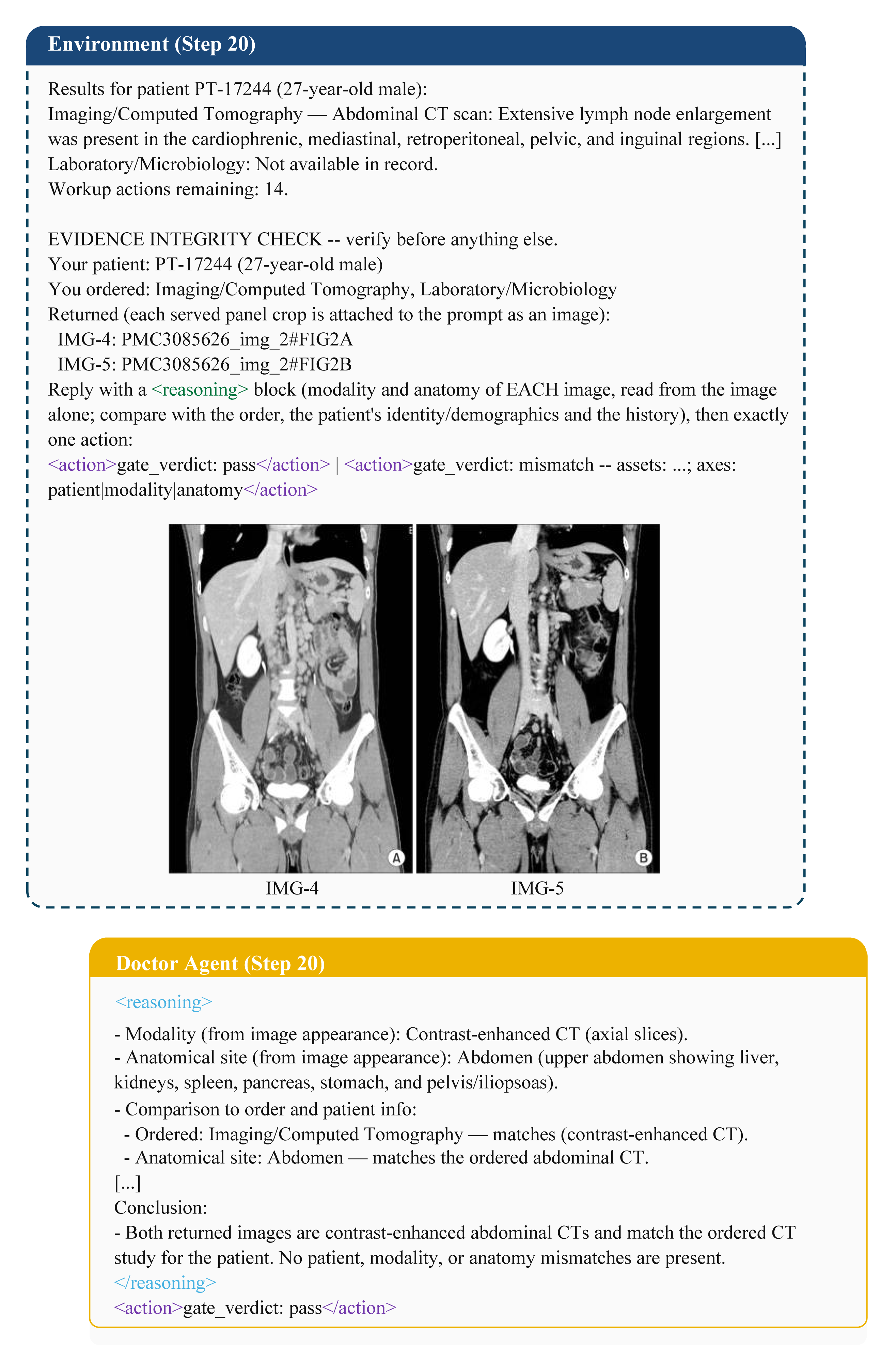}
    \label{fig:reasoning-14}
\end{figure}

\begin{figure}[!htbp]
    \centering
    \includegraphics[width=\textwidth]{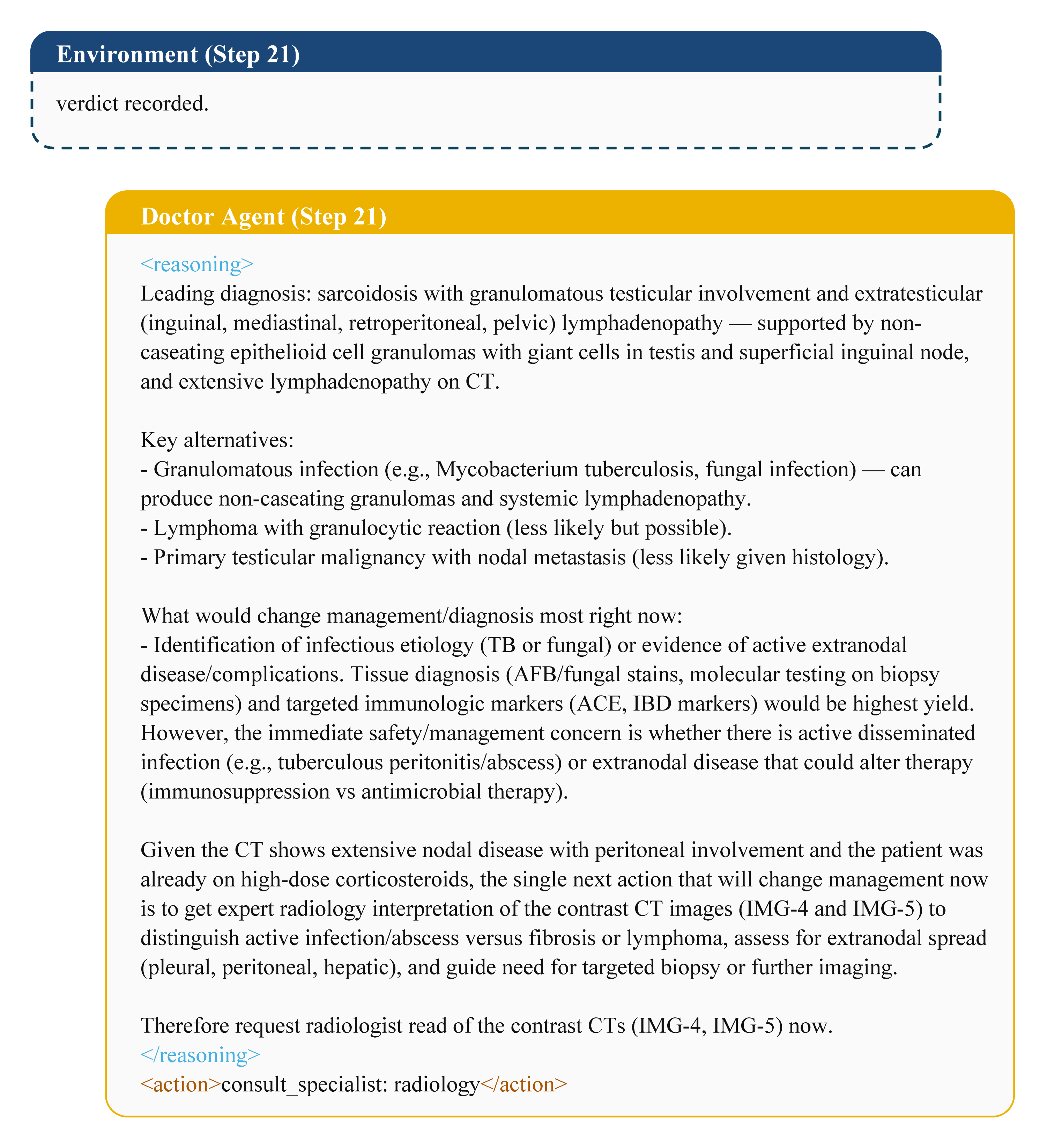}
    \label{fig:reasoning-15}
\end{figure}

\begin{figure}[!htbp]
    \centering
    \includegraphics[width=\textwidth]{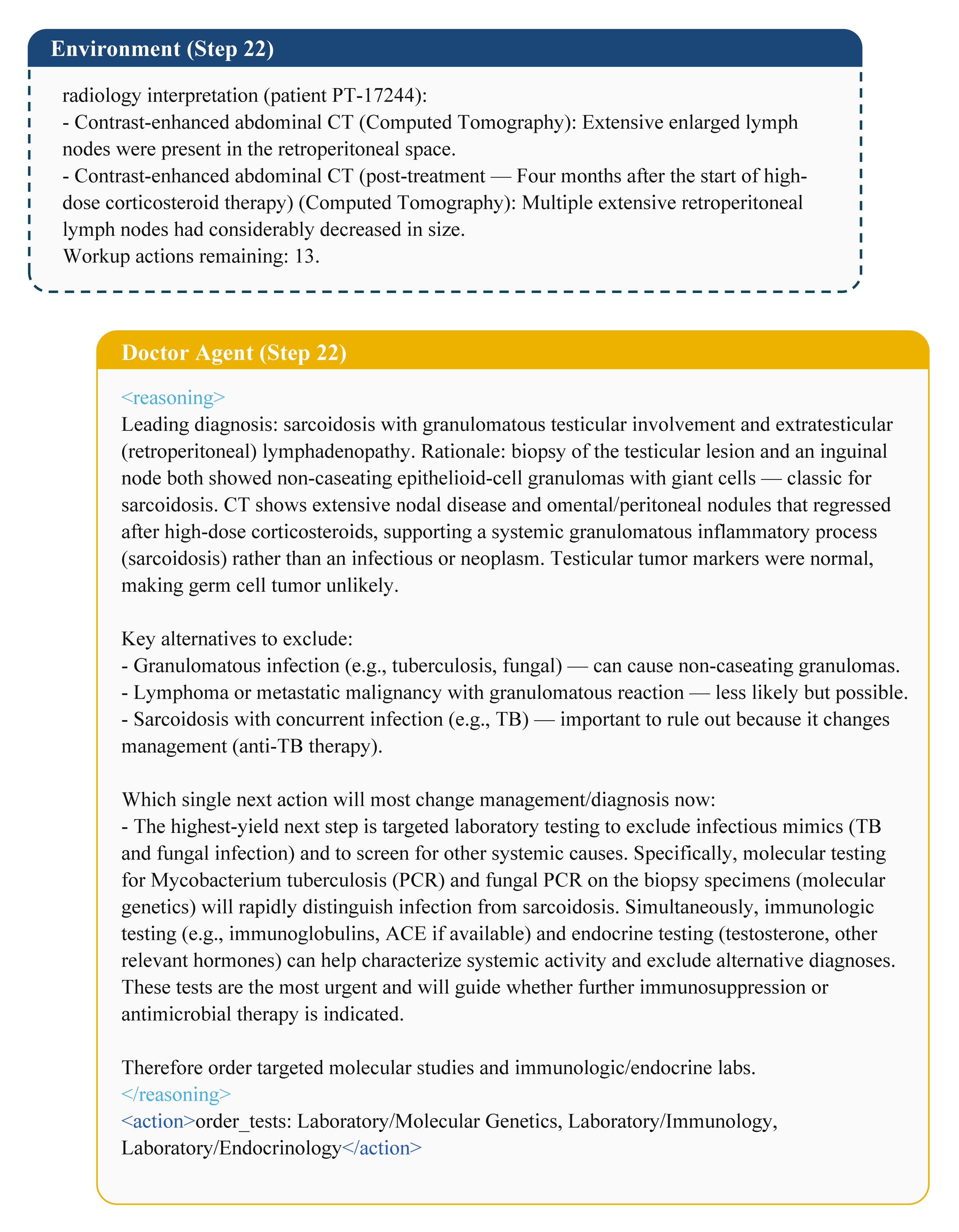}
    \label{fig:reasoning-16}
\end{figure}

\begin{figure}[!htbp]
    \centering
    \includegraphics[width=\textwidth]{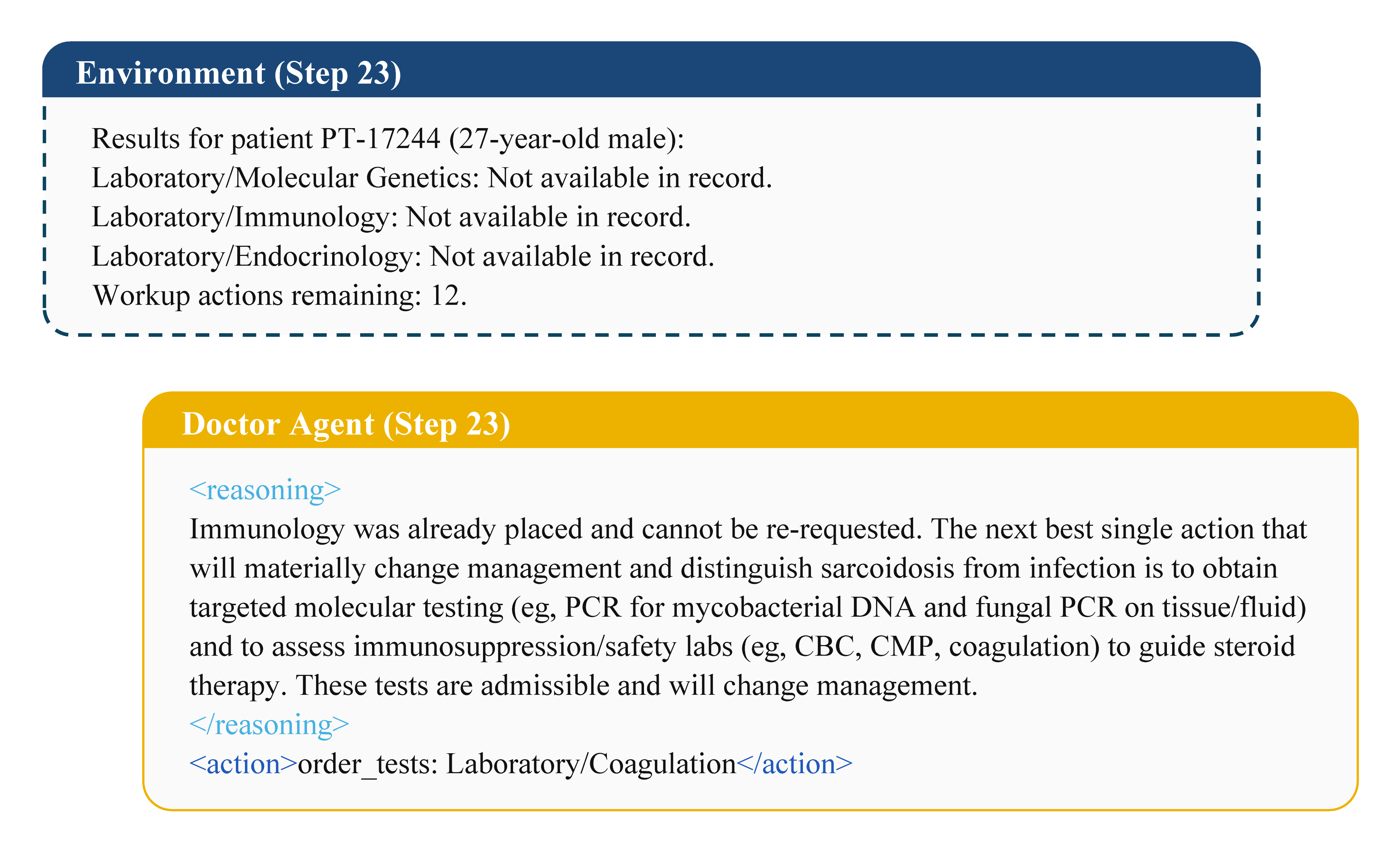}
    \label{fig:reasoning-17}
\end{figure}

\begin{figure}[!htbp]
    \centering
    \includegraphics[width=\textwidth]{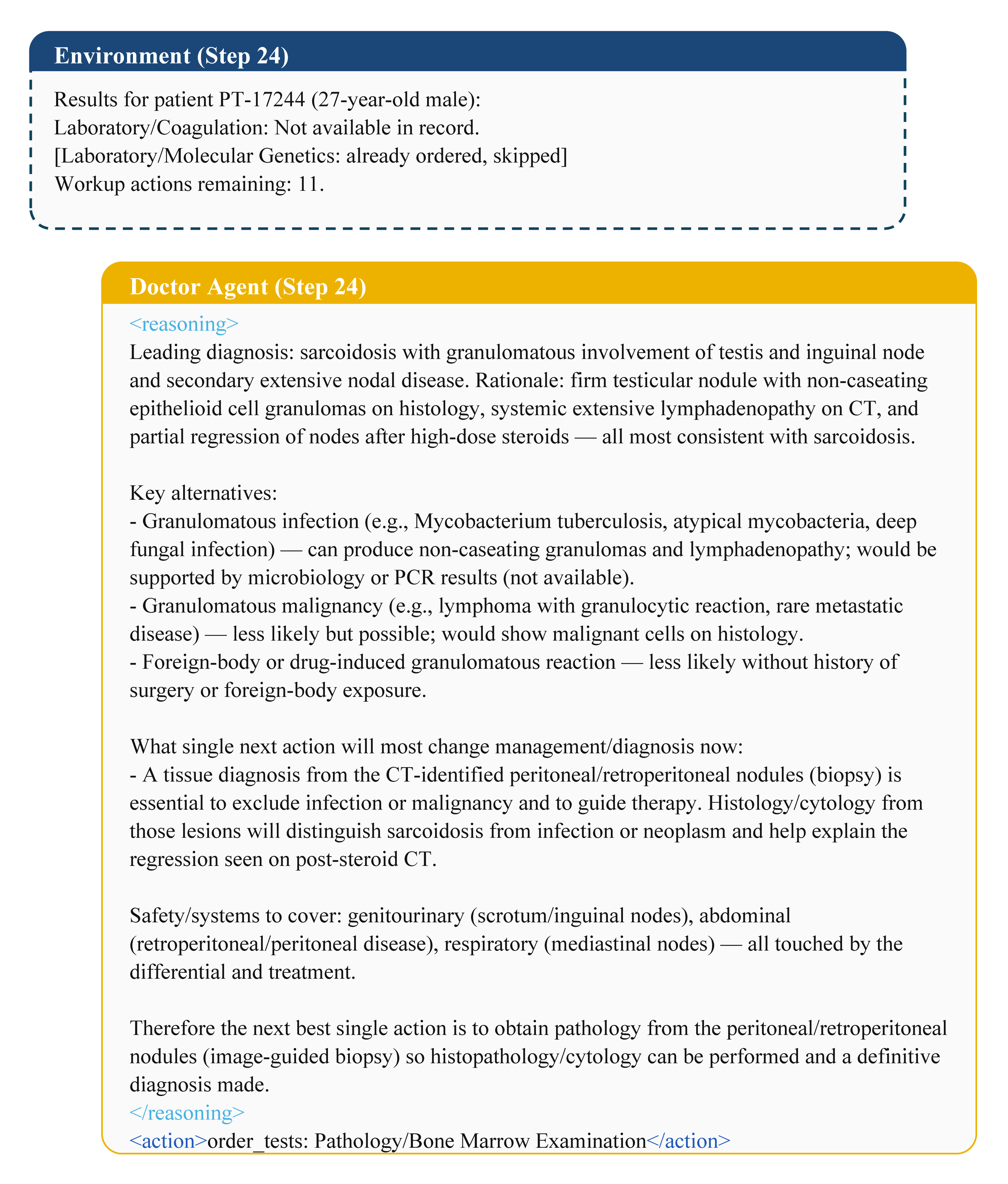}
    \label{fig:reasoning-18}
\end{figure}

\begin{figure}[!htbp]
    \centering
    \includegraphics[width=\textwidth]{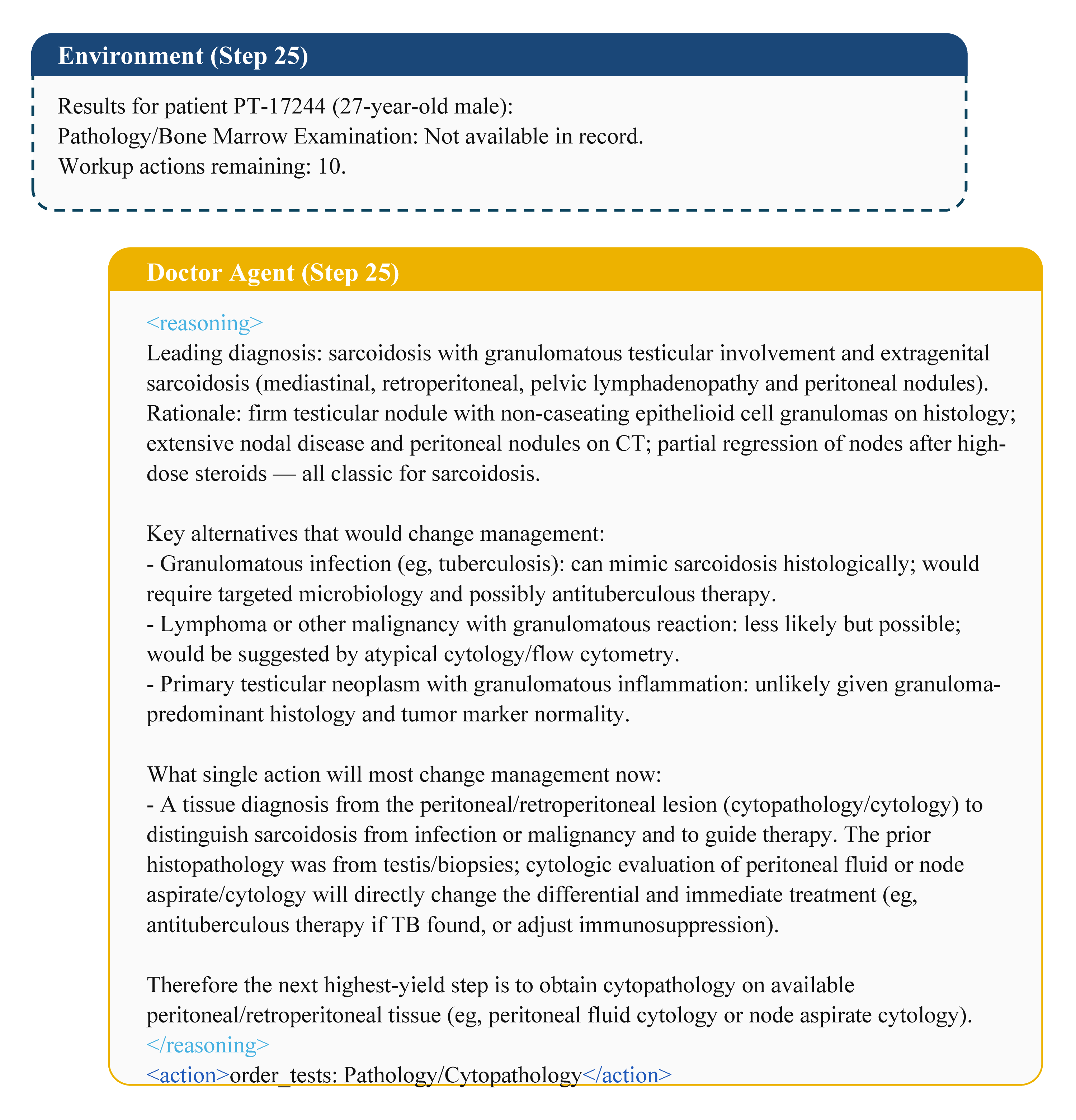}
    \label{fig:reasoning-19}
\end{figure}

\begin{figure}[!htbp]
    \centering
    \includegraphics[width=\textwidth]{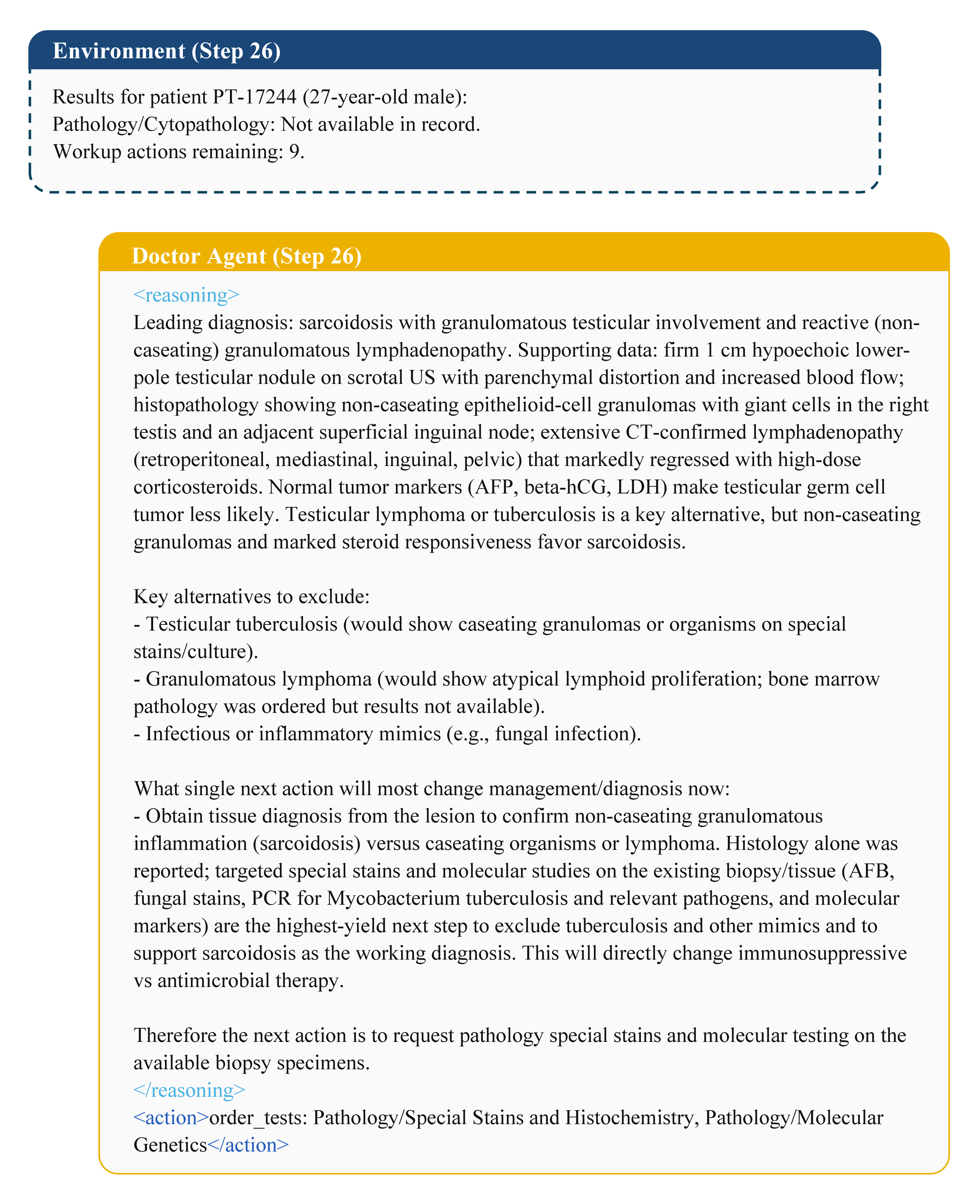}
    \label{fig:reasoning-20}
\end{figure}

\begin{figure}[!htbp]
    \centering
    \includegraphics[width=\textwidth]{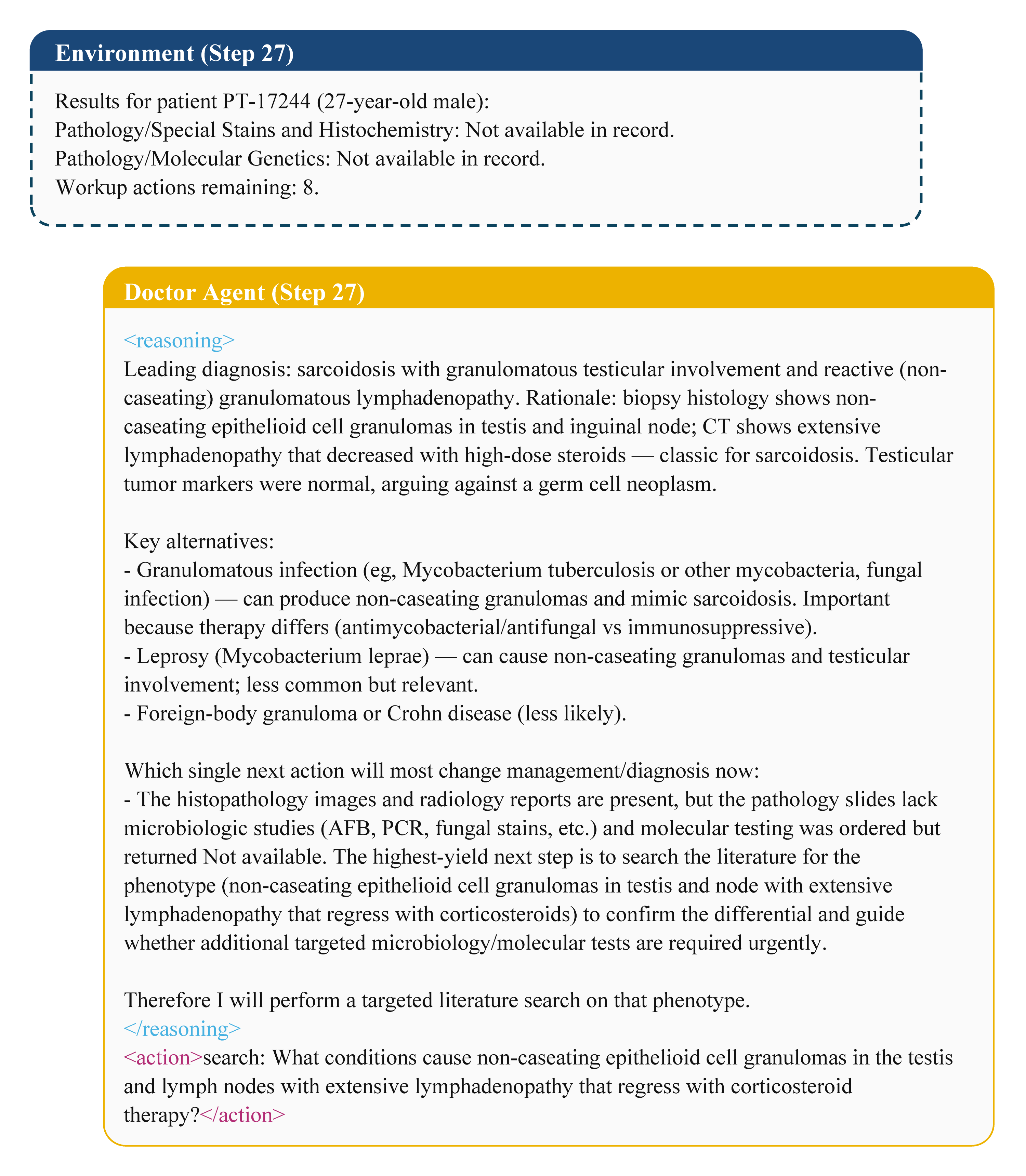}
    \label{fig:reasoning-21}
\end{figure}

\begin{figure}[!htbp]
    \centering
    \includegraphics[width=\textwidth]{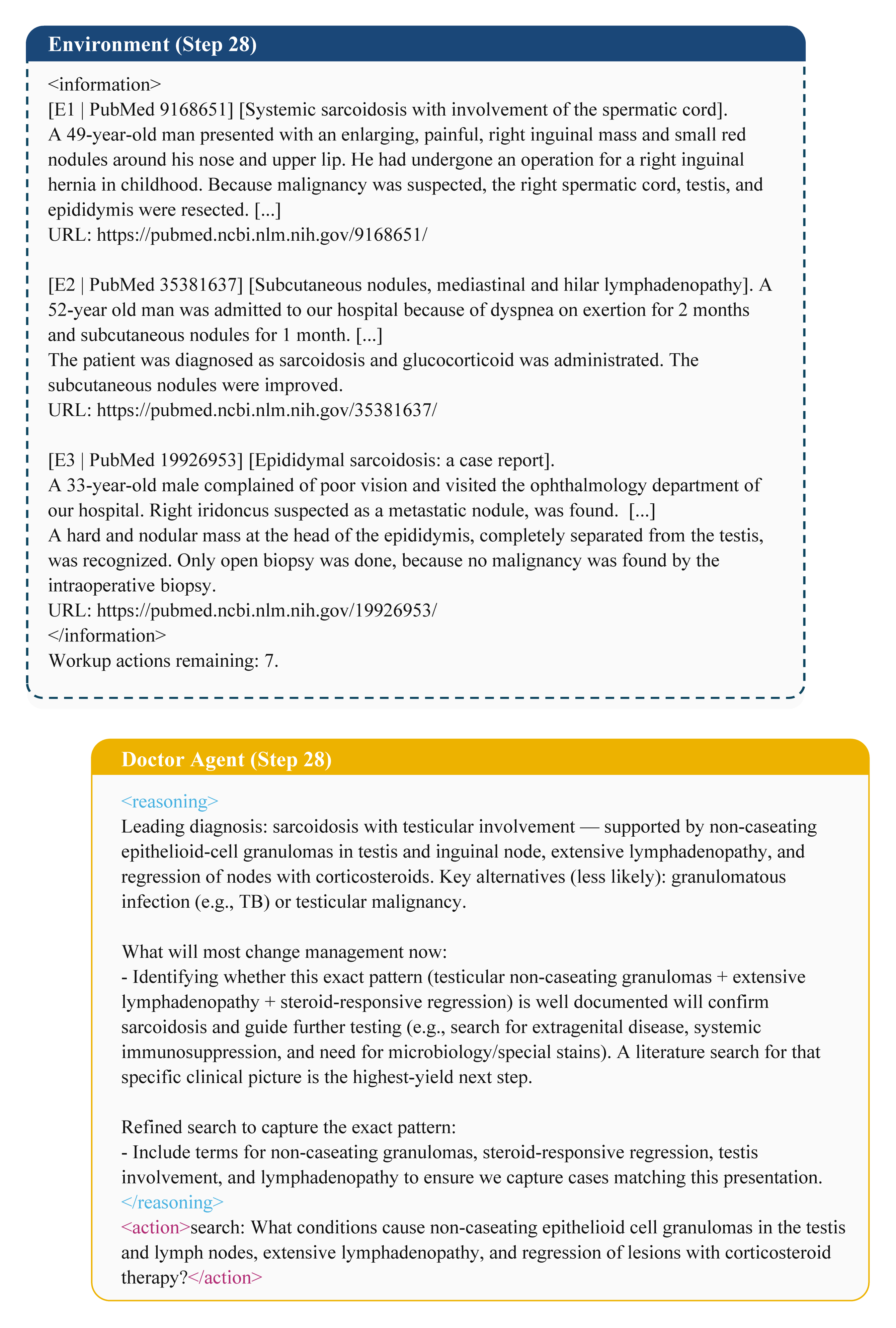}
    \label{fig:reasoning-22}
\end{figure}

\begin{figure}[!htbp]
    \centering
    \includegraphics[width=\textwidth]{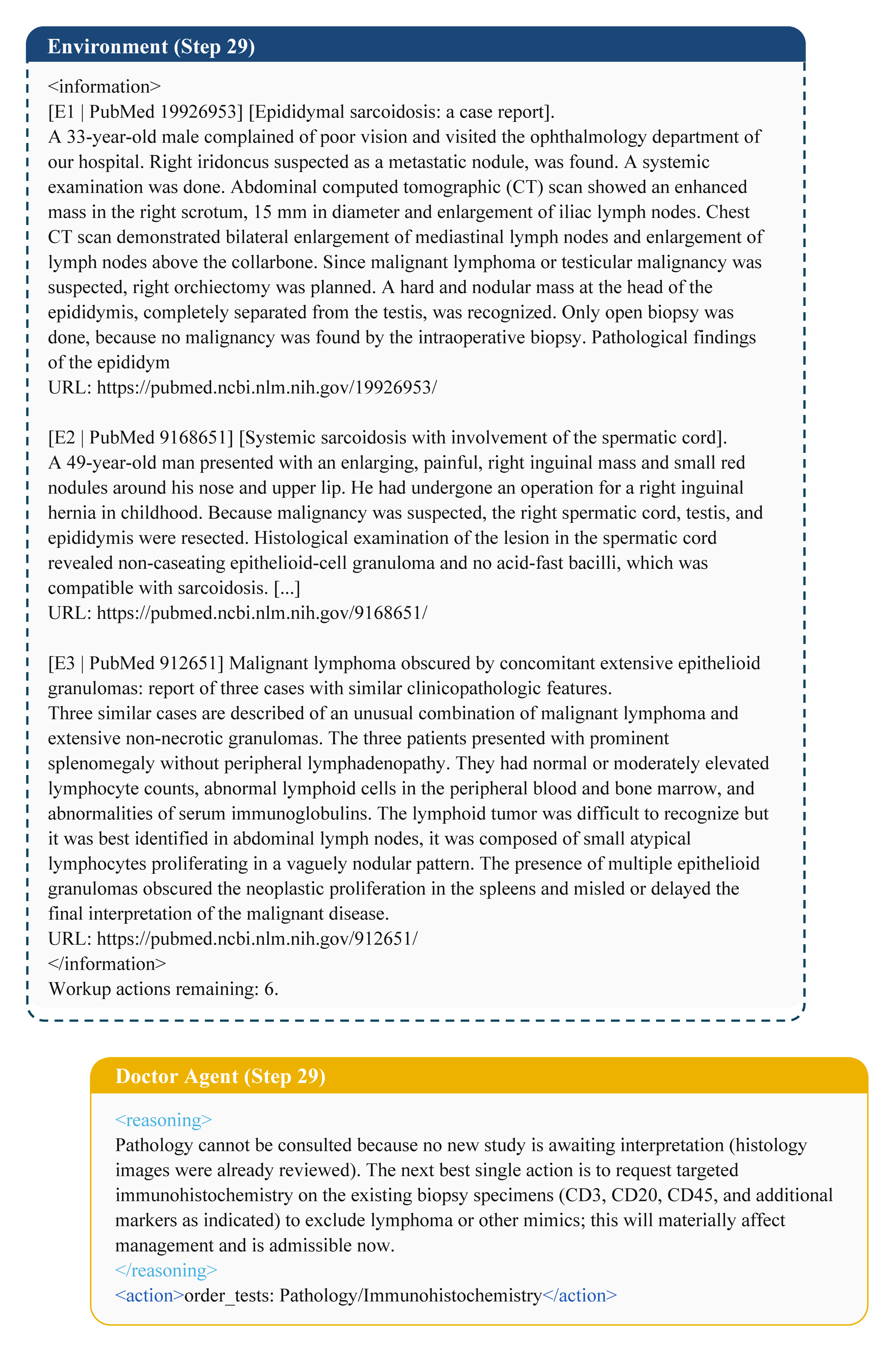}
    \label{fig:reasoning-23}
\end{figure}

\begin{figure}[!htbp]
    \centering
    \includegraphics[width=\textwidth]{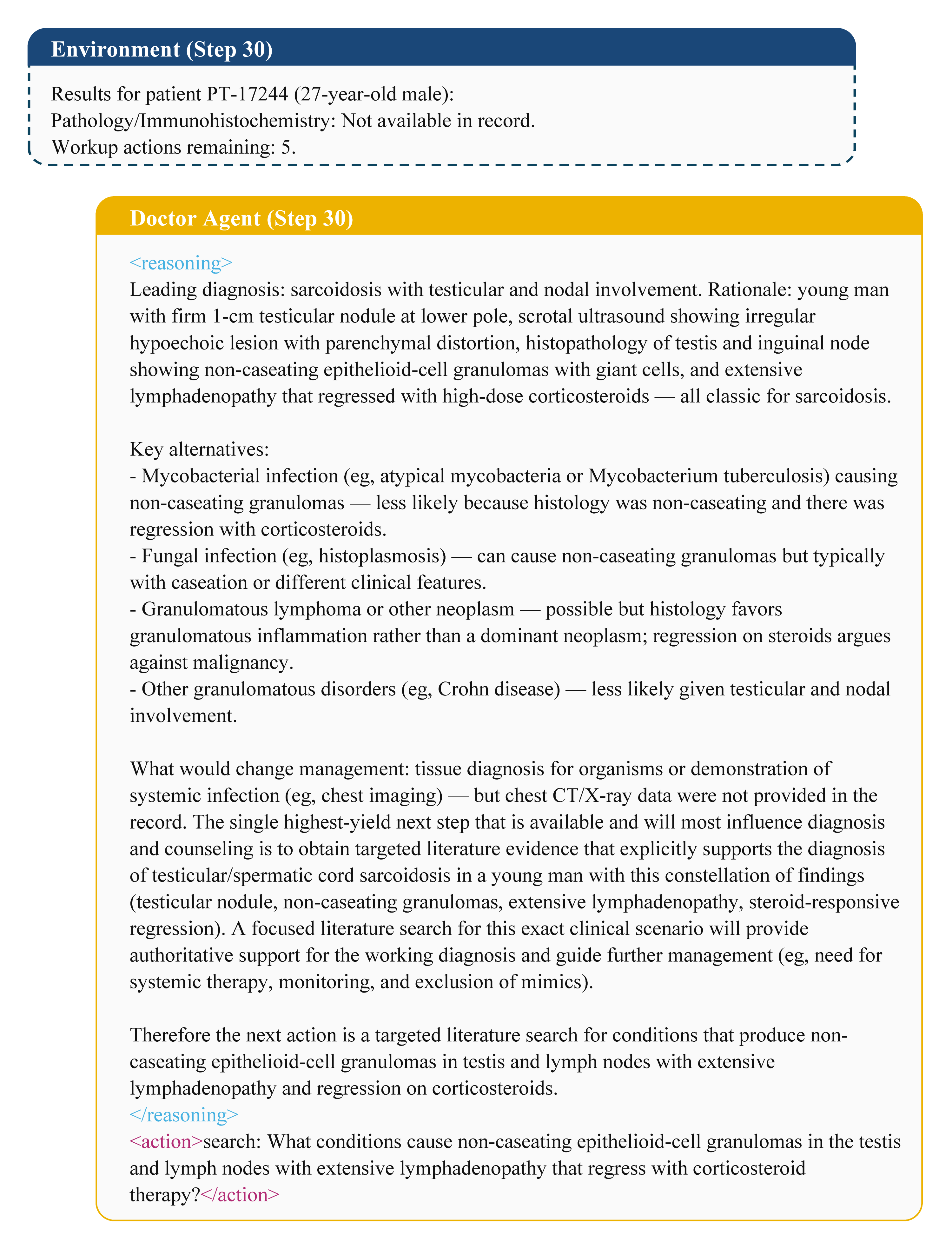}
    \label{fig:reasoning-24}
\end{figure}

\begin{figure}[!htbp]
    \centering
    \includegraphics[width=\textwidth]{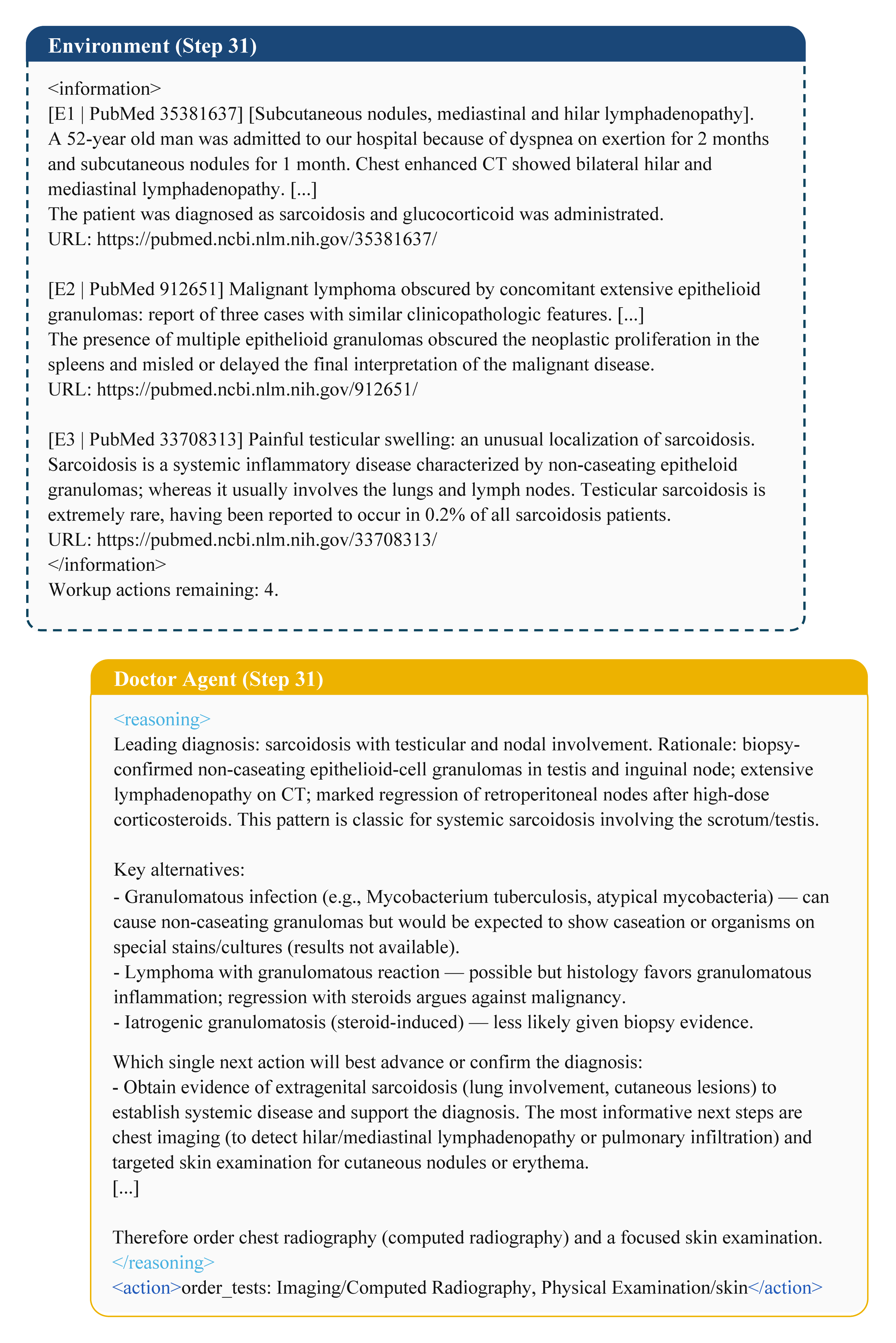}
    \label{fig:reasoning-25}
\end{figure}

\begin{figure}[!htbp]
    \centering
    \includegraphics[width=\textwidth]{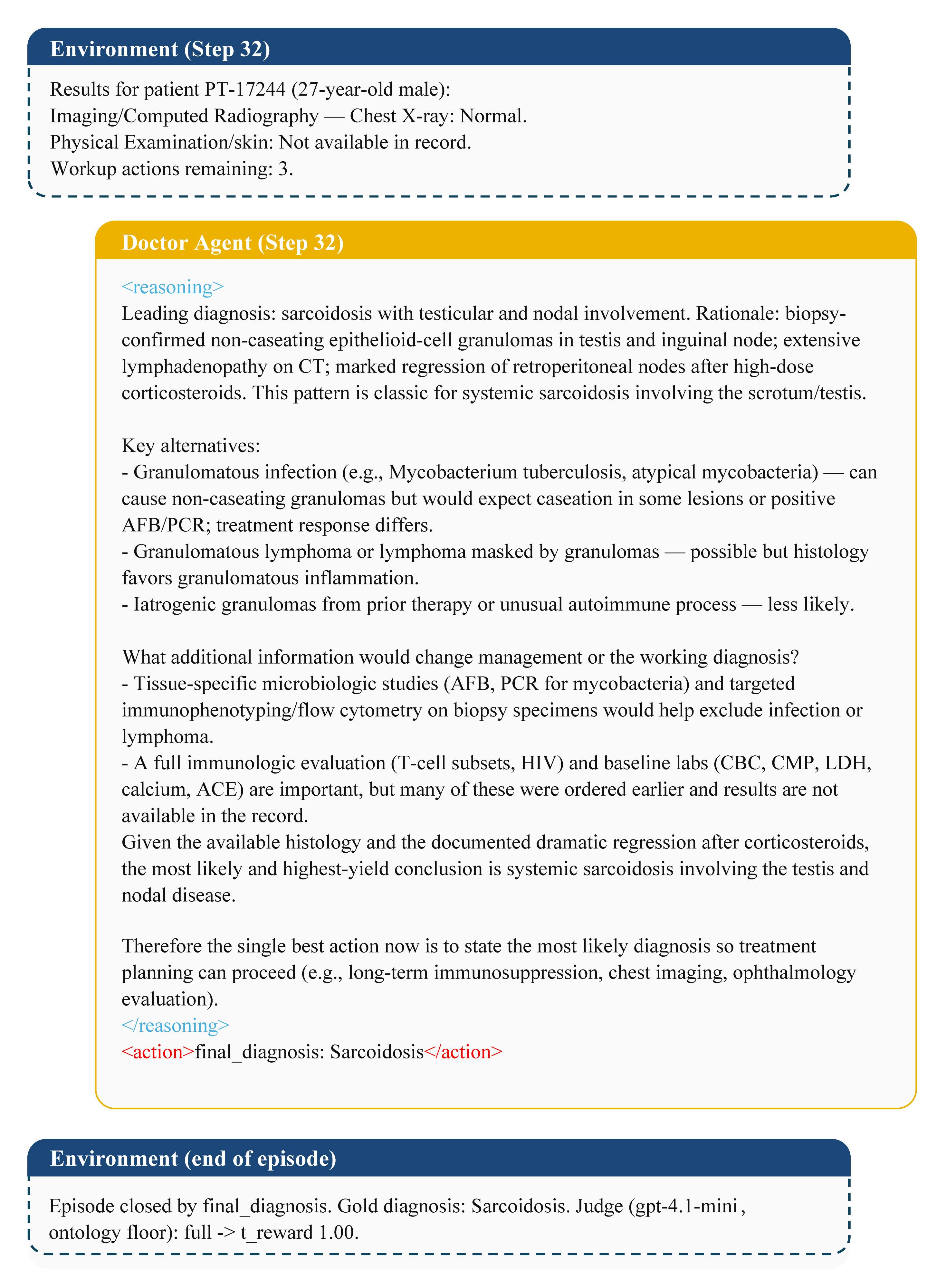}
    \label{fig:reasoning-26}
\end{figure}

%% file: appendix/h_OSEC.tex
\section{OSCE style Format}
\begin{figure*}[t]
    \centering
    \includegraphics[width=0.95\textwidth]
    {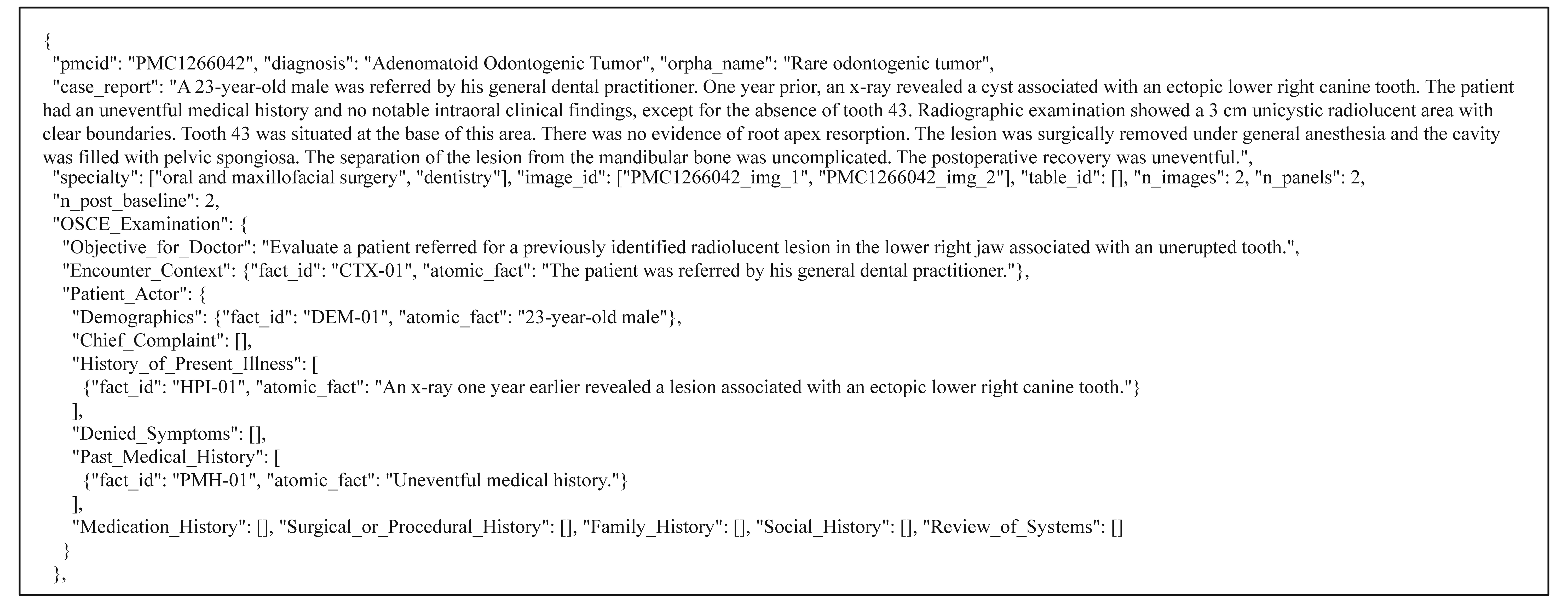}

    \vspace{2mm}

    \includegraphics[width=0.95\textwidth]
    {figures/OSEC/clinical_example_horizontal_frames_Page_2.jpg}

    \caption{Example of a multi-turn clinical diagnostic trajectory in our interactive environment.}
    \label{fig:clinical-example}
\end{figure*}